\documentclass[lettersize,journal]{IEEEtran}
\usepackage{amsmath,amsfonts}
\usepackage{algpseudocode}
\usepackage{algorithm}
\usepackage{algorithmicx}

\usepackage{array}
\usepackage{textcomp}
\usepackage{stfloats}
\usepackage{url}
\usepackage{verbatim}

\usepackage{graphicx}
\graphicspath{{./figures/}}

\usepackage{cite}
\usepackage{multirow}
\usepackage{caption}
\usepackage{subcaption}
\usepackage{url}

\usepackage[dvipsnames,table]{xcolor}

\newcommand{\boldx}{\mbox{${\mathbf x}$}}

\newcommand{\IKEMO}{IK-EMO}

\begin{document}
\title{An Interactive Knowledge-based Multi-objective Evolutionary Algorithm Framework for Practical Optimization Problems}


\author{Abhiroop~Ghosh,
	        Kalyanmoy~Deb,
	        Erik~Goodman, and
	        Ronald~Averill
	\thanks{Authors are with Michigan State University, East Lansing, MI 48824, USA, e-mail: \{ghoshab1, kdeb, goodman, averillr\}@msu.edu (see https://www.coin-lab.org).}}


\maketitle
\begin{abstract}
Experienced users often have useful knowledge and intuition in solving real-world optimization problems. User knowledge can be formulated as inter-variable relationships to assist an optimization algorithm in finding good solutions faster. Such inter-variable interactions can also be automatically learned from high-performing solutions discovered at intermediate iterations in an optimization run -- a process called \textit{innovization}. These relations, if vetted by the users, can be enforced among newly generated solutions to steer the optimization algorithm towards practically promising regions in the search space. Challenges arise for large-scale problems where the number of such variable relationships may be high. 
This paper proposes an interactive knowledge-based evolutionary multi-objective optimization (\IKEMO) framework that extracts hidden variable-wise relationships as knowledge from evolving high-performing solutions, shares them with users to receive feedback, and applies them back to the optimization process to improve its effectiveness. The knowledge extraction process uses a systematic and elegant graph analysis method which scales well with number of variables. 
The working of the proposed \IKEMO\ is demonstrated on three large-scale real-world engineering design problems. The simplicity and elegance of the proposed knowledge extraction process and achievement of high-performing solutions quickly indicate the power of the proposed framework. The results presented should motivate further such interaction-based optimization studies for their routine use in practice.  

\end{abstract}


\section{Introduction}
%

\IEEEPARstart{F}{or} practical multi-objective optimization problems (MOPs), additional knowledge may often be available from the users who have years of knowledge and experience in solving such problems. However, such information is often ignored by researchers while developing an algorithm due to concerns regarding loss of generality. But computational resources for design problems may be limited in time, cost or availability. Thus, in many cases, it may be important to use any available information that may help an optimization algorithm in finding good solutions.

For complex single-objective practical problems, evolutionary algorithms (EAs) with generic recombination and mutation operators \cite{Deb1995,Price2005} may be too slow to lead to high-performing regions of the search space. Good performance of an algorithm in solving benchmark problems such as ZDT \cite{Zitzler2000}, DTLZ \cite{Deb2002a}, and WFG \cite{Huband2005} 
does not always translate to good performance on practical problems. For such cases, creating customized algorithms leveraging additional problem information is necessary. Deb and Myburgh \cite{Deb2016} proposed a customized evolutionary algorithm that exploited the linearity of constraint structures to solve a billion-variable resource allocation problem. A micro-genetic algorithm \cite{Szollos2009} combining range-adaptation and knowledge-based re-initialization was applied to an airfoil optimization problem. Semi-independent variables \cite{Gandomi2019a} can be used to handle user-specified monotonic relationships among variables in the form of $x_i \leq x_{i+1} \leq x_{i+2} \leq\ldots\leq x_j$. 
Some techniques for combining EAs with problem knowledge are given in \cite{Landa-Becerra2008}.

Alternatives to pre-specifying problem information exist, such as cultural algorithms which encode domain knowledge inside a belief space \cite{Coello2021}. Self-organizing maps (SOMs) can provide information about important design variable clusters \cite{Obayashi2003}. Recent {\em innovization} studies \cite{Deb2006,Bandaru2013, Gaur2016a} aim to extract additional problem information from high-performing solutions during the optimization process.


Interactive optimization is when the user, referred to as the decision maker (DM), provides guidance during the optimization \cite{Xin2018}. Multiple ways to interactively specify information exist, such as aspiration levels \cite{Deb2006a}, importance of individual objectives \cite{Miettinen2008}, pairwise comparison of solutions \cite{Greco2010}, etc.

Using additional problem information comes with a set of challenges. An effective knowledge representation method needs to be designed which can be used effectively by an optimization algorithm. At the same time, it should also be comprehensible to the user. Validating any user-provided knowledge is necessary since the quality of supplied information may vary. This reduces the possibility of a premature or false convergence. The user may wish to periodically monitor and review optimization progress as well as any learned information. If necessary, the user can also supply information in a collaborative fashion \cite{Gombolay2018}. However, care needs to be taken to ensure that any user feedback does not lead the search process towards sub-optimal solutions. 
For large-scale problems resulting in a potentially huge rule set, how do we efficiently encode the rule information? How do we ensure that enforcing one rule does not violate one or more of the other rules? How can maximum rule compliance among new solutions be achieved?

This paper aims to address the issues mentioned above by proposing a generic knowledge-based evolutionary multi-objective optimization (EMO) framework with user interactivity (\IKEMO) for solving practical multi-objective optimization problems. Users can provide a preference among the learned relationships. The possibility of learned and user-provided knowledge being imperfect is also taken into account and the algorithm can adjust the extent of their influence accordingly. \IKEMO\ performance is demonstrated on three practical MOPs.

\section{Variable Relationships as Knowledge in an Optimization Task}
\label{sec:knowledge}
Knowledge is a generic term and can be interpreted in many different ways depending on the context. For an optimization task, here, we restrict the definition of knowledge to be additional information provided or extracted about the optimization problem itself. Specifically, we are interested in variable-variable relationships that commonly exist in high-performing solutions of the problem. A practical optimization task minimizes a number of objectives and satisfies a number of constraints, all stated as functions of one or more variables. Thus, understanding the variable-to-variable relationships which are common to feasible solutions (each represented by a variable vector) with small objective values is critically important. A supply of such knowledge \textit{a priori} by the users, in addition to the optimization problem description, or a discovery process of such knowledge from the evolving high-performing optimization solutions, can be directly utilized by the optimization algorithm to speed up its search process. Moreover, if such knowledge is discovered during the optimization process, users will benefit from having this knowledge in addition to the optimal solutions of the problem.   

\subsection{Past studies}
\textit{Innovization} is the process of extracting commonalities among Pareto-optimal solutions, first proposed by Deb and Srinivasan \cite{Deb2006}. 
The basic principle of innovization is to generate rules representing inter-variable relationships in simple forms such as power laws $(x_i x_j^{b} = c)$. In \cite{Bandaru2015}, the authors have proposed a method which is able to express relationships involving operators like summation $ (+) $, difference $ (-) $, product $(\times)$, etc. 

Bandaru and Deb \cite{Bandaru2013} introduced the concept of higher- and lower-level innovization. 
A genetic programming-based innovization framework was proposed in \cite{Bandaru2015} and was applied on an inventory management problem. 
An MOEA combined with a local search procedure was employed in \cite{Deb2012a} to ensure faster convergence. A combination of innovization and data mining approaches were used in \cite{Ng2013} to achieve faster convergence. Gaur and Deb \cite{Gaur2016a} proposed an adaptive innovization method that treats the innovization process as a machine learning problem and repairs the solutions directly, based on the learned model. A combination of user guidance and inequality relation-based online innovization \cite{Ghosh2021} was used to solve three practical problems.


\subsection{Structure of rules considered in this study}
For an interactive knowledge-based optimization algorithm to work, a standard form of knowledge representation is necessary which is simple enough for users to understand but has enough complexity to capture problem knowledge accurately. Using algebraic expressions or `rules' is one way of representing knowledge and has been extensively used in the `innovization' literature \cite{Deb2006,Gaur2016a}. 
A rule can take the form of an equality or an inequality, as shown below:
\begin{align}
\phi(\mathbf{\boldx}) = 0,
\label{eq:rulegeneraleq}\\
\psi(\mathbf{\boldx}) \leq 0.
\label{eq:rulegeneralineq}
\end{align}
Any arbitrary form of rules involving many variables from a decision variable vector ($\mathbf{x}$) and complicated mathematical structures of functions $\phi$ or $\psi$ may be considered, but such rules would not only be difficult to learn, they would also be difficult to interpret by the user. In this study, we restrict the rules to have simple structures involving a maximum of two variables, as discussed below.

\subsubsection{Constant rule}
\label{sec:const}
This type of rule involves only one variable taking a constant value ($x_i = \kappa_i$). In terms of Equation \ref{eq:rulegeneraleq}, for the $i$-th variable, the structure of the rule becomes $\phi_{i}(\mathbf{x})= x_i - \kappa_i$. This type of rule can occur if multiple high-performing solutions are expected to have in common a fixed value of a specific variable \cite{Ghosh2020a}.

\subsubsection{Power law rule}
\label{sec:powerlaw}
Power law rules \cite{Deb2006} for two variables $x_i$ and $x_j$ can be represented by Equation~\ref{eq:rulegeneraleq} as $\phi_{ij}(\mathbf{x}) = x_i x_j^b - c$, where $b$ and $c$ are constants. This form makes power laws versatile enough to encode a wide variety of rules, such as proportionate or inversely proportionate relationships among two variables. Interestingly, an inequality power law using a $\psi$ function can also be implemented, but such a rule may represent a relationship loosely and we do not consider it here.

\subsubsection{Equality rule}
\label{sec:equiv}
This type of rule can express the equality principle of two variables $x_i$ and $x_j$ observed in high-performing solutions. In terms of Equation \ref{eq:rulegeneraleq}, $\phi_{ij}(\mathbf{x}) = x_i - x_j$ is the rule's structure. 

\subsubsection{Inequality rule}
\label{sec:ineq}
This type of rule can represent relational properties of two variables $x_i$ and $x_j$ as $x_i \leq x_j$ or $x_i \geq x_j$. In terms of Equation \ref{eq:rulegeneralineq}, $\psi_{ij}(\mathbf{x}) = x_i - x_j$ or $\psi_{ij}(\mathbf{x}) = x_j - x_i$ are the respective rules. 
For example, the radius of two beams in a truss \cite{Ghosh2021} might be related via this type of rule.

After describing the chosen rule structures, we are now ready to discuss the procedures of extracting such rules from high-performing variable vectors and applying the extracted rules to the optimization algorithm. A summary of their representations and use in our analysis are provided in Table~\ref{tab:RuleType}.

\section{Proposed Interactive Knowledge-based \IKEMO\ Framework}
\begin{figure*}[hbt]
	\centering
	\includegraphics[width=0.8\linewidth]{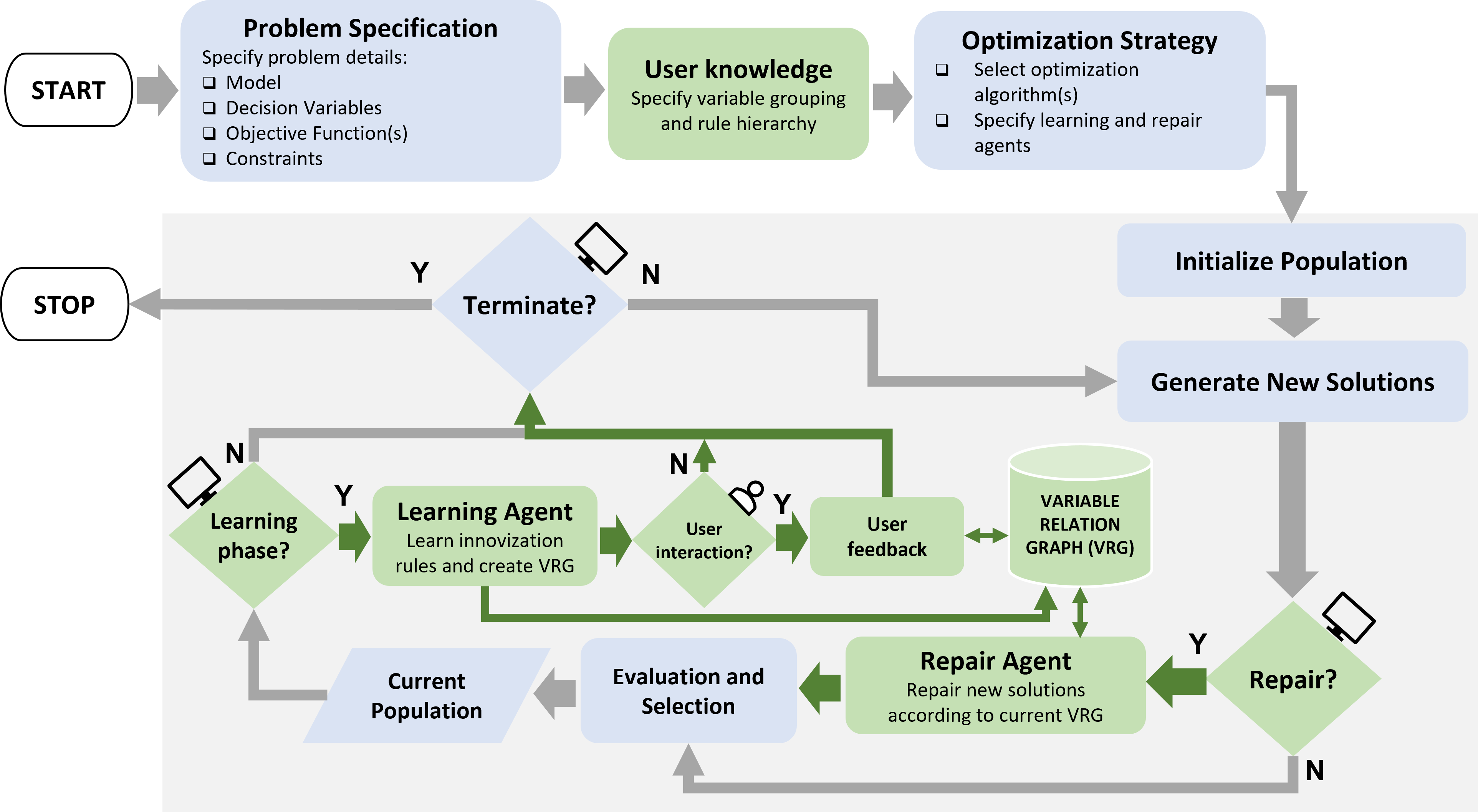}
	\caption{Interactive knowledge-based EMO framework (\IKEMO) showing user interaction, learning and repair agents. Blue blocks represent a normal EMO. Green blocks represent the components responsible for knowledge extraction and application, as well as user interaction.}
	\label{fig:imoea}
\end{figure*}

In this study, we restrict our discussions to multi-objective optimization problems, so high-performing solutions refer to the entire {\em non-dominated\/} (ND) solution set discovered by the optimization algorithm from the start of the run to the current iteration. Figure~\ref{fig:imoea} shows the proposed \IKEMO\ framework.

The framework starts with a description of the multi-objective optimization problem, as shown in the top-left box in the figure. In addition, if any additional problem information is available, that is also specified. The penultimate step before starting the optimization is to select a suitable EMO and methods to algorithmically extract and apply any problem knowledge. The subsequent sections describe the various components in more detail.

\subsection{User knowledge}
Before the start of the optimization, the user may provide some initial information which will affect how the framework operates, details of which are given below.
\subsubsection{Variable grouping}
For a problem with $n$ variables, there can be $\frac{n(n - 1)}{2}$ pairwise variable interactions. For any reasonable-sized problem, such a huge number of meaningful relationships may not exist. In practice, the user may be interested in only a handful of relationships that relate some critical decision variables. In order to reduce the complexity, variables can be divided into different groups $G_k$ for $ k = 1, 2, ..., n_g$. Each group consists of variables that the user thinks are likely to be related. Group information is specified prior to the optimization. If no group specification exists, then all $N$ variables are considered as part of a single group and all $\frac{n(n - 1)}{2}$ pairwise variable combinations will be considered. Inter-group relationships are not discoverable under this scheme. Variables that are not part of any group are assumed not to be related to other variables. For example, assume there are two variable groups $G_1 = \{2, 3, 5\}$ and $G_2 = \{1, 4, 7\}$ for an 8-variable problem. For $G_1$ all pairwise combinations ($x_2, x_3$), ($x_2, x_5$), and ($x_3, x_5$) will be checked for the existence of any possible relationships. A similar process is repeated for $G_2$. Since inter-group relationships are not explored, combinations like $(x_3, x_7)$ will not be considered. Variables $x_6,$ and $x_8$ are not part of any group, hence they are assumed not to be related to the other variables in any meaningful way. 

\subsubsection{Rule hierarchy}
\label{sec:ruleHierarchy}
In the proposed framework we consider four types of rules as presented in Section \ref{sec:knowledge}. A pair of variables may be related by more than one rule type. In that case, we will select one type of rule according to a predefined hierarchy. At the start of the optimization each rule is assigned a rank. The existence of a particular rule type for one or more variables is checked rank-wise. For example, if constant rules are ranked 1, followed by power laws (rank 2) and inequalities (rank 3), then the variables in every group will be checked for constant rules first. The variables which do not exhibit constant rules will then be checked for power laws, and so on. For relations having equal ranking, a scoring criterion needs to be used to determine which rule better represents the non-dominated (ND) front and will be used by the algorithm.

\subsection{Learning agent}
A learning agent is a procedure used to identify different innovization rules present in the ND solutions in a population. The rules involve a single variable or a pair of variables, as required by a rule's description. Each type of rule (inequality, power law, etc.) requires a different rule satisfaction condition. A score (within [0,1], as presented in Table \ref{tab:RuleType}) is assigned to each rule to quantify how well the rule represents the ND set. Different learning agents applicable to the rule types covered in Section \ref{sec:knowledge} are presented below. A summary of the various rules, their scoring procedures and satisfaction conditions are provided in Table \ref{tab:RuleType}.

\begin{table*}[!hbt]
	\renewcommand{\arraystretch}{1.5}
	\caption{Rule types and the corresponding mathematical representation. $\mathcal{X}$ represents the set of ND solutions. $x_{i}$ and $x_{j}$ refer to the $i$-th and $j$-th variables, respectively, of a ND solution $\mathbf{x} \in \mathcal{X}$. The corresponding variables in a new solution $\mathbf{x}_r$ to be repaired are labeled as $x_{ir}$ and $x_{jr}$, respectively. Normalized variables are represented by a hat ($\hat{x}_i, \hat{x}_j$). Higher ranked rules are preferred while performing repair. The score ($s$) is a measure of how well $\mathcal{X}$ follows the rule in the representation column. Satisfaction condition dictates whether $\mathbf{x}_r$ follows the respective rule.}
	\centering
	\begin{tabular}{|l|c|c|c|}
		\hline
		Rule type  & Representation & Score & Satisfaction condition \\
		\hline
		Constant   & $\phi_{i}(\mathbf{x})= x_i - \kappa_i = 0$ & \multicolumn{1}{l|}{$s_{\phi_{i}} =  \frac{|A_i|}{|\mathcal{X}|}$, where $A_i=\{1: \forall \mathbf{x}\in \mathcal{X}, |x_i - \kappa_i| \leq \rho_i \}, \kappa_i = \tilde{x}_i$}	& $ |x_{ir} - \kappa_i| \leq \rho_i $ \\
		Power law  & $\phi_{ij}(\mathbf{x}) = \hat{x}_i \hat{x}_j^b - c = 0$ 	& \multicolumn{1}{l|}{$s_{\phi_{ij}} = R^2$ score of linear regression shown in Equation \ref{eq:PowerLawRegression}}	& $\left(\hat{x}_{jr} - \left(\frac{c}{\hat{x}_{ir}}\right)^{\frac{1}{b}}\right)^2$ $\leq $ $e_{ij}^{\min}$ \\
		Equality   &  $\phi_{ij}(\mathbf{x}) = x_i - x_j = 0$ & \multicolumn{1}{l|}{$s_{\phi_{ij}} = \frac{|B_{ij}|}{|\mathcal{X}|}$, where $B_{ij}=\{1: \forall \mathbf{x}\in \mathcal{X}, |x_{i} - x_{j}| \leq \varepsilon_{ij}\}$}	& $|x_{ir} - x_{jr}| \leq \varepsilon_{ij}$ \\
		Inequality ($\leq$) & $\psi_{ij}(\mathbf{x}) = (x_i - x_j) \leq 0$ & \multicolumn{1}{l|}{$s_{\psi_{ij}} = \frac{|C_{ij}|}{|\mathcal{X}|}$, where $C_{ij}=\{1: \forall \mathbf{x}\in \mathcal{X}, x_{i} \leq x_{j}\}$}	& $(x_{ir} - x_{jr}) \leq 0 $\\
		Inequality ($\geq$) & $\psi_{ij}(\mathbf{x}) = (x_j - x_i) \leq 0$ & \multicolumn{1}{l|}{$s_{\psi_{ij}} = \frac{|D_{ij}|}{|\mathcal{X}|}$, where $D_{ij}=\{1: \forall \mathbf{x}\in \mathcal{X}, x_{i} \geq x_{j}\}$}	& $(x_{jr} - x_{ir}) \leq 0$ \\
		\hline
	\end{tabular}
	\label{tab:RuleType}
\end{table*}

\subsubsection{Constant rule}
In order to learn constant rules, we have to analyze the values of the variable under consideration for every ND solution and check if one or more of them converge to specific values. Since variables can be of different scales and units, we need a generalized criterion to determine if a variable is taking on a constant value. First, the median ($\tilde{x}_i$) is calculated. The proportion of ND solutions which satisfy $|x_i - \tilde{x}_i| \leq \rho_i$ is said to be the score ($s_{\phi_i}$) of the constant rule $x_i = \kappa_i = \tilde{x}_i$. $\rho_i$ is a small tolerance used for determining whether variable $x_i$'s value is in the neighborhood of $\tilde{x}_i$. It must be defined separately for each variable. An alternative option is to normalize the variables and define a singular $\rho$ for the normalized variable space. 
%
To check whether a new solution ($\mathbf{x}_r$) follows $x_{ir} = \kappa_i$, we check whether $x_{ir}$ lies in the neighborhood of $\kappa_i$ using the condition: $ |x_{ir} - \kappa_i| \leq \rho_i $.


\subsubsection{Power law rule}
In order to learn power laws ($x_i x_j^b = c$) we use the method proposed in \cite{Gaur2016} with a  modification. Each variable is initially normalized to [1, 2]. A training dataset is created from the ND solution set with the logarithms of normalized variables $\hat{x}_i$ and $\hat{x}_j$ as features, leading to Eqn.~\ref{eq:PowerLawRegression}: 
\begin{eqnarray}
\label{eq:PowerLawNormalized}
\hat{x}_i \hat{x}_j^b &=& c,\\
\Rightarrow \quad \log \hat{x}_i &=& \beta \log \hat{x}_j + \epsilon, \label{eq:PowerLawRegression}
\end{eqnarray}
where $\beta = -b$ is the weight and $\epsilon = \log c$ is the intercept. Normalization prevents 0 or negative values from appearing in the logarithm terms. Then we apply ordinary least squares linear regression to the logarithm of $\hat{x}_i$ and $\hat{x}_j$. Linear regression finds the best-fit line for the training data defined by the parameters $\beta$ and $\epsilon$. In order to evaluate the quality of the fit, we use the coefficient of determination ($R^2$) metric. A new solution ($\mathbf{x}_r$) follows the power law given in Equation \ref{eq:PowerLawNormalized} if the difference between the actual value ($x_{ir}$ or $x_{jr}$) and the predicted value ($\hat{x}_{ir}$ or $\hat{x}_{jr}$) is lower than a pre-defined threshold error ($e_{ij}^{\min}$). Table \ref{tab:RuleType} shows the formulation for the satisfaction condition. 

\subsubsection{Equality rule}
Two variables can be considered equal if $|x_i - x_j| \leq \varepsilon_{ij}$ with $\varepsilon_{ij}$ being a tolerance parameter for variable pair $x_i$ and $x_j$. The proportion of ND solutions following this condition is the score ($s_{\phi_{ij}}$) of the equality rule. The need to define $\varepsilon_{ij}$ for every variable pair can be avoided if normalized variables are used. 

\subsubsection{Inequality rule}
Inequality rules can be of the form $x_i \leq x_j$ or $x_i \geq x_j$. The proportion of ND solutions satisfying either condition is the score of the respective rules.

After the learning agent identifies specific rules from a set of ND solutions, the rules can be used to repair offspring solutions of the next generation. The repair mechanism for each rule is described next. 

\subsection{Repair agent}
\label{sec:RepairAgent}
Once the rules are learned from the current ND solutions by the learning agent, the next task is to use these rules to repair the offspring solutions for the next few generations. There are two questions to ponder. First, how many rules should we use in the repair process? Second, how closely should we adhere to each rule while repairing? A small fraction of learned rules may not embed requisite properties present in the ND solutions in offspring solutions. But the usage of too many rules may reduce the effect of each rule. Similarly, a tight adherence to observed rules may encourage premature convergence to a non-optimal solution, while a loose adherence may not pass on properties of ND solutions to the offspring. We propose four different rule usage schemes (10\% (RU1) to 100\% (RU4)) and three rule adherence schemes (RA1 (tight) to RA3 (loose)) for power law and inequality rules.   


\subsubsection{Constant rule}
To apply a constant rule $x_i = \kappa_i$ to a particular offspring solution $\boldx^{(k)}$, the variable $x_i^{(k)}$ is simply set to $\kappa_i$, thereby implementing the learned rule from previous ND solutions to the current offspring solutions. Constant rules are always included in the rule set and used with tight adherence. 

\subsubsection{Power law rule}
For a power law rule $\hat{x_i} \hat{x_j}^b = c$, one variable is selected as the base (independent) variable and the other variable is set according to the rule. For example, for a particular offspring solution $\boldx^{(k)}$, if $\hat{x_i}^{(k)}$ is selected as the base variable, $\hat{x}_{j}^{(k)}$ is set as follows: $\hat{x}_{j}^{(k)} = (\frac{c}{\hat{x}_i})^{\frac{1}{b}}$. Despite theoretically being able to represent constant relationships by having $b=0$, in practice, extremely low values of $b$ can cause the repaired variable $\hat{x}_{j}^{(k)}$ to have a large value outside the variable range. Hence, in this study, we first check whether a variable follows constant rules, and if it does, then that variable's involvement in a power law rule is ignored. 

A repair of a power law rule is followed with three different confidence levels by adjusting to an updated $c$-value: $\hat{x_i} \hat{x_j}^b = c_r$. PL-RA1 uses $c_r = c$ (tight adherence); PL-RA2 uses $c_r \in \mathcal{N}(c, \sigma_c)$ (medium adherence), and PL-RA3 uses $c_r \in \mathcal{N}(c, 2\sigma_c)$ (loose adherence), where $\sigma_c$ is the standard deviation of $c$-values for the power law observed among the ND solutions during learning process. 
PL-RA1 puts the greatest trust into the learned power law rule, whereas PL-RA3 has the least amount of trust and provides the most flexibility in the repair process. 

\subsubsection{Inequality and equality rules}
In order to repair an offspring solution $\boldx^{(k)}$, we have to select one variable ($x_i^{(k)}$) as the base variable and the other ($x_j^{(k)}$) as the dependent variable to be repaired. 
The generalized inequality repair operation is shown below:
\begin{align}
	\label{eq:ir}
	x_j^{(k)} &= x_i^{(k)} + \nu_{r1} (x_i^U - x_i^{(k)}), &\text{ for } x_i^{(k)} \leq x_j^{(k)},\\
	x_j^{(k)} &= \frac{x_i^{(k)} - \nu_{r2}x_i^U}{1-\nu_{r2}}, &\text{ for } x_i^{(k)} \geq x_j^{(k)}.
\end{align}
Three different rule adherence (RA) schemes are considered. IQ-RA1 uses $\nu_{r1} = \mu_{\nu1}$ and $\nu_{r2} = \mu_{\nu2}$ (tight adherence with no standard deviation), which are computed as the means of $\nu_1$ and $\nu_2$ from ND solutions during the learning process, as follows:
\[\nu_1 = \frac{x_j - x_i}{x_i^U - x_i}, \quad \nu_2 = \frac{x_i - x_j}{x_i^U - x_j}.\]  
For IQ-RA2, $\nu_{r1} \in \mathcal{N}(\mu_{\nu1}, \sigma_{\nu1})$ and $\nu_{r2} \in \mathcal{N}(\mu_{\nu2}, \sigma_{\nu2})$ (medium adherence with one standard deviation) are used, where $\sigma_{\nu_1}$ and $\sigma_{\nu_2}$ are standard deviations of $\nu_1$ and $\nu_2$, respectively. Both $\nu_{r1}$ and $\nu_{r2}$ are set to zero, if they come out to be negative. For IQ-RA3, $\nu_{r1}, \nu_{r2} \in U(0,1)$ (loose adherence with a uniform distribution) are used. 

\subsection{Ensemble repair agent}
\label{sec:moeai_es}
Both power law and inequality/equality rules have three rule adherence options for repair. For a new problem, it is not clear which option will work the best, so we also propose an ensemble approach (PL-RA-E and IQ-RA-E) in which all three options are allowed, but based on the success of each option, more probability is assigned to each. 
The ensemble method also considers a fourth option in which no repair to an offspring is made. 
The survival rate $(r_s^i)$ of offspring generated by the $i$-th repair operator is a measure of its quality. The greater the survival rate of the offspring created by an operator is, the higher is the probability of its being used in subsequent offspring generation. The probability ($\widehat{p^{i}_{r}}$) update operation for the $i$-th operator is presented below:
\begin{align}
\label{eq:p_ensemble}
p^{i}_{r}(t+1) &= \max\left(p_{\min}, \ \alpha \frac{r^i_{s}}{\sum_{i} r^i_{s}} + (1 - \alpha) \widehat{p^{i}_{r}}(t)\right)\!,\\
\label{eq:p_es_normal}
\widehat{p^{i}_{r}}(t+1) &= \frac{p^{i}_{r}(t+1)}{\underset{i}{\sum} p^{i}_{r}(t+1)},
\end{align}
where $\alpha$ is the learning rate, $r^i_{s} = \frac{n^i_{s}}{n_{\rm off}}$, where $n^i_{s}$ and $n_{\rm off}$ are the number of offspring created by the $i$-th operator that survive in generation $t$ and the total number of offspring that survive in generation $t$, respectively. 
It is possible that at any point during the optimization, no solution generated by one of the repair operators survives. This might cause the corresponding selection probability to go down to zero without any possibility of recovery. To prevent this, in Equation~\ref{eq:p_ensemble}, the probability update step ensures that a minimum selection probability ($p_{\min}$) is always assigned to each repair operator present in the ensemble. Equation~\ref{eq:p_es_normal} normalizes the probability values for each operator so that their total sum is one.

The learning rate ($\alpha$) determines the rate of change of the repair probabilities. A high $\alpha$ would increase the sensitivity, and can result in large changes in repair probabilities over a short period of time. A low $\alpha$ exerts a damping effect which causes the probability values to update slowly. Through trial and error, $\alpha = 0.5$ and $p_{\min} = 0.1$ are found to be suitable for the problems of this study.

\subsection{Mixed rule repair agent}
A mixed rule repair agent is designed to work on two or more different types of rules. Since multiple rules (for example, an inequality rule and a power law rule) can show up for the same variable pair, a rule hierarchy needs to exist as defined in Section~\ref{sec:ruleHierarchy}. Table~\ref{tab:RuleRanking} shows the rule hierarchical rank used for all the repair agents in this study.
\begin{table}[!hbt]
	\caption{Rule hierarchy by rank for each repair agent.}
	\centering
	\begin{tabular}{|c|c|c|}
		\hline
		Repair agent & Rule type  & Rank \\
		\hline
		\multirow{2}{*}{PL-RA1, PL-RA2, PL-RA3, PL-RA-E} & Constant 	& 1 \\
													& Power law & 2 \\
		\hline
		\multirow{4}{*}{IQ-RA1, IQ-RA2, IQ-RA3, IQ-RA-E} & Constant 	& 1 \\
													& Equality & 2 \\
													& Inequality ($\leq$) & 3 \\
													& Inequality ($\geq$) & 3 \\
		\hline
		\multirow{5}{*}{Mixed (Power law and inequality)} 	& Constant 	& 1 \\
															& Power law & 2 \\
														   	& Equality & 2 \\
														   	& Inequality ($\leq$) & 2 \\
														   	& Inequality ($\geq$) & 2 \\
		\hline
	\end{tabular}
	\label{tab:RuleRanking}
\end{table}

\subsection{User's ranking of rules}
The user forms the basis of the interactivity of the \IKEMO\ framework. At any point during the optimization, the user has the option to review the optimization results and provide feedback to the optimization algorithm in one or more of the following ways:
\begin{itemize}
	\item Rule ranking: The user may provide a ranking of rules (rank 1 is most preferred) provided by the algorithm. The algorithm will then try to implement the rules in the rank order provided by the user. 
	
	\item Rule exclusion: The user may select to remove certain rules provided by the algorithm, based on their knowledge of the problem. 
	
	\item Rule specificity: The user may specify details for considering a rule further.  For example, the user may specify that only variables having a correlation above a specified value should be considered. Another criterion could be to select all rules having a score greater than a threshold as rank 1 and exclude the others.
\end{itemize}
In this paper, the proposed rule usage schemes (RU1-RU4) can also be considered as artificial users \cite{Barba-Gonzalez2018} who select a certain percentage of the learned rules every few generations. This systematically illustrates the interactive ability of \IKEMO\ while showing the effect of different numbers of rules used for repair on the performance.

%

\subsection{Variable relation graph (VRG)}
The possible number of pair-wise relations among $n$ variables is $\frac{n(n - 1)}{2}$ or $O(n^2)$. Thus, for a large number of variables, the amount of bookkeeping required to track individual pairwise relations is large. Moreover, the observed relationships should not contradict each other. For example, for inequality rules $x_i \leq x_j$ and $x_j \leq x_k$, the transitive property can be maintained by choosing to repair $x_j$ based on $x_i$, followed by repairing $x_k$ based on $x_j$ using Equation \ref{eq:ir}. But repairing both $x_j$ and $x_k$ separately based on $x_i$ can potentially contradict the rule $x_j \leq x_k$. 
To solve these two challenges, we propose using a graph-based data structure, called variable relation graph (VRG), to encode and track relationships observed between multiple variable pairs. A customized graph-traversal algorithm ensures that all repairs are performed with minimal or no contradictions. In the following sections, steps 1 to 5 show the process of using learning agents to construct a VRG (learning phase). A learning interval ($T_{L}$) is defined as the number of generations or function evaluations (FEs) after which a new learning phase begins. Step 6 shows the process of applying the VRG to repair an offspring solution using one or more repair agents (repair phase). A repair interval ($T_R$) is defined as the number of generations or FEs between any two repair phases. 

\subsubsection{Create a complete VRG}
\label{sec:vrgstep1}
\begin{figure}[ht]
	\begin{subfigure}{0.48\columnwidth}
		\centering
		\includegraphics[height=0.9\columnwidth]{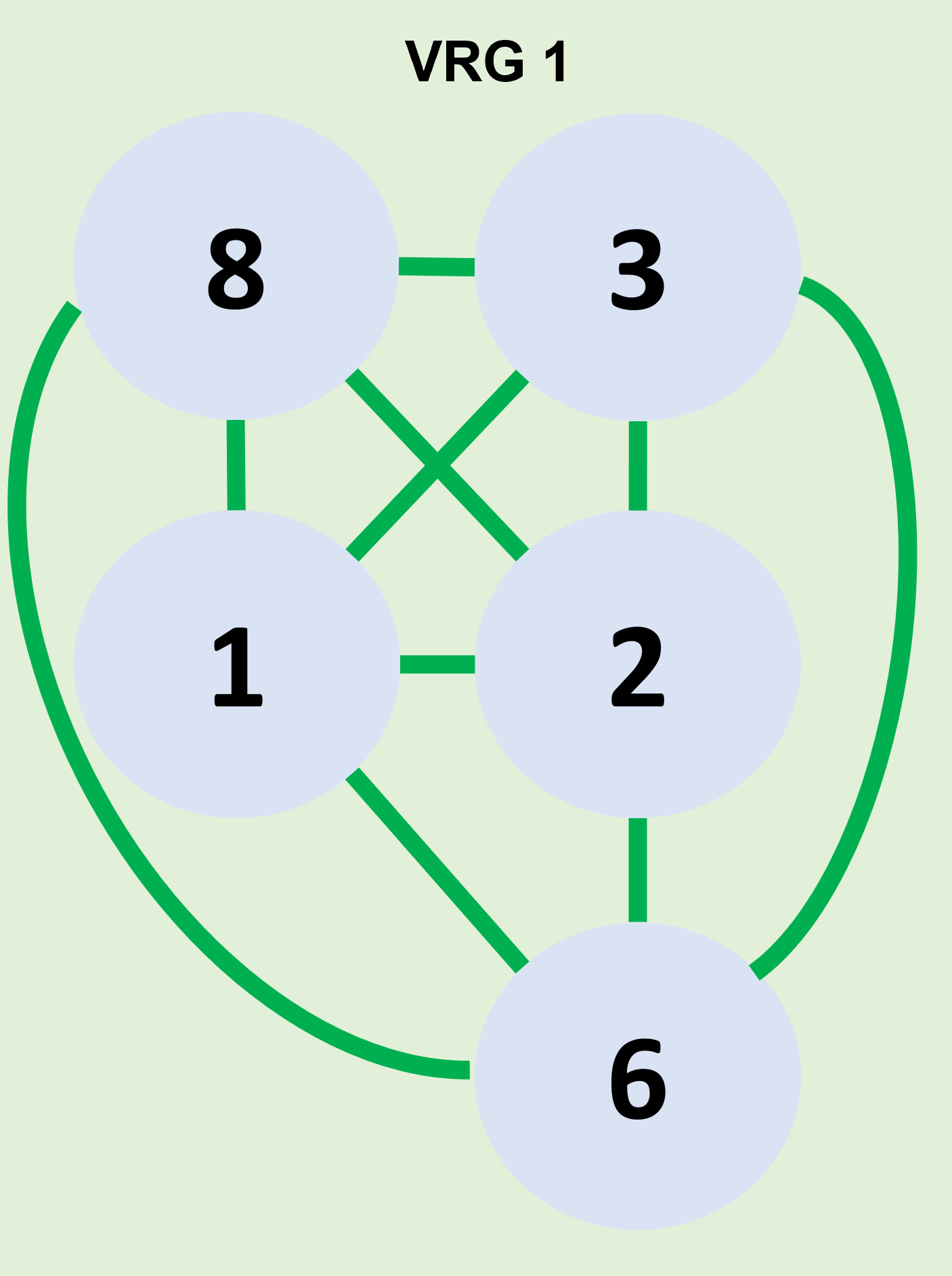}
		\caption{Group $G_1$.}
		\label{fig:vrg1step1}
	\end{subfigure}
	\begin{subfigure}{0.48\columnwidth}
		\centering
		\includegraphics[height=0.9\columnwidth]{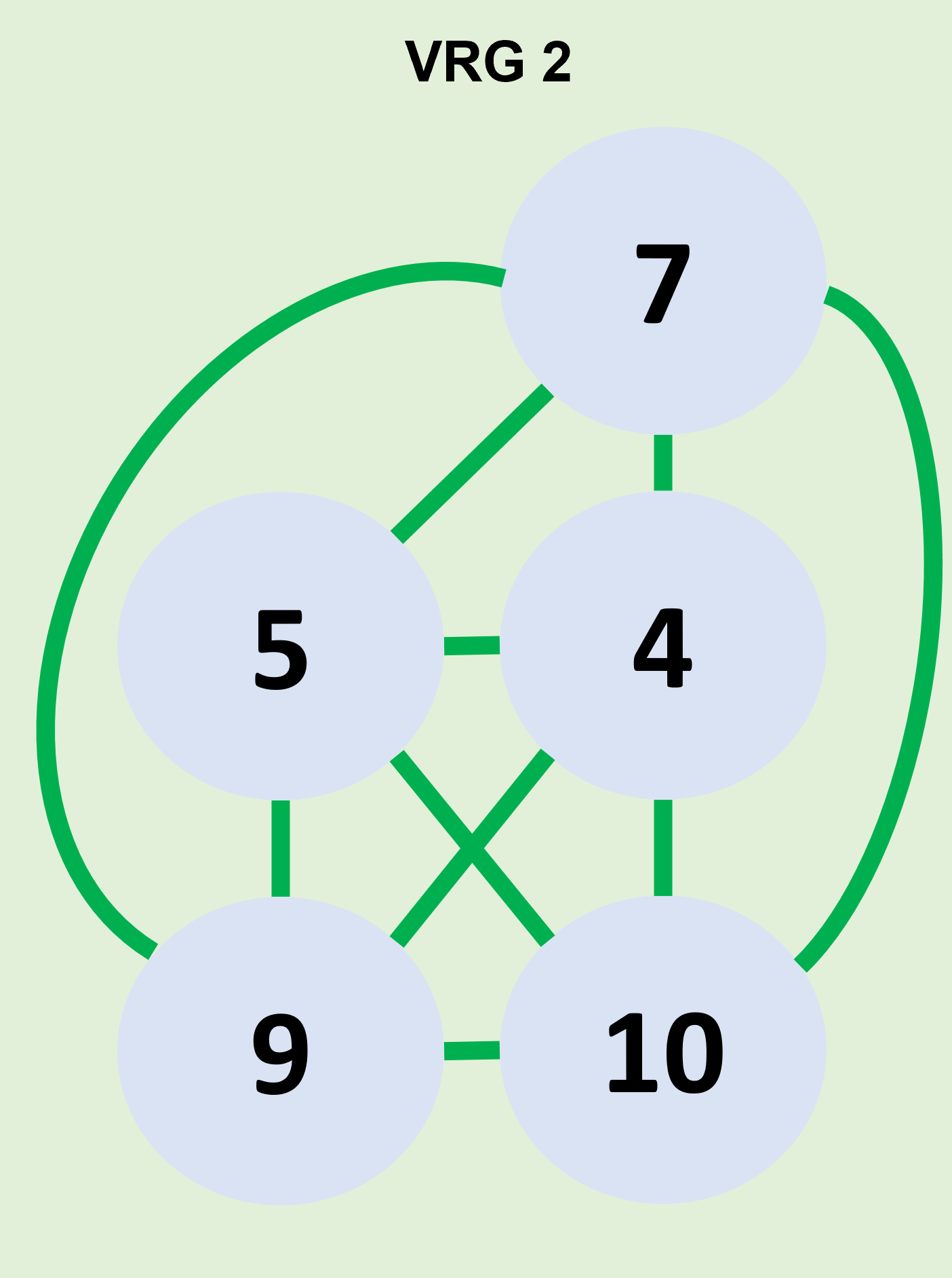}
		\caption{Group $G_2$.}
		\label{fig:vrg2step1plete}
	\end{subfigure}
	\caption{Ten variables in two non-interacting groups are represented in complete graphs.}
	\label{fig:exampleVRGstep1}
\end{figure}

A vertex (or node) of a VRG represents a variable and an edge connecting two nodes indicates the existence of a relationship between the corresponding variables. For every group $G_k$ of variables, all pairwise variable combinations are connected by an edge. This will result in a \textit{complete} graph where every pair of vertices is connected by a unique undirected edge. An example with two variable groups ($G_1 = \{1, 2, 3, 6, 8\}$ and $G_2 = \{4, 5, 7, 9, 10\}$) having five variables each is illustrated in Figure~\ref{fig:exampleVRGstep1}.

\subsubsection{Rule selection}
\begin{figure}[ht]
	\begin{subfigure}{0.48\columnwidth}
		\centering
		\includegraphics[height=0.9\columnwidth]{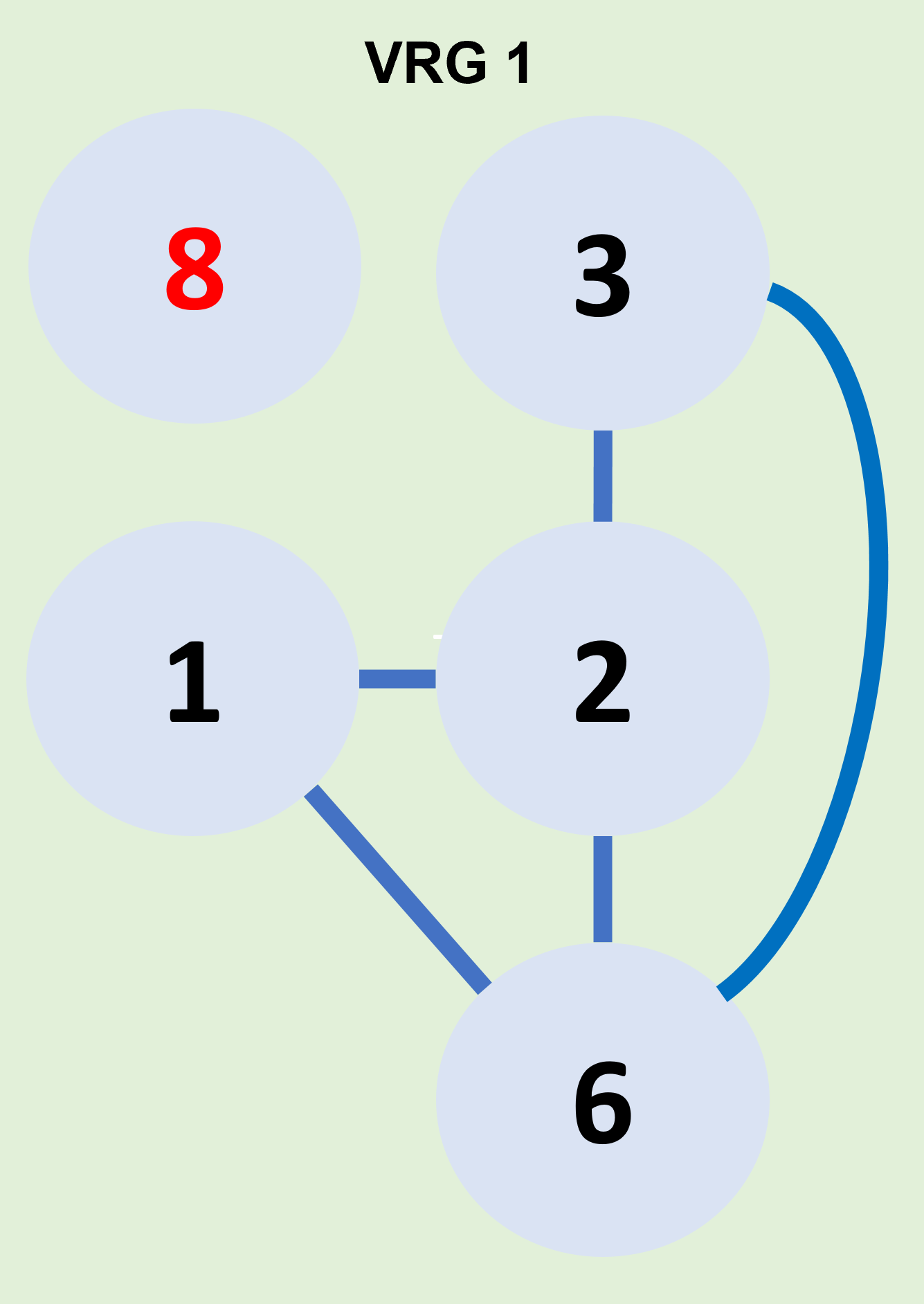}
		\caption{Group $G_1$.}
		\label{fig:vrg1step2}
	\end{subfigure}
	\begin{subfigure}{0.48\columnwidth}
		\centering
		\includegraphics[height=0.9\columnwidth]{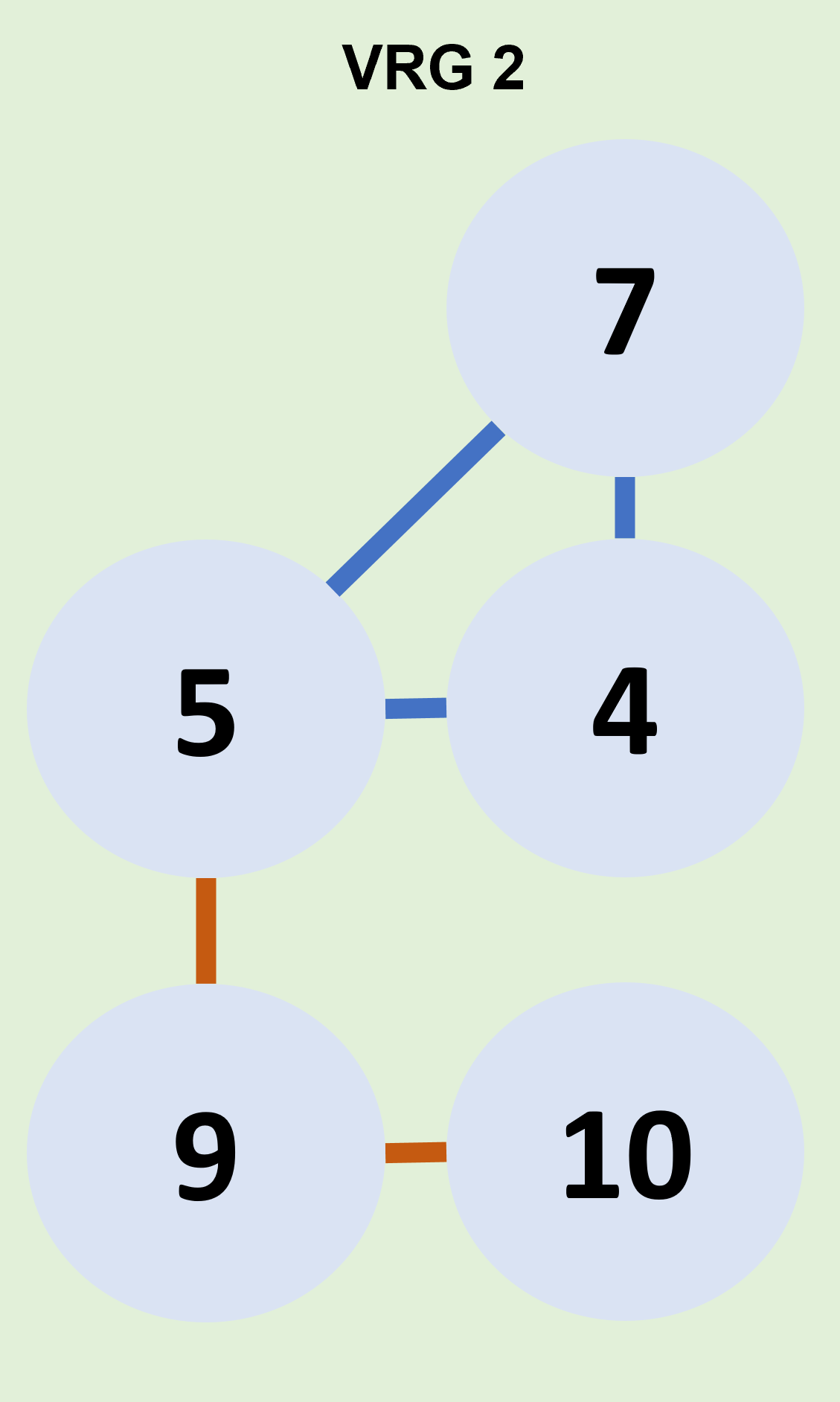}
		\caption{Group $G_2$.}
		\label{fig:vrg2step2}
	\end{subfigure}
	\caption{Rule selection.}
	\label{fig:exampleVRGstep2}
\end{figure}

In this step, learned rules are used to modify the VRGs according to two selection criteria. First, all rules having a score (defined in Table \ref{tab:RuleType}) above a certain threshold $(s_{\min})$ are considered. Second, they are applied in the order of user's preference ranking. A connection may be removed if it does not satisfy the selection criteria. 
If a single-variable (constant) rule satisfies the selection criterion, then the corresponding node is removed from the VRG and that rule will be implemented separately. 
If no two-variable rule involving $x_i$ and $x_j$ satisfies the minimum score criterion, the corresponding VRG edge ($i$-$j$) is removed. An example is shown in Figure~\ref{fig:exampleVRGstep2}, which uses the rule hierarchy for mixed rule repair operators (third row) shown in Table~\ref{tab:RuleRanking}, except that inequalities are ranked 3 for illustration here. A blue or brown edge represents a power law rule or an inequality rule, respectively. An edge ranking is also assigned based on the rule hierarchy. In this case, edges representing power laws and inequalities will be ranked 1 and 2 by default, unless overruled by the user. Both graphs have a reduced number of edges after the rule selection process is complete. Node~8 in Figure~\ref{fig:vrg1step2} (marked in red) is found to have a constant rule associated with it and hence removed. In Figure~\ref{fig:vrg2step2}, variables ($x_5$, $x_9$) and ($x_9$, $x_{10}$) are not related by power laws having a score greater than $s_{\min}$. However, they are found to follow inequality relationships with a score greater than $s_{\min}$. Hence, they are connected by brown edges. The rest of the edges represent power law rules and are marked by blue. The approach to set the direction of the edges is discussed next. 

\subsubsection{Create a directed acyclic VRG}
\begin{figure}[ht]
	\begin{subfigure}[b]{0.48\columnwidth}
		\centering
		\includegraphics[height=0.9\columnwidth]{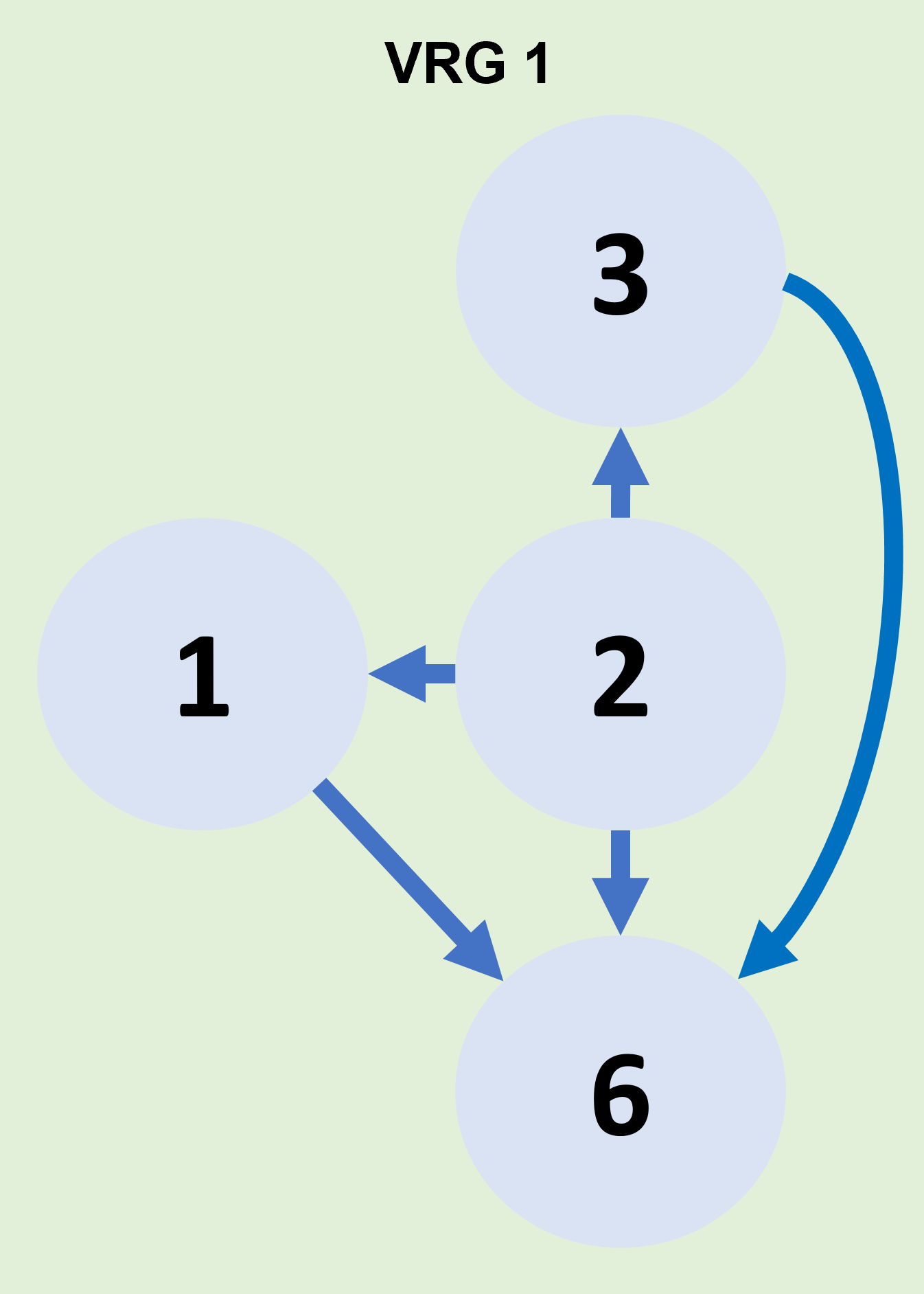}
		\caption{Group $G_1$.}
		\label{fig:vrg1step3}
	\end{subfigure}
	\begin{subfigure}[b]{0.48\columnwidth}
		\centering
		\includegraphics[height=0.9\columnwidth]{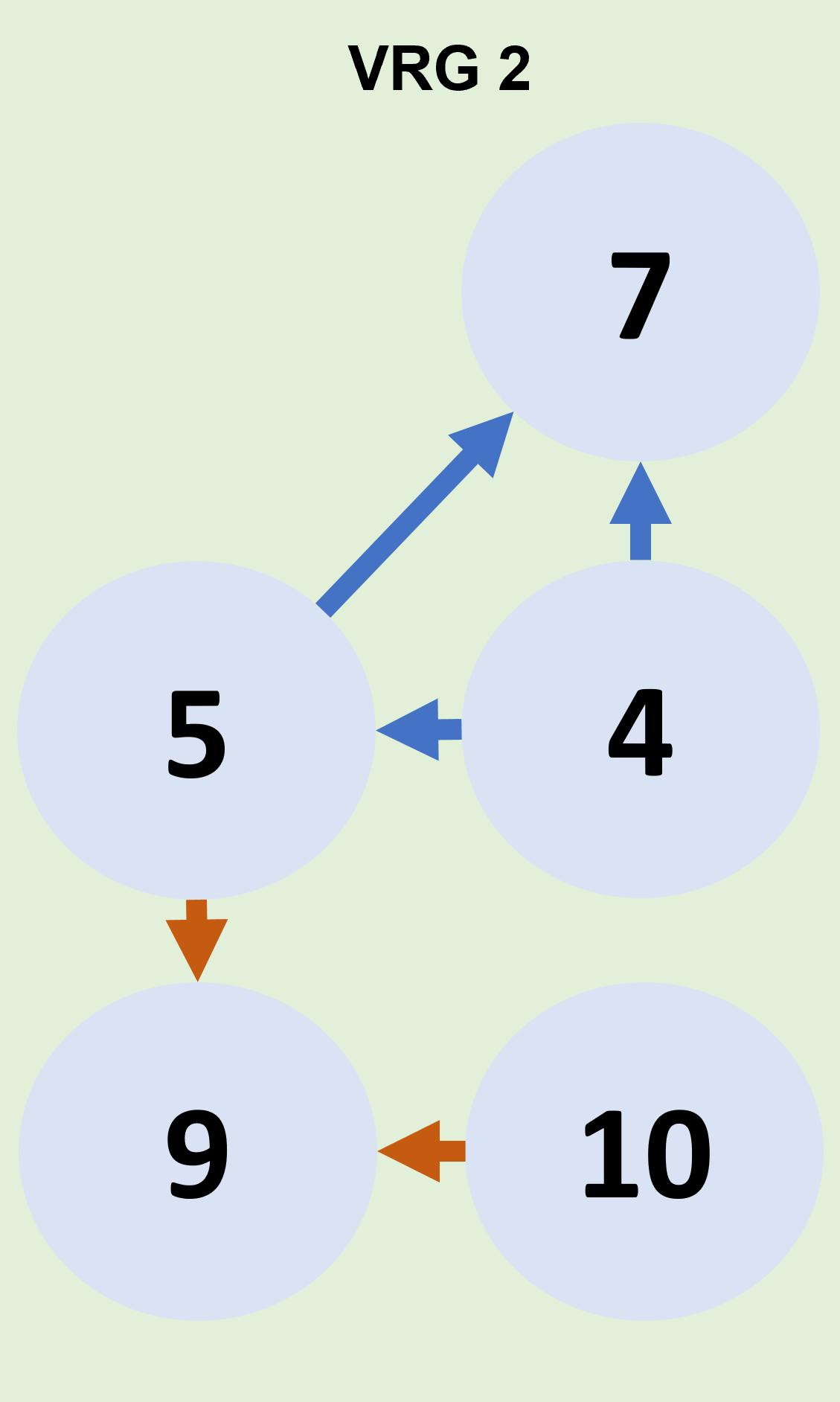}
		\caption{Group $G_2$.}
		\label{fig:vrg2step3}
	\end{subfigure}
	\caption{Creating a directed acyclic VRG.}
	\label{fig:exampleVRGstep3}
\end{figure}

In order to apply a repair agent to the VRG, it needs to be converted to a directed acyclic graph (DAG). This step ensures graph traversal is possible without getting stuck in loops. The members of every group $G_k$ are randomly permuted to create a sequence $D_k$. 
If $i$ appears before $j$ in $D_k$, an undirected edge between nodes $i$ and $j$ is converted to a directed edge from $i$ to $j$. In the example shown in Figure~\ref{fig:exampleVRGstep3}, two random sequences 
$D_1 = (2, 1, 3, 6)$ and $D_2 = (10, 4, 5, 9, 7)$ are created for groups $G_1$ and $G_2$, respectively. Since node~2 appears before node~1 in $D_1$, a blue arrow goes from node~2 to node~1, as shown in the figure. 
This process is repeated for every population member so as to create diverse VRGs. 

\subsubsection{Transitive reduction}
\begin{figure}[ht]
	\begin{subfigure}{0.48\columnwidth}
		\centering
		\includegraphics[height=0.9\columnwidth]{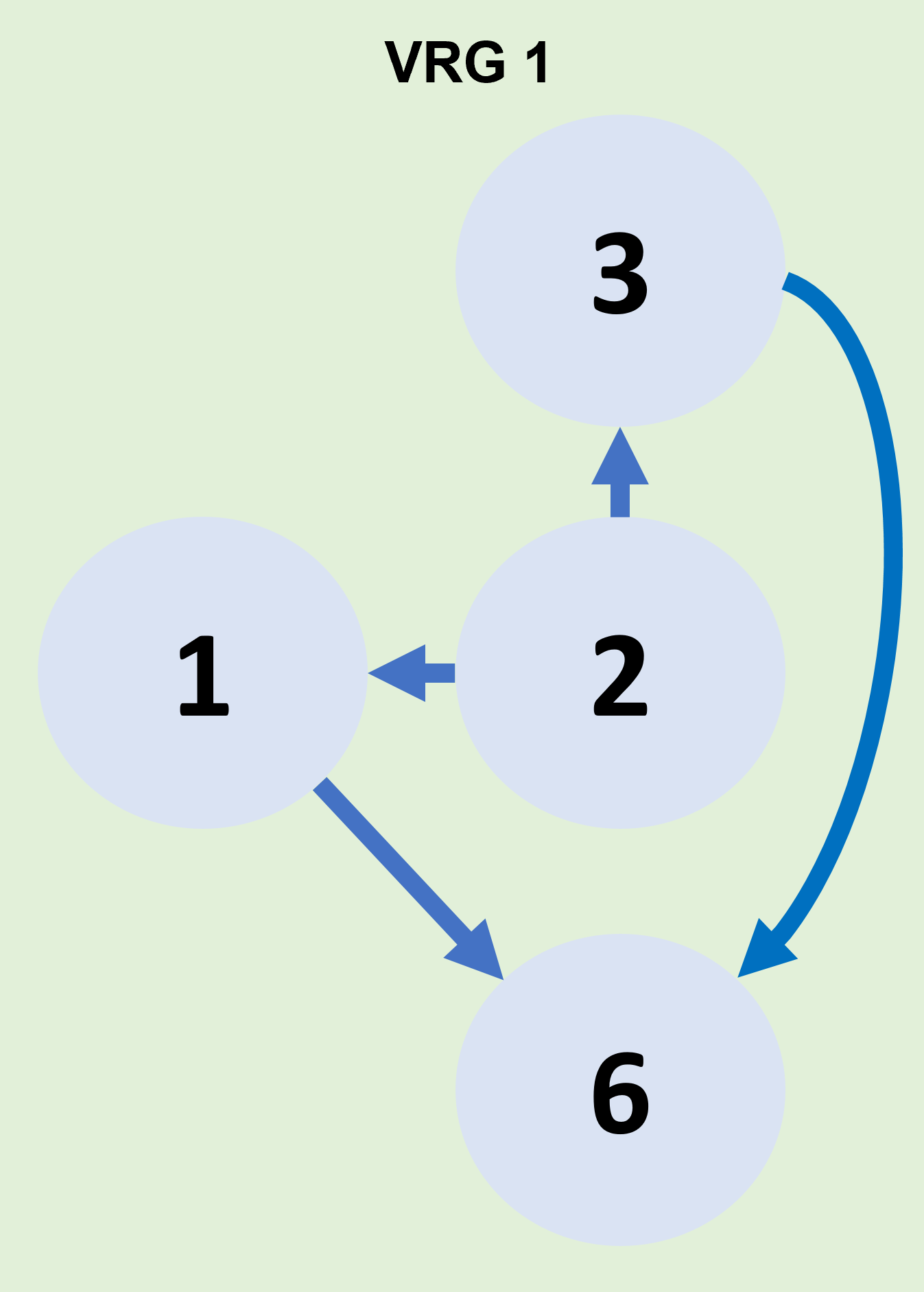}
		\caption{Group $G_1$.}
		\label{fig:vrg1step4}
	\end{subfigure}
	\begin{subfigure}{0.48\columnwidth}
		\centering
		\includegraphics[height=0.9\columnwidth]{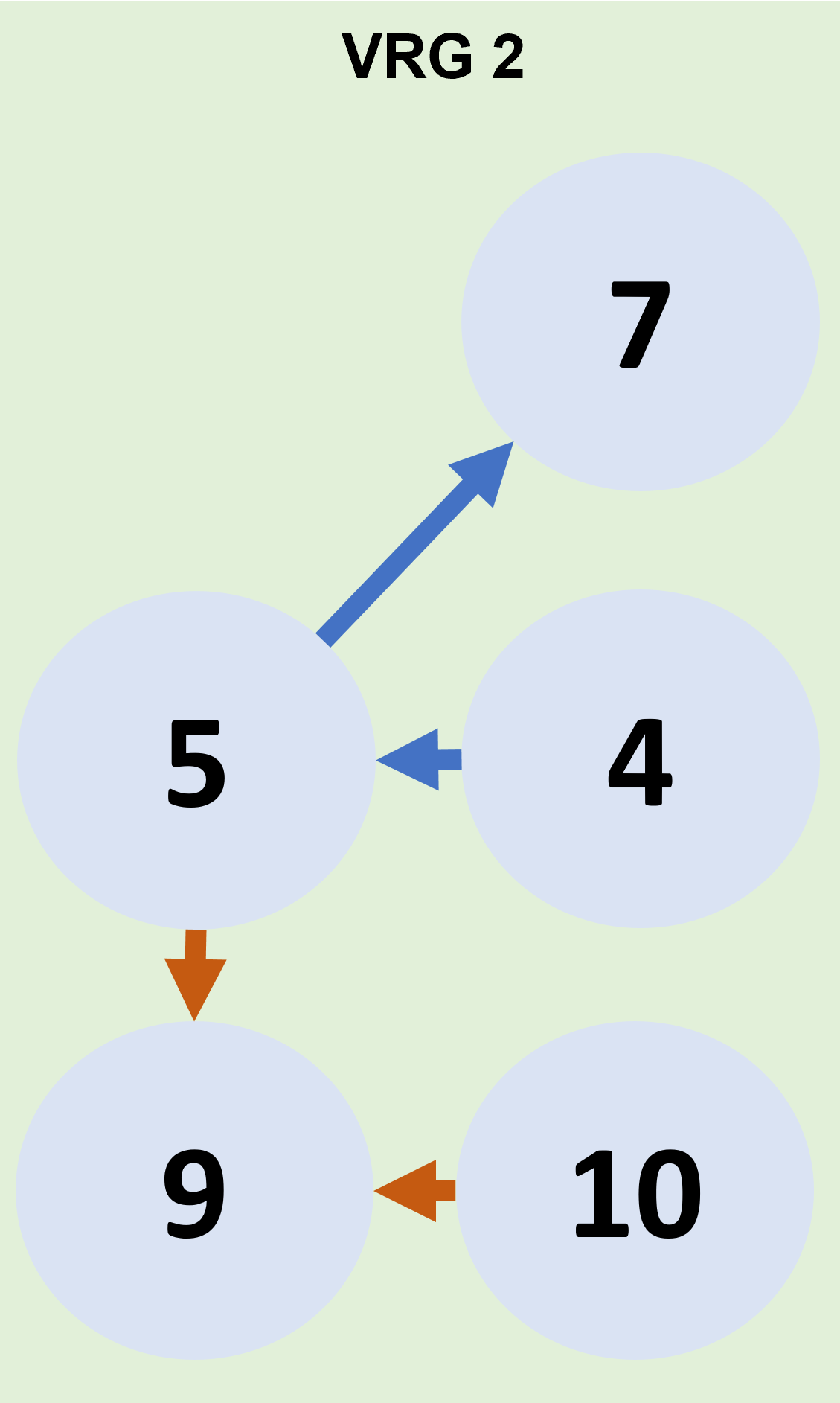}
		\caption{Group $G_2$.}
		\label{fig:vrg2step4}
	\end{subfigure}
	\caption{Transitive reduction.}
	\label{fig:exampleVRGstep4}
\end{figure}

Next, a transitive reduction \cite{Aho2006} is performed on the VRG corresponding to each variable group. For VRGs having both power law and inequality edges, transitive reduction is performed on subgraphs consisting only of the edges of the same type. This step eliminates redundant directed edges between two different rule types. An example of eliminating an arrow from node~2 to node~6 is shown in Figure~\ref{fig:vrg1step4}.
%

\subsubsection{Modify VRG according to user's feedback}
\label{sec:vrgstep5}
\begin{figure}[ht]
	\begin{subfigure}{0.48\columnwidth}
		\centering
		\includegraphics[height=0.9\columnwidth]{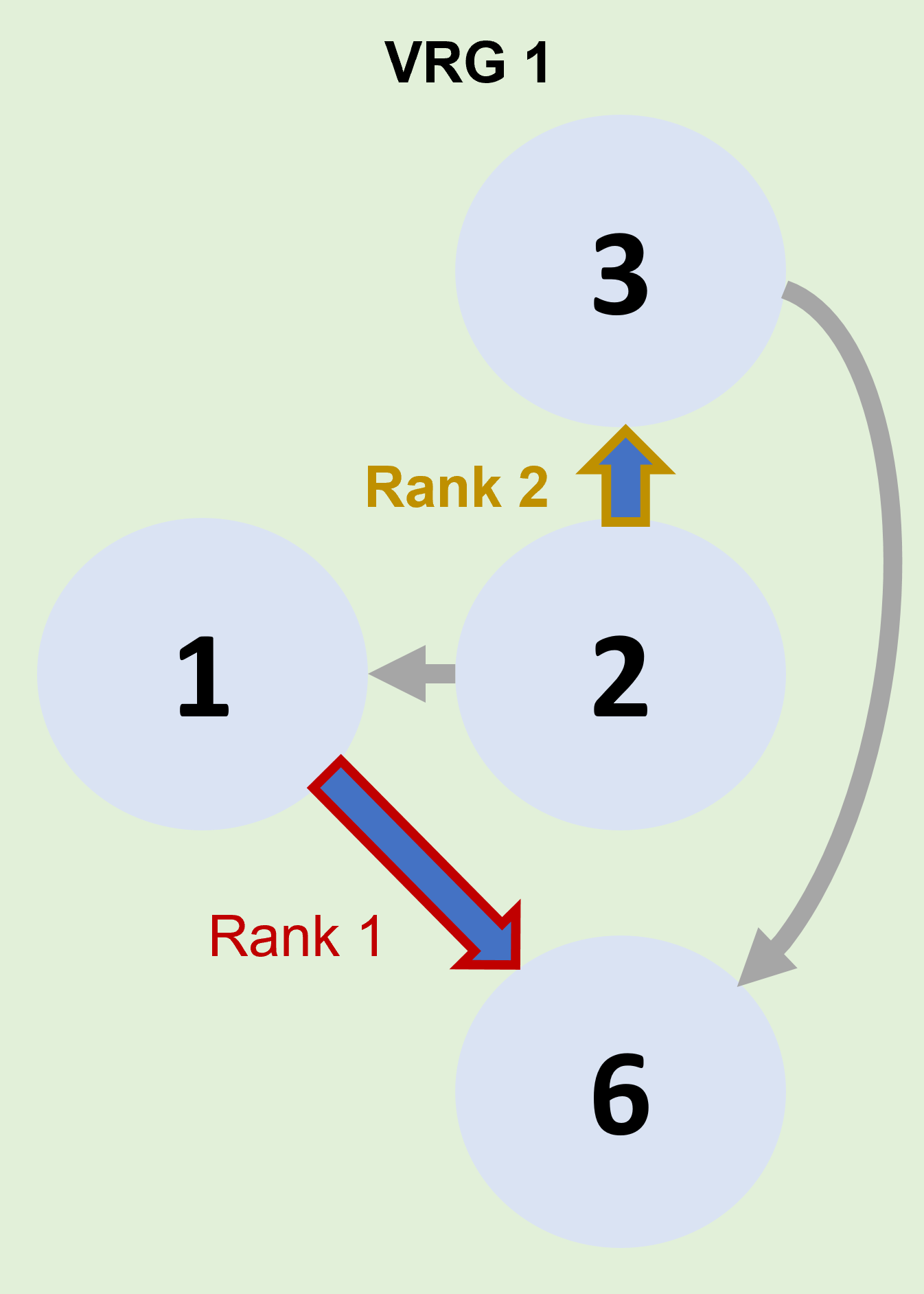}
		\caption{Group $G_1$.}
		\label{fig:vrg1step5}
	\end{subfigure}
	\begin{subfigure}{0.48\columnwidth}
		\centering
		\includegraphics[height=0.9\columnwidth]{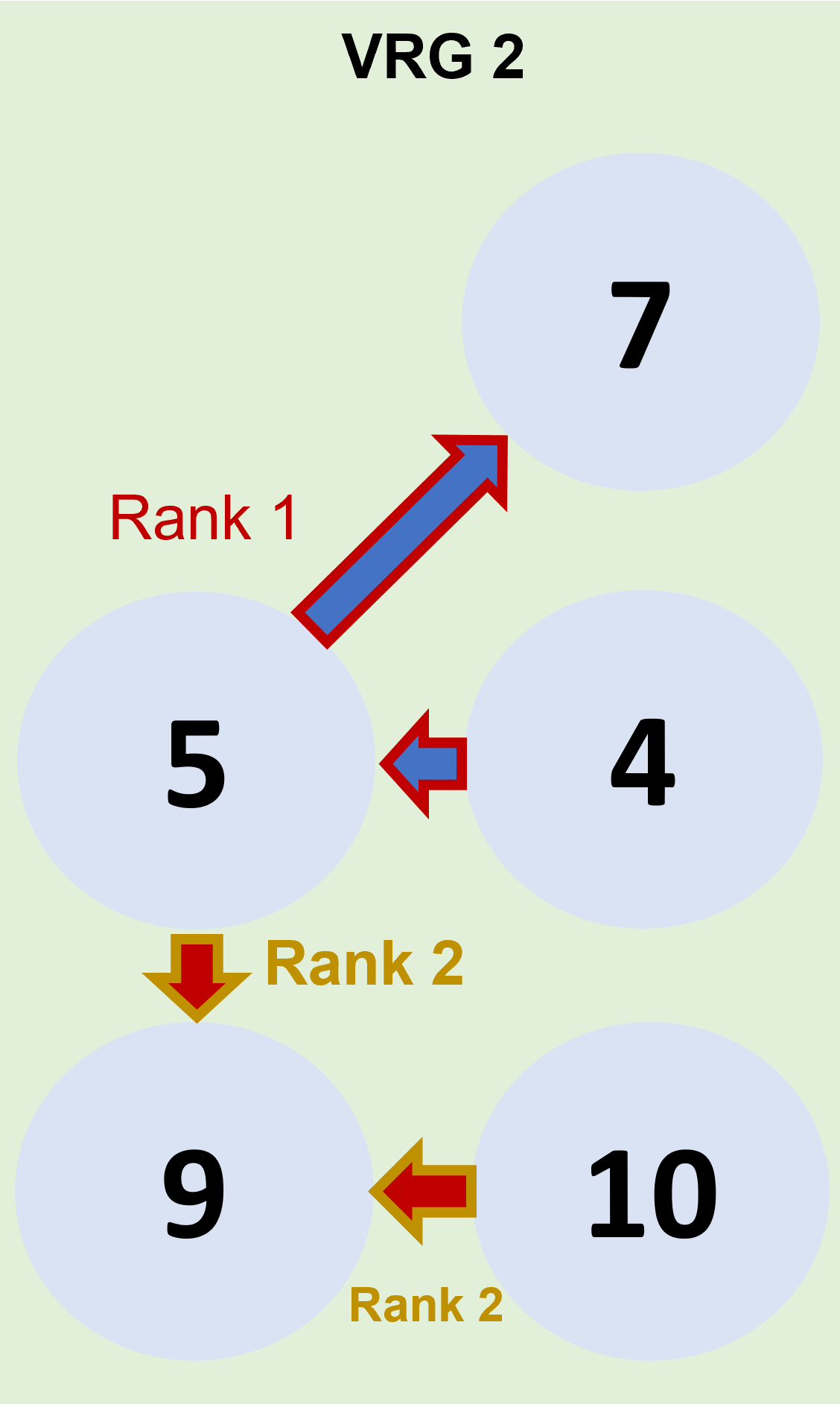}
		\caption{Group $G_2$.}
		\label{fig:vrg2step5}
	\end{subfigure}
	\caption{Implementing user feedback.}
	\label{fig:exampleVRGstep5}
\end{figure}

A user can provide feedback in the form of a ranking, or select only a subset of the available rules. In the former case, the VRG edge rankings are updated to reflect the user's choice. Edges corresponding to the rules discarded by the user are removed. Figure~\ref{fig:vrg1step5} shows an example where the rule involving $x_1$ and $x_6$ are ranked 1 (marked by arrows with a red border) and $x_2$ and $x_3$ are ranked 2 (marked by arrows with a dark yellow border). The gray edges represent the rules discarded by the user. Figure~\ref{fig:vrg2step5} shows a similar ranking process.

\subsubsection{Repair new offspring solutions}
\label{sec:vrgstep6}
For every new solution, the corresponding VRGs are traversed. A random rank 1 starting point is selected and the VRG is traversed recursively in a depth-first fashion. From every node, the algorithm first moves forward via the outgoing edges and repairs the connected node based on the current node. Once all outgoing edges are traversed, and the algorithm comes back to the same node, traversal is performed by following the incoming edges. This is repeated for all ranks. Algorithm~\ref{alg:VRGrepair} presents the pseudocode of the repair process. For ease of understanding, some of the terminology used in the pseudocode is explained in this section. In the pseudocode, the VRG data structure has the attributes Nodes and Edges. The Edges attribute representing an edge $(i, j)$ has multiple sub-attributes: StartVertex ($i$ in this case), EndVertex ($j$ in this case), EdgeType (rule type and correspdonding repair agent), EdgeRank (rank of an edge). A function \textit{TraverseGraph} is used which recursively traverses the VRG from a random start node for a particular rule rank. The function \textit{Repair} called by \textit{TraverseGraph} calls the correct repair agent based on EdgeType.
{\small\begin{algorithm}[hbt]
	\caption{VRG traversal and repair pseudocode.}
	\label{alg:VRGrepair}
	{\small \begin{algorithmic}[1]
			\Require{New solution set $(\mathcal{X}_r)$, variable groups $({G})$, VRGs for every solution and group, rule hierarchy.}
			\Ensure{Repaired solution set $\mathcal{X}_r$}.
			\Function{TraverseGraph}{x, Graph, CurrentNode, PreviousNode, NodesVisited, CurrentRank}
			\If{CurrentNode in NodesVisited}
				\State \Return
			\EndIf
			\State CurrentEdges $\gets$ Graph.Edges[CurrentNode];
			\For{each outgoing edge (e) in CurrentEdges}
				\State NextNode $\gets$ e.EndVertex;
				\If{NextNode not in NodesVisited}
					\State EdgeType $\gets$ e.EdgeType;
					\State EdgeRank $\gets$ e.EdgeRank;
					\If{EdgeRank = CurrentRank}
						\State Repair(x, CurrentNode, NextNode, EdgeType, EdgeRank);
					\EndIf
					\State TraverseGraph(x, Graph, NextNode, CurrentNode, NodesVisited);
				\EndIf
			\EndFor
			\For{each incoming edge (e) in CurrentEdges}
				\State NextNode $\gets$ e.StartVertex;
				\If{NextNode not in NodesVisited \textbf{and} NextNode $\neq$ PreviousNode}
					\State EdgeType $\gets$ e.EdgeType;
					\State EdgeRank $\gets$ e.EdgeRank;
					\If{EdgeRank = CurrentRank}
						\State Repair(x, CurrentNode, NextNode, EdgeType, EdgeRank);
					\EndIf
					\State TraverseGraph(x, Graph, NextNode, CurrentNode, NodesVisited);
				\EndIf
			\EndFor
			\State Add CurrentNode to NodesVisited;
			\EndFunction
			
			\For{each group $G_k$ in $G$} \Comment Repair procedure begins
				\For{each solution $x$ in $\mathcal{X}_r$}
					\State CurrentGraph $\gets$ VRG assigned to $\mathbf{x}$ for $G_k$;
					\For{CurrentRank = 1, 2, ..., $n_{ranks}$}
					\State StartNode $\gets$ Select random node having atleast one edge of rank CurrentRank;
					\State TraverseGraph(x, CurrentGraph, StartNode, NULL, [], CurrentRank);
					\EndFor
				\EndFor
			\EndFor
	\end{algorithmic}}
\end{algorithm}}

\section{Simply-supported stepped beam design}
\label{sec:SteppedBeam}
\begin{figure}[htb]
	\centering
	\includegraphics[width=1.05\linewidth]{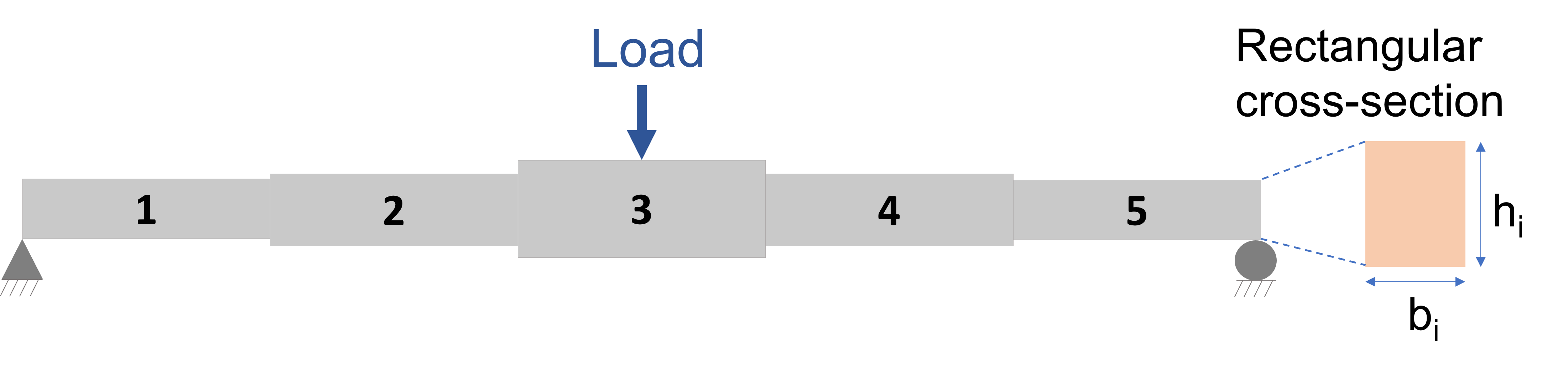}
	\caption{Simply-supported stepped beam with five segments.}
	\label{fig:SteppedBeam}
\end{figure}

Beam design problems are common in the literature \cite{Gandomi2019a,Rothwell2017} and can be used to benchmark an optimization algorithm. In this paper, we consider a simply-supported stepped beam design with multiple segments having a rectangular cross-section. An example with five segments is shown in Figure~\ref{fig:SteppedBeam}.
A vertical load of 2 kN is applied at the middle of the beam. All $n_{\rm seg}$ segments are of equal length. The area of the rectangular cross-section is determined by its width ($b_i$) and height ($h_i$) for the $i$-th segment, where $i \in [1, n_{\rm seg}]$, The volume ($V$) and maximum deflection ($\Delta$) are to be minimized by finding an optimal width $b_i$ and height $h_i$ of each segment, totalling $2n_{\rm seg}$ variables. The maximum stress $\sigma_i(\boldx)$ of $i$-th member and deflection $\delta_j(\boldx)$ at $j$-th node need to be kept below strength of the material $\sigma_{\max}$ and a specified limit $\delta_{\max}$, respectively. The aspect ratio (ratio of height to width) of each segment is also restricted within a particular range (in $[a_L, a_U]$), as constraints. The MOP formulation is shown below:
\begin{align}
\label{eq:SteppedBeam}
\text{Minimize } & V(\boldx) = \sum_{i=1}^{n_{\rm seg}} b_i h_i l_i,\\
\text{Minimize } & \Delta(\boldx) = \max_{i=1}^{n_{\rm seg}} \delta_i(\boldx), \\
\text{Subject to } & \max_{i=1}^{n_{\rm seg}} \sigma_i(\boldx) \leq \sigma_{\max},\\
& \max_{j=1}^{n_{\rm seg}} \delta_j(\boldx) \leq \delta_{\max}, \\
&a_L \leq a_i \leq a_U,\ \mbox{for $i=1,\ldots,n_{\rm seg}$.}
\end{align}
Here, two cases with 39 and 59 segments are considered. Problem parameters are described in Table~\ref{tab:SteppedBeamDetails}.
\begin{table*}[hbt]
	\centering
	\caption{Number of decision variables and constraints for 39 and 59-segment stepped beams.}
	{\small\begin{tabular}{|c|c|c|c|c|c|c|c|}
			\hline
			$n_{seg}$ & $\sigma_{\max} (MPa)$ & $\delta_{\max} (m)$ & Width ($b_i$)  & Height ($h_i$)  & Aspect ratio ($ a_i $)  & Variables & Constraints \\
			\hline
			39 & 20 & 0.04 & [0.1, 40] & [0.1, 40] & [0.5, 2] & 78 & 41 \\
			
			59 & 20 & 0.06 & [0.1, 60] & [0.1, 60] & [0.5, 2] & 118 & 61\\
			\hline
	\end{tabular}}
	\label{tab:SteppedBeamDetails}
\end{table*}

\subsection{Experimental settings}
NSGA-II \cite{Deb2002}, a state-of-the-art MOEA, is applied with the proposed \IKEMO\ procedure to solve both cases. 
This problem is intended to demonstrate the performance of our proposed algorithm with minimal initial user knowledge. Thus, no grouping information is provided, resulting in all variables being in a single group. \IKEMO\ is combined separately with each repair agent described in Section \ref{sec:RepairAgent}. In addition, there are two cases where mixed relationships are used: the first case with PL-RA2 and IQ-RA2, and the second case with PL-RA-E and I-ES. The rule hierarchy is described in Table~\ref{tab:RuleRanking}. 

Four rule usage schemes RU1, RU2, RU3 and RU4 select the top 10\%, 20\%, 50\% and 100\% of the learned rules sorted according to their scores. They also act as artificial users with a consistent behavior.  Each rule usage scheme is paired with one or more repair agents. Eight cases with a single repair agent are considered: PL-RA1, PL-RA2, PL-RA3, PL-RA-E, IQ-RA1, IQ-RA2, IQ-RA3, IQ-RA-E. Two cases with a combination of repair agents are considered: one with PL-RA2 and IQ-RA2, and the other with PL-RA-E and IQ-RA-E. From Table \ref{tab:RuleType}, $\rho_i$ is set to be 0.1, $\varepsilon_{ij}$ is set as 0.1, and $e_{ij}^{min}$ is set to 0.01. Table~\ref{tab:NSGA2Param} shows the parameter settings for this problem.

For each combination of a repair agent and user, 20 runs are performed and the Hypervolume (HV) \cite{Zitzler1998} values are recorded at the end of each generation. 
The Wilcoxon rank-sum test \cite{Hollander2015a} is used to compare the statistical performance of the algorithms tested here with respect to the best performing algorithm for each scenario. As an example, let $p_1$ and $p_2$ represent the performance metric values for two algorithms $A_1$ and $A_2$. For each simulation run, $p_1$ and $p_2$ exist as paired observations. Here, the null hypothesis states that there is no statistically significant difference between $p_1$ and $p_2$. The hypothesis is tested with 95\% significance level and the $p$-values are recorded. A $p$-value less than $0.05$ means that there is a statistically significant performance difference between $A_1$ and $A_2$. The median number of FEs taken to achieve a target hypervolume (HV$^T$) is used as a performance metric for the Wilcoxon test. HV$^T$ is set to be 80\% of the highest median HV among all the algorithms. In order to make the problem challenging for the proposed approach, a small population size of 40 is used and a maximum number of generations of 500 is set, thereby allowing a maximum computational budget of 20,000 FEs for each run. 
\begin{table}[hbt]
	\centering
	\renewcommand{\arraystretch}{1.1}
	\caption{Parameter settings of \IKEMO.}
	\label{tab:NSGA2Param}
	{
		\begin{tabular}{|l|c|} \hline
			Parameter & Value \\ \hline
			Population size & Problem-specific \\
			Maximum generations & Problem-specific \\
			Mutation operator & Polynomial mutation \cite{deb-book-01}\\
			Mutation probability ($p_m$) and index ($\eta_m$) & $1/n_{var}$, {50} \\
			Crossover operator & SBX \cite{Deb1995}\\
			Crossover probability ($p_c$) and index ($\eta_c$) & 0.9, 30 \\
			Minimum rule score, $s_{\min}$ & 0.7 \\
			Learning interval ($T_{L}$, in generations) & 10\\
			Repair interval ($T_R$, in generations) & 10\\
			$\alpha$ and $p_{\min}$ in Equation \ref{eq:p_ensemble} & 0.5, 0.1\\
			Rule parameters $\rho_i$, $\varepsilon_{ij}$,  $e_{ij}^{\min}$ &  Problem-specific\\
			User feedback lag, $T_U$ in Section~\ref{sec:syncAsync} & User-dependent\\
			\hline
	\end{tabular}}
\end{table}

\subsection{Experimental results and discussion}
\begin{figure*}[htb]
\begin{center}
    \begin{subfigure}{0.48\textwidth}
	\centering\includegraphics[width=0.8\columnwidth]{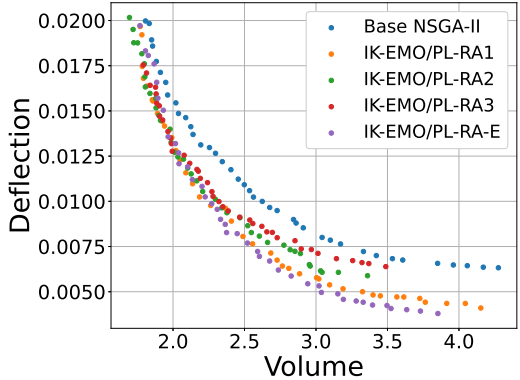}
		\caption{ND Front for one run.}
		\label{fig:pfSB59}
	\end{subfigure}\hfill
	\begin{subfigure}{0.48\textwidth}
	\centering\includegraphics[width=0.8\columnwidth]{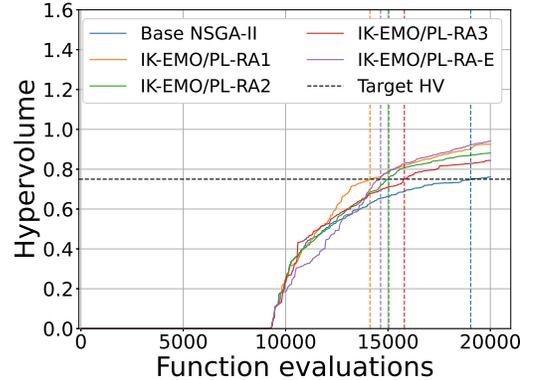}
		\caption{HV plot over 20 runs.}
		\label{fig:hvSB59}
	\end{subfigure}
	\end{center}
	\caption{ND fronts and hypervolume plots obtained by \IKEMO\ with RU2 and power law repair agents for 
	59-segment stepped beam problem.}
	\label{fig:SBResults}
\end{figure*}

Tables~\ref{tab:SteppedBeamResult39} and \ref{tab:SteppedBeamResult59} show the optimization results for the 39 and 59-segment stepped beam problems, respectively. Base NSGA-II results without any rule extraction and repair are shown in the first row. The best performance case in each row is marked in bold. For every column, the best performing algorithm is marked with a shaded gray box. The Wilcoxon p-values show the relative performance of each algorithm with the column-wise best performance. Algorithms with a statistically similar performance to the column-wise best are shown in italics. The ND front obtained in a particular run using the power law repair operators for RU2 are shown in 
Figure~\ref{fig:pfSB59} for the 59-segment case. The corresponding median HV plot over the course of the optimization run are shown in 
Figure~\ref{fig:hvSB59}. Similar behaviors are observed for 39-segment case (see supplementary materials).
\begin{table*}[hbt]
	\centering
	\caption{FEs required to achieve HV$\protect^T$ = 0.81 for 39-segment beams. Best performing algorithm for row is marked in bold. Best performing algorithm in each column is marked by a shaded gray box. Algorithms with statistically similar performance to the best algorithm column-wise are marked in italics. The corresponding Wilcoxon p-values are given in braces.}
	\begin{tabular}{|c|c|c|c|c|c|}
		\hline
		Rule Type & Repair agent & RU1 & RU2 & RU3 & RU4 \\
		\hline\hline
		None 						& None (base) & 10.6k $\pm$ 1.0k (p = 0.0145) & 10.6k $\pm$ 1.0k (p=0.0118) & 10.6k $\pm$ 1.0k (p=0.0238) & 10.6k $\pm$ 1.0k (p=0.0412)\\
		\hline
		\hline
		\multirow{4}{1.2cm}{Power law rule}  & PL-RA1 & 10.4k $\pm$ 0.8k (p=0.0416) & \textbf{10.3k $\pm$ 1.0k} (p=0.0225) & 10.5k $\pm$ 0.9k (p=0.0420) & 10.6k $\pm$ 1.0k (p=0.0319)\\
									& PL-RA2 & \textit{9.8k  $\pm$ 0.8k} (p=0.0661) & \cellcolor{gray!25}\textbf{9.4k $\pm$ 0.9k}& \cellcolor{gray!25}9.5k $\pm$ 1.2k & \textit{10.2k  $\pm$ 1.3k} (p=0.0551)\\
									& PL-RA3 & 10.4k $\pm$ 0.9k (p=0.0195) & 10.3k $\pm$ 1.0k (p=0.0422) & \textbf{9.8k $\pm$ 1.0k} (p=0.0106)  & 11.6k $\pm$ 1.0k (p=0.0147) \\
									& PL-RA-E & \cellcolor{gray!25}9.6k  $\pm$ 1.0k & \textbf{\textit{9.5k $\pm$ 1.0k}} (p=0.0762)  & \textit{9.6k $\pm$ 1.4k} (p=0.0841)  & \cellcolor{gray!25}9.9k $\pm$ 1.2k\\
		\hline
		\hline
		\multirow{4}{1.2cm}{Inequality rule} & IQ-RA1 & 10.7k $\pm$ 0.7k (p=0.0016) & \textbf{10.5k $\pm$ 1.0k} (p=0.0471) & 10.6k $\pm$ 0.9k(p=0.0483)  & 10.8k $\pm$ 1.0k (p=0.0308) \\
										& IQ-RA2 & \textbf{10.2k $\pm$ 0.9k (p=0.0125)} & 10.3k $\pm$ 1.1k (p=0.0263)  & 10.3k $\pm$ 0.8k(p=0.0486)  & 10.7k $\pm$ 1.2k (p=0.0210) \\
										& IQ-RA3 & 10.7k $\pm$ 1.2k (p=0.0483) & 10.6k $\pm$ 1.1k(p=0.0340)  & \textbf{10.5k $\pm$ 1.0k} (p=0.0318)  & 10.6k $\pm$ 1.4k (p=0.0247) \\
										& IQ-RA-E & 10.6k $\pm$ 0.8k (p=0.0463) & \textbf{\textbf{10.5k $\pm$ 0.9k}} (p=0.0207)  & 10.6k $\pm$ 1.3k(p=0.0342)  & 10.6k $\pm$ 1.6k (p=0.0177) \\
		\hline
		\hline
		\multirow{2}{1.2cm}{Mixed rule} & PL-RA2+IQ-RA2 & \textit{9.8k $\pm$ 0.9k}  (p=0.0517) & \textbf{\textit{9.6k $\pm$ 1.0k}} (p=0.0957)   &  \textit{9.7k $\pm$ 1.1k} (p=0.0586)   &  \textit{10.4k $\pm$ 0.8k} (p=0.0778)  \\
								& PL-RA-E+IQ-RA-E &  \textbf{\textit{9.7k $\pm$ 1.2k}} (p=0.0913)   &  \textit{9.6k $\pm$ 0.4k} (p=0.0713)   &  \textit{9.8k $\pm$ 0.7k }(p=0.0616)   &  \textit{10.0k $\pm$ 0.7k} (p=0.0506)  \\
		\hline
	\end{tabular}
	\label{tab:SteppedBeamResult39}
\end{table*}
\begin{table*}[hbt]
	\centering
	\caption{FEs required to achieve HV$\protect^T$ = 0.75 for 59-segment beams.}
	\begin{tabular}{|c|c|c|c|c|c|}
		\hline
		Rule Type & Repair agent & RU1 & RU2 & RU3 & RU4 \\
		\hline\hline
		None 						&  None (base) & 19.0k $\pm$ 2.5k (p=0.0102)& 19.0k $\pm$ 2.5k (p=0.0015)& 19.0k $\pm$ 2.5k (p=0.0011)& 19.0k $\pm$ 2.5k (p=0.0027)\\
		\hline
		\hline
		\multirow{4}{1.2cm}{Power law rule}  & PL-RA1 & 15.6k $\pm$ 1.8k (p=0.0105) & \cellcolor{gray!25}\textbf{14.1k $\pm$ 2.3k} & 15.2k $\pm$ 3.1k (p=0.0164) & 16.1k $\pm$ 2.5k (p=0.0371)\\
									& PL-RA2 &\cellcolor{gray!25}14.8k $\pm$ 1.8k  & 15.0k $\pm$ 2.0k (p=0.0225)& \textbf{\textit{14.4k $\pm$ 2.8k }}(p=0.1015) & \textit{15.5k $\pm$ 1.8k} (p=0.0510)\\
									& PL-RA3 & 16.0k $\pm$ 2.2k (p=0.0042) & \textbf{15.8k $\pm$ 1.5k} (p=0.0218)& 16.6k $\pm$ 3.5k (p=0.0215) & 17.0k $\pm$ 2.6k (p=0.0446)\\
									& PL-RA-E & \textit{14.9k $\pm$ 3.1k} (p=0.1165) & \textit{\textbf{14.2k $\pm$ 3.6k}} (p=0.0911)& \cellcolor{gray!25}14.2k  $\pm$ 2.9k & \cellcolor{gray!25} 14.9k $\pm$ 2.5k \\
		\hline
		\hline
		\multirow{4}{1.2cm}{Inequality rule} & IQ-RA1 & 16.6k $\pm$ 4.1k (p=0.0215) & \textbf{16.0k $\pm$ 2.6k} (p=0.0411) & 16.6k $\pm$ 2.0k (p=0.0182) & 18.1k $\pm$ 2.2k (p=0.0341)\\
										& IQ-RA2 & 16.8k  $\pm$ 4.0k (p=0.0193) & \textbf{15.9k $\pm$ 2.9k} (p=0.0365) & 16.8k $\pm$ 3.0k (p=0.0179) & 17.6k $\pm$ 2.6k(p=0.0335)\\
										& IQ-RA3 & \textbf{16.5k  $\pm$ 4.3k} (p=0.0317) & 17.2k $\pm$ 3.1k (p=0.0357) & 17.2k $\pm$ 3.2k (p=0.0155) & 17.9k $\pm$ 2.4k(p=0.0273)\\
										& IQ-RA-E & 16.6k  $\pm$ 3.5k (p=0.0287) & \textbf{16.4k $\pm$ 4.2k} (p=0.0282) & 16.9k $\pm$ 2.7k (p=0.0293) & 17.8k $\pm$ 1.9k(p=0.0228)\\
		\hline
		\hline
		\multirow{2}{1.2cm}{Mixed rule} & PL-RA2+IQ-RA2 & \textit{15.1k $\pm$ 3.0k} (p=0.0583) & \textit{\textbf{14.5k $\pm$ 3.1k}} (p=0.0917) & \textit{14.6k $\pm$ 3.0k} (p=0.0715) & \textit{15.4k $\pm$ 2.5k} (p=0.0713)\\
								& PL-RA-E+IQ-RA-E & \textit{14.9k $\pm$ 2.6k} (p=0.0917) & \textit{\textbf{14.4k $\pm$ 2.9k}} (p=0.0663) & \textit{14.6k $\pm$ 2.6k} (p=0.0681) & \textit{15.2k $\pm$ 2.2k} (p=0.0558)\\
		\hline
	\end{tabular}
	\label{tab:SteppedBeamResult59}
\end{table*}

The results show many interesting observations as stated below.
\subsubsection{General observations}
Despite the median FEs being close, statistically base NSGA-II does not perform well compared to knowledge-based NSGA-II methods for both 39- and 59-segment problems. 
A positive aspect of the proposed algorithm is that it is still able to achieve a good performance with significantly low population size. 
For problems with expensive evaluation functions, this may stay beneficial for saving computational time.

\subsubsection{Power law vs inequality rules}
As can be seen from the table, for both 39- and 59-segment problems, the power law repair alone results in the best performance for each user case. Inequality rule-based repair alone in general results in worse performance compared to power-law-based repair. One possible reason could be the greater versatility of power laws in modeling complex relationships compared to simple inequality rules.

\subsubsection{Best performing algorithm for each rule usage scheme}
In the 39-segment case, PL-RA-E is the best performer for both RU1 and RU4. For RU2 and RU3, PL-RA2 is the best performer. It is to be noted that PL-RA-E has statistically similar performance to PL-RA2 for both RU2 and RU3. This shows that the ensemble method can be used to get good performance without the need for selecting a proper repair process. For the 59-segment case, PL-RA2 and PL-RA1 are the best performers for RU1 and RU2. In the cases of RU3 and RU4, PL-RA-E gives the best performance, with PL-RA2 having a comparable performance. PL-RA1 and PL-RA3 do not show comparable performance with PL-RA2 or PL-RA-E in most cases. For PL-RA1, adhering closely to the learned power law rules constrains NSGA-II in finding good solutions. PL-RA3 introduces a large amount of variance which is detrimental to the optimization process. A compromise between these two extremes, provided by PL-RA2 or PL-RA-E, is the logical step.

Figure~\ref{fig:SBResults} illustrates the results when RU2 is combined with the power law repair operators for both problem cases. The difference in the quality of solutions obtained after 20,000 FEs is prominent in the 59-segment ND front. In the median HV plots it is seen that the FEs required to reach HV$^T$ for PL-RA-E is close to the number needed by the best performing repair agent. Base NSGA-II and the repair operators all give good quality solutions at the end of the run for the 39-segment case. However, for the 59-segment case, base NSGA-II performs significantly worse.

\subsubsection{Relative performance of each rule usage scheme}
It can be seen from Tables~\ref{tab:SteppedBeamResult39} and \ref{tab:SteppedBeamResult59} that RU2 produces the best performance in 6 out of 10 cases for the 39-segment case, and 8 out of 10 cases for the 59-segment case. This shows that in terms of rule usage, using too few or too many of the learned rules is not effective in improving the algorithm's performance.

\subsubsection{Mixed relation repair agents}
The mixed relation repair agents (PL-RA2+IQ-RA2) and (PL-RA-E+IQ-RA-E), have statistically similar performance to the best algorithm for each user and both problem cases. Even though inequality repair operators perform worse than power law repair operators individually, their presence in the mixed repair agents do not hinder the performance, since only the high-performing rules are added to the VRG during creation. The proposed framework is robust enough to give good performance irrespective of the number of repair agents and type of rules.

\section{Optimal Power Flow Problem}
\label{sec:OPF}

Optimal power flow (OPF) is a common problem in power system engineering with MOEAs being used to solve the problem \cite{Datta2017,Basu2011}. 
The following objective functions are minimized: fuel cost, emissions, voltage deviation, and real power loss. In many cases, one or more of these objectives are considered in the literature, with the rest being kept as constraints. In this study, we consider two objectives: minimizing fuel cost and reducing fossil fuel emissions. Voltage deviation and power loss are kept as constraints. This version of the OPF problem is also known as the environmental economic dispatch (EED) problem \cite{Basu2011}.  

{\footnotesize\begin{align}
\label{eq:OPFProblem}
\text{Minimize } &C_F(\mathbf{P_G}, \mathbf{V_G}) = \sum_{i=1}^{N_{G}}\left(a_i + b_i P_{Gi} + c_i P_{Gi}^2\right),\\
\text{Minimize } &C_E(\mathbf{P_G}, \mathbf{V_G}) = \sum_{i=1}^{N_{G}} \left(\alpha_i + \beta_i P_{Gi} + \gamma_i P_{Gi}^2 + \zeta_i e^{(\lambda_i P_{Gi})}\right),\\
\text{Subject to } & \sum_{i=1}^{N_{bus}}(P_i - P_D - P_L)=0,\\
& VD_{\min} \leq VD \leq VD_{\max},\quad P_{Lmin} \leq P_L \leq P_{Lmax}, \nonumber \\
&Q_{Gimin} \leq Q_{Gi} \leq Q_{Gimax},\quad P_{smin} \leq P_s \leq P_{smax},\nonumber \\
&V_{smin} \leq V_s \leq V_{smax},\quad V_{PQimin} \leq V_{PQi} \leq V_{PQimax}, \nonumber
\end{align}}%
where $C_F$ is the fuel cost, $C_E$ is the emission cost, $N_G$ is the number of generators, $P_{Gi}$ is the real power output and $V_{Gi}$ is the voltage output of the $i^{th}$ generator, $(a_i, b_i, c_i)$ are the fuel cost coefficients, $(\alpha_i,\beta_i,\gamma_i,\zeta_i,\lambda_i)$ are the emission cost coefficients. $VD$ is the total voltage deviation of all the load buses, $P_L$ is the total real power loss, $Q_{Gi}$ is the reactive power output of the $i^{th}$ generator, $P_s$ is the real power output and $V_s$ is the voltage output of the slack bus, $V_{PQi}$ is the voltage at the $i^{th}$ load/P-Q bus. $P_D$ is the power demand and $N_{bus}$ is the total number of buses. A load flow analysis must be performed to satisfy the equality constraint. We use MATPOWER \cite{Zimmerman2011} as the load flow solver. We consider IEEE 118-bus and 300-bus systems in this study. 

\subsection{Experimental settings}
The bus details, along with the numbers of decision variables and constraints, are given in Table~\ref{tab:IEEEBus}. The types of decision variables and their corresponding ranges are given in Table~\ref{tab:IEEEBusVar}.
\begin{table}[hbt]
	\renewcommand{\arraystretch}{1.2}
	\centering
	\caption{IEEE bus system specifications.}
	{\scriptsize \begin{tabular}{|c|c|c|p{0.5cm}|p{0.9cm}|p{1cm}|}
		\hline
		System & Generators & Transformers & Load bus & Decision variables & Constraints \\
		\hline
		IEEE 118-bus & 54 & 11 & 64 & 115 & 240 \\
		IEEE 300-bus & 69 & 107 & 231 & 243 & 604 \\
		\hline
	\end{tabular}}
	\label{tab:IEEEBus}
\end{table}

\begin{table}[hbt]
	\centering
	\caption{OPF decision variable types and ranges.}
	\begin{tabular}{|c|c|}
		\hline
		Variable & Range \\
		\hline
		Generator power output ($P_{Gi}$) & [30, 100]  \\
		Generator voltage ($V_{Gi}$) & [0.95, 1.05]  \\
		Transformer tap ratio ($T_i$) & [0.9, 1.1]\\
		\hline
	\end{tabular}
	\label{tab:IEEEBusVar}
\end{table}

Experimental settings are the same as in the stepped beam problem except that the population size is set to be 50 and the maximum number of generations is set as 400 for both IEEE 118 and 300-bus systems. From Table \ref{tab:RuleType}, $\rho_i$ and $\varepsilon_{ij}$ are set as 1, and $e_{ij}^{min}$ is set to 0.01. Two variable groups are defined and shown in Table~\ref{tab:OPFGroups}.
\begin{table}[hbt]
	\centering
	\caption{OPF variable groups.}
	\begin{tabular}{|c|c|p{1.2cm}|p{1.2cm}|}
		\hline
		\multirow{2}{*}{Group} & \multirow{2}{*}{Variable Type} & \multicolumn{2}{c|}{Variable Indices} \\
		\cline{3-4} & & 118-bus & 300-bus \\
		\hline
		 $G_{\rm opf1}$ & Generator power and voltage & [1-104] & [1-136]\\
		 $G_{\rm opf2}$ & Transformer tap ratio & [105-115] & [137-243]\\
		\hline
	\end{tabular}
	\label{tab:OPFGroups}
\end{table}

\subsection{Experimental results and discussion} 
\begin{figure*}[htb]
\begin{center}	
\begin{subfigure}{0.48\textwidth}
	\centering\includegraphics[width=0.8\columnwidth]{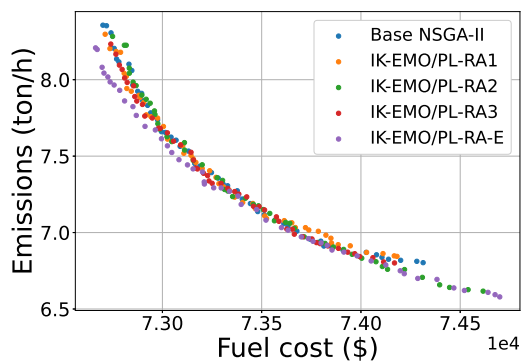}
		\caption{ND Front for one run.}
		\label{fig:pfOpf300}
	\end{subfigure}\hfill
	\begin{subfigure}{0.48\textwidth}
	\centering\includegraphics[width=0.8\columnwidth]{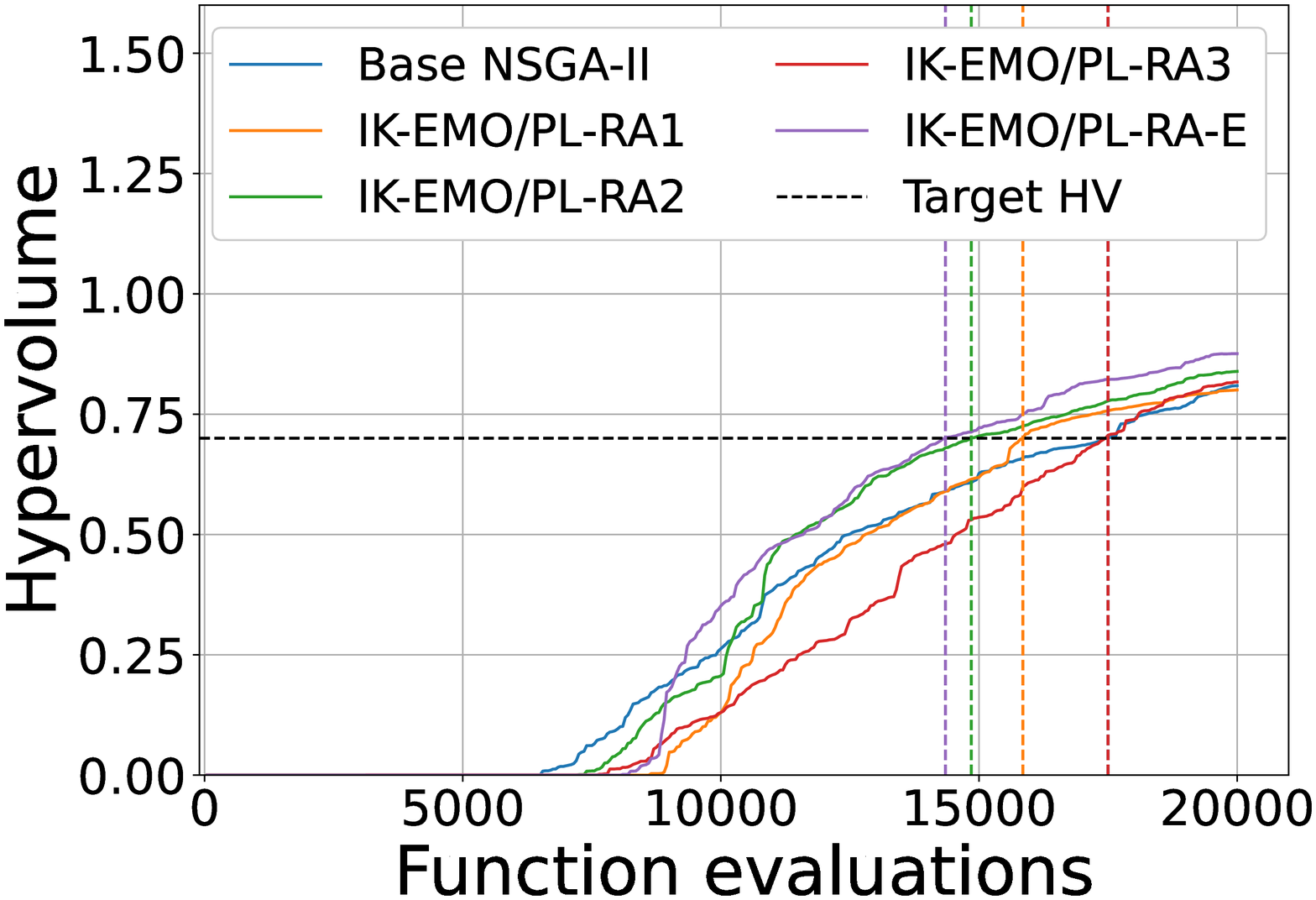}
		\caption{HV plot over 20 runs.}
		\label{fig:hvOpf300}
	\end{subfigure}
	\end{center}
	\caption{ND front and hypervolume plots obtained with RU2 and power law repair agents for IEEE 300-bus OPF problem.}
	\label{fig:ieeeResults}
\end{figure*}

Tables~\ref{tab:OPFResult118} and \ref{tab:OPFResult300} show the optimization results for the IEEE 118- and 300-bus systems, respectively. The ND front obtained in a single run using four power law repair methods for RU2 are shown in  
Figure~\ref{fig:pfOpf300} for the IEEE 300-bus system. The corresponding median HV plots over the course of the optimization are shown in 
Figure~\ref{fig:hvOpf300}. Similar behavior is observed for the 118-bus system (see supplementary document).
\begin{table*}[hbt]
	\centering
	\caption{FEs required to achieve HV$\protect^T$ = 0.74 for IEEE 118-bus system.}
	\begin{tabular}{|c|c|c|c|c|c|}
		\hline
		Rule Type & Repair agent & RU1 & RU2 & RU3 & RU4 \\
		\hline\hline
		None 						& None (base) &  6.5k $\pm$ 0.6k (p=0.0216)  & 6.5k $\pm$ 0.6k (p=0.0286) &  6.5k $\pm$ 0.6k (p=0.0337)  &  6.5k $\pm$ 0.6k (p=0.0432) \\
		\hline
		\hline
		\multirow{4}{1.2cm}{Power law rule}  & PL-RA1 & 5.9k $\pm$ 0.2k (p=0.0101) & \textbf{5.8k $\pm$ 0.4k} (p=0.0417)& 5.8k $\pm$ 0.5k (p=0.0119) & \cellcolor{gray!25}6.0k $\pm$ 0.2k\\
									& PL-RA2 & \cellcolor{gray!25}4.5k $\pm$ 0.4k & \textbf{\textit{5.0k $\pm$ 0.9k}} (p=0.0805)& \cellcolor{gray!25}5.2k $\pm$ 0.2k & \textit{6.1k $\pm$  0.3k} (p=0.0813)\\
									& PL-RA3 & 6.8k $\pm$ 0.5k (p=0.0152) & 6.3k $\pm$ 0.7k (p=0.0398)& \textbf{6.2k $\pm$ 0.4k} (p=0.0817) & 7.0k $\pm$ 0.6k (p=0.0086)\\
									& PL-RA-E & \textit{4.7k $\pm$ 0.4k} (p=0.0656) & \cellcolor{gray!25}\textbf{4.2k $\pm$ 0.5k} & \textit{5.4k $\pm$ 0.3k} (p=0.0680) & \textit{6.1k $\pm$ 0.2k} (p=0.0727)\\
		\hline
		\hline
		\multirow{4}{1.2cm}{Inequality rule} & IQ-RA1 & 6.6k $\pm$ 0.4k (p=0.00119) & 6.9k $\pm$ 0.7k (p=0.0255) & \textbf{6.1k $\pm$ 0.1k} (p=0.0341) & 7.0k $\pm$ 0.5k (p=0.0338)\\
										& IQ-RA2 & 6.6k $\pm$ 0.2k (p=0.0065) & 6.8k  $\pm$ 0.5k (p=0.0021) & \textbf{6.4k $\pm$ 0.3k} (p=0.0279) & 6.8k $\pm$ 0.1k (p=0.0332)\\
										& IQ-RA3 & 6.5k  $\pm$ 0.4k (p=0.0138) & 6.4k $\pm$ 0.2k (p=0.0018) & \textbf{6.1k $\pm$ 0.3k} (p=0.0275) & 7.5k $\pm$ 0.4k(p=0.0320)\\
										& IQ-RA-E & 6.5k $\pm$ 0.1k (p=0.0129) & \textbf{6.3k $\pm$ 0.4k} (p=0.0121) & 6.8k $\pm$ 0.3k (p=0.0116) & 7.0k $\pm$ 0.2k (p=0.0112)\\
		\hline
		\hline
		\multirow{2}{1.2cm}{Mixed rule} & PL-RA2 + IQ-RA2 & \textit{\textbf{4.8k $\pm$ 0.3k}} (p=0.0722) & \textit{\textbf{4.8k $\pm$ 0.4k}} (p=0.0713) & \textit{5.4k $\pm$ 0.3k} (p=0.0841) & \textit{6.2k $\pm$ 0.1k} (p=0.0748)\\
								& PL-RA-E + IQ-RA-E & \textit{4.6k $\pm$ 0.1k} (p=0.0903) & \textit{\textbf{4.4k $\pm$ 0.2k}} (p=0.0667) & 5.6k $\pm$ 0.1k (p=0.0144) & \textit{6.2k $\pm$ 0.2k} (p=0.0919)\\
		\hline
	\end{tabular}
	\label{tab:OPFResult118}
\end{table*}
\begin{table*}[hbt]
	\centering
	\caption{FEs required to achieve HV$\protect^T$ = 0.70 for IEEE 300-bus system.}
	\begin{tabular}{|c|c|c|c|c|c|}
		\hline
		Rule Type & Repair agent & RU1 & RU2 & RU3 & RU4 \\
		\hline\hline
							None 	& None (base) &  17.5k $\pm$ 1.2k (p=0.0142) & 17.5k $\pm$ 1.2k (p=0.0205) & 17.5k $\pm$ 1.2k (p=0.0130) & 17.5k $\pm$ 1.2k (p=0.0091)\\
		\hline
		\hline
		\multirow{4}{1.2cm}{Power law rule}  & PL-RA1 &  \textit{\textbf{14.8k	$\pm$ 0.7k}} (p=0.0878) & 15.9k $\pm$ 0.8k (p=0.0110) & 16.2k $\pm$ 0.6k (p=0.0035) & 16.8k $\pm$ 0.6k (p=0.0063)\\
									& PL-RA2 & \textit{15.1k $\pm$ 0.6k} (p=0.0753)  & \textit{\textbf{14.9k $\pm$ 0.5k}} (p=0.0518) & 15.9k $\pm$ 0.2k (p=0.0239)  & \textit{14.9k $\pm$ 0.2k} (p=0.0657) \\
									& PL-RA3 &  18.5k	$\pm$ 0.3k (p=0.0413) & \textbf{17.5k $\pm$ 1.0k} (p=0.0017) & 19.3k $\pm$ 1.2k (p=0.0181) & 18.8k $\pm$ 0.5k (p=0.0025)\\
									& PL-RA-E &  \cellcolor{gray!25}14.5k	$\pm$ 0.2k & \cellcolor{gray!25}\textbf{14.4k $\pm$ 0.3k} & \cellcolor{gray!25}14.8k $\pm$ 0.6k & \cellcolor{gray!25}15.6k $\pm$ 0.9k\\
		\hline
		\hline
		\multirow{4}{1.2cm}{Inequality rule} & IQ-RA1 & \textbf{15.2k $\pm$ 0.9k} (p=0.0315) & 16.0k $\pm$ 1.1k (p=0.0033) & 18.5 $\pm$ 1.3k (p=0.0059) & 18.2 $\pm$ 1.0k (p=0.0024)\\
										 & IQ-RA2 & 16.9k $\pm$ 1.2k (p=0.0122) & \textbf{16.5k $\pm$ 1.0k} (p=0.0015) & 17.6 $\pm$ 0.7k (p=0.0073) & 18.0 $\pm$ 1.3k (p=0.0022)\\
										 & IQ-RA3 & 16.8k $\pm$ 1.0k (p=0.0286) & \textbf{16.1k $\pm$ 0.8k} (p=0.0112) & 16.6 $\pm$ 1.7k (p=0.0012) & 19.5 $\pm$ 1.5k (p=0.0016)\\
										 & IQ-RA-E & 15.9k $\pm$ 0.8k (p=0.0252) & 16.8k $\pm$ 0.8k (p=0.0104) & \textbf{15.7 $\pm$ 0.6k} (p=0.0076) & 18.3 $\pm$ 1.1k (p=0.0032)\\
		\hline
		\hline
		\multirow{2}{1.2cm}{Mixed rule} & PL-RA2 + IQ-RA2 & \textit{14.9k $\pm$ 0.8k} (p=0.0991) & \textbf{\textit{14.8k $\pm$ 0.7k}} (p=0.0836) & 16.3k $\pm$ 0.2k (p=0.0103) & \textit{16.0k $\pm$ 3.0k} (p=0.0528)\\
		 					   & PL-RA-E + IQ-RA-E & \textit{14.7k $\pm$ 0.7k} (p=0.1013) & \textit{\textbf{14.6k $\pm$ 0.9k}} (p=0.0811) & \textit{15.0k $\pm$ 0.8k} (p=0.0713) & \textit{16.2k $\pm$ 0.9k} (p=0.661)\\
		\hline
	\end{tabular}
	\label{tab:OPFResult300}
\end{table*}

\subsubsection{General observations}
Base NSGA-II is outperformed by PL-RA1, PL-RA2 and PL-RA-E as well as the mixed rule repair agents. However, for PL-RA3 and the inequality repair operators, base NSGA-II produces comparable performance in most cases.  Good performance with low population size is obtainable by the power-law-based repair agents.

\subsubsection{Power law vs inequality rules}
For both problem cases, a power law repair operator is the best performer for each user, as in the stepped beam problem. Inequality-rule-based repair operators in general result in worse performance compared to power-law-based repair operators as well as the base NSGA-II. This, as in the stepped beam problem, is a result of the power laws being able to more accurately model the inter-variable relationships. 
For PL-RA3, having a high variance ($2\sigma_c$) during repair is harmful to the optimization, resulting in comparable or worse performance than base NSGA-II in general.

\subsubsection{Best performing algorithm for each user}
In the IEEE 118-bus system, PL-RA-E is the best performer for RU2, with PL-RA2 showing comparable performance. For RU1 and RU3, PL-RA2 is the best performer, with PL-RA-E showing comparable performance. For RU4, PL-RA1 is the best, with PL-RA2 and PL-RA-E giving statistically similar performance. As in the case of the stepped beam problems, PL-RA-E is either the best or gives statistically similar performance. Thus, good performance can be obtained without the need to determine which power-law-based repair operator is the best.

Figure~\ref{fig:ieeeResults} illustrates the results with RU2 combined with the power law repair operators for both problem cases. The difference in the quality of solutions obtained after 20,000 FEs is more prominent in the IEEE 300-bus case. In the median HV plots it is seen that the number of FEs required to reach HV$^T$ for PL-RA-E is close to that of PL-RA2. Base NSGA-II and PL-RA3 show worse performance than the others.

\subsubsection{Relative performance of each user}
It can be seen from Tables~\ref{tab:OPFResult118} and \ref{tab:OPFResult300} that RU2 produces the best performance in 6 out of 10 cases for the IEEE 118-bus case, and 7 out of 10 cases for the IEEE 300-bus case. This shows that in terms of rule usage, using too few or too many of the learned rules is detrimental to the optimization performance in general, which is similar to the conclusions made in the stepped beam design problems.

\subsubsection{Mixed relation repair agents}
The mixed relation repair agents (PL-RA2+IQ-RA2) and (PL-RA-E+IQ-RA-E), have statistically similar performance to the best algorithm for each user and both problem cases. As in the stepped beam problems, the worse performance of the inequality repair operators does not hinder the performance of the mixed relation operators.

\section{Truss Design Problem}
\label{sec:truss_problem}
Finally, we consider a commonly-used truss design problem involving two objectives, and 1,416 highly nonlinear constraints. The truss has 1,100 members and 316 nodes, making a total of 1,179 variables, making it a large-scale problem. The details of the problem description are provided in the supplementary document.

Experimental settings are similar to those of the previous problems. Population size is set to 100 and the maximum number of generations is set as 10,000. Thus, the total computational budget comes out to be 1 million FEs. From Table \ref{tab:RuleType}, $\rho_i$ and $\varepsilon_{ij}$ are set as 0.1, and $e_{ij}^{min}$ is set to 0.01. Multiple variable groups are defined for this problem based on the relative location and alignment of the beams as shown in Table~\ref{tab:TrussGroups1100}.
\begin{table}[hbt]
	\renewcommand{\arraystretch}{1.1}
	\centering
	\caption{Variable groups for the 1,100-member truss cases. Each group has comparable variables having identical units and scales.}
	\footnotesize{
		\begin{tabular}{|c|c|c|} \hline
			{Group} & {Variable Type} & Variable Indices\\
			\hline
			$G_{t1}$ & $l_i$ of vertical members & $[1101-1179]$ \\
			$G_{t2}$ & $r_i$ of top longitudinal members & $[79-156], [235-312]$ \\
			$G_{t3}$ & $r_i$ of bottom longitudinal members & $[1-78], [157-234]$ \\
			$G_{t4}$ & $r_i$ of vertical members & $[313-391]$ \\
			\hline
		\end{tabular}
		\label{tab:TrussGroups1100}}
\end{table}

\subsection{Experimental results and discussion}
Results are presented in Table~\ref{tab:TrussResults1380}. 
\begin{table*}[hbt]
	\centering
	\caption{FEs required to achieve HV$\protect^T$ = 0.78 for 1,100-member truss.}
	\begin{tabular}{|c|c|c|c|c|c|}
		\hline
		Rule Type & & RU1 & RU2 & RU3 & RU4 \\
		\hline
		None 						& Base &  874k $\pm$ 10k (p = 0.0043)  & 874k $\pm$ 10k (p = 0.0017) &  874k $\pm$ 10k (p = 0.0062)  &  874k $\pm$ 10k (p = 0.0053) \\
		\hline
		\hline
		\multirow{4}{1.2cm}{Power law rule}  & PL-RA1 & 802k $\pm$ 8k (p=0.0057) & 792k $\pm$ 8k (p=0.0129) & \textbf{786k $\pm$ 13k} (p=0.0140) & 812k $\pm$ 6k (p=0.0023)\\
									& PL-RA2 & \textit{680k $\pm$ 11k} (p=0.0633) & \textbf{\textit{678k $\pm$ 15k}} (p=0.0793) & \cellcolor{gray!25}688k $\pm$ 10k  & \cellcolor{gray!25}744k $\pm$ 17k \\
									& PL-RA3 & \textbf{963k $\pm$ 21k} (p=0.0005) & 1M (HV=0.74) & 1M (HV=0.71)  & 1M (HV=0.66) \\
									& PL-RA-E & \cellcolor{gray!25}672k $\pm$ 9k  & \cellcolor{gray!25}\textbf{656k $\pm$ 18k} & \textit{693k $\pm$ 16k} (p=0.1016) & \textit{754k $\pm$ 24k} (p=0.1163)\\
		\hline
		\hline
		\multirow{4}{1.2cm}{Inequality rule} & IQ-RA1 & 843k $\pm$ 15k (p=0.0015) & 828k $\pm$ 20k (p=0.0115) & \textbf{822k $\pm$ 23k} (p=0.0169) & 851k $\pm$ 8k (p=0.0325)\\
										& IQ-RA2 & 836k $\pm$ 12k (p=0.0039) & 838k $\pm$ 19k (p=0.0248) &  \textbf{803k $\pm$ 18k} (p=0.0465) & 834k $\pm$ 6k (p=0.0318)\\
										& IQ-RA3 & 839k $\pm$ 21k (p=0.0036) & \textbf{819k $\pm$ 29k} (p=0.0219) & 826k $\pm$ 10k (p=0.0454) & 837k $\pm$ 5k (p=0.0414)\\
										& IQ-RA-E &  840k $\pm$ 27k (p=0.0024) & 816k $\pm$ 22k (p=0.0351) & \textbf{798k $\pm$ 13k} (p=0.0311) & 843k $\pm$ 11k (p=0.0223)\\
		\hline
		\hline
		\multirow{2}{1.2cm}{Mixed rule} & PL-RA2 + IQ-RA2 & \textit{682k $\pm$ 10k} (p=0.0669) & \textbf{\textit{677k $\pm$ 14k}} (p=0.0816) & \textit{691k $\pm$ 8k} (p=0.0772) & \textit{751k $\pm$ 10k} (p=0.0522)\\
							   & PL-RA-E + IQ-RA-E & \textit{676k $\pm$ 9k} (p=0.0714) & \textit{\textbf{659k $\pm$ 12k}} (p=0.0814) & \textit{696k $\pm$ 13k} (p=0.0611) &\textit{752k $\pm$ 23k} (p=0.0699)\\
		\hline
	\end{tabular}
	\label{tab:TrussResults1380}
\end{table*}
\begin{figure*}[htb]
\centering
\begin{subfigure}{0.48\textwidth}
	\centering\includegraphics[width=0.8\columnwidth]{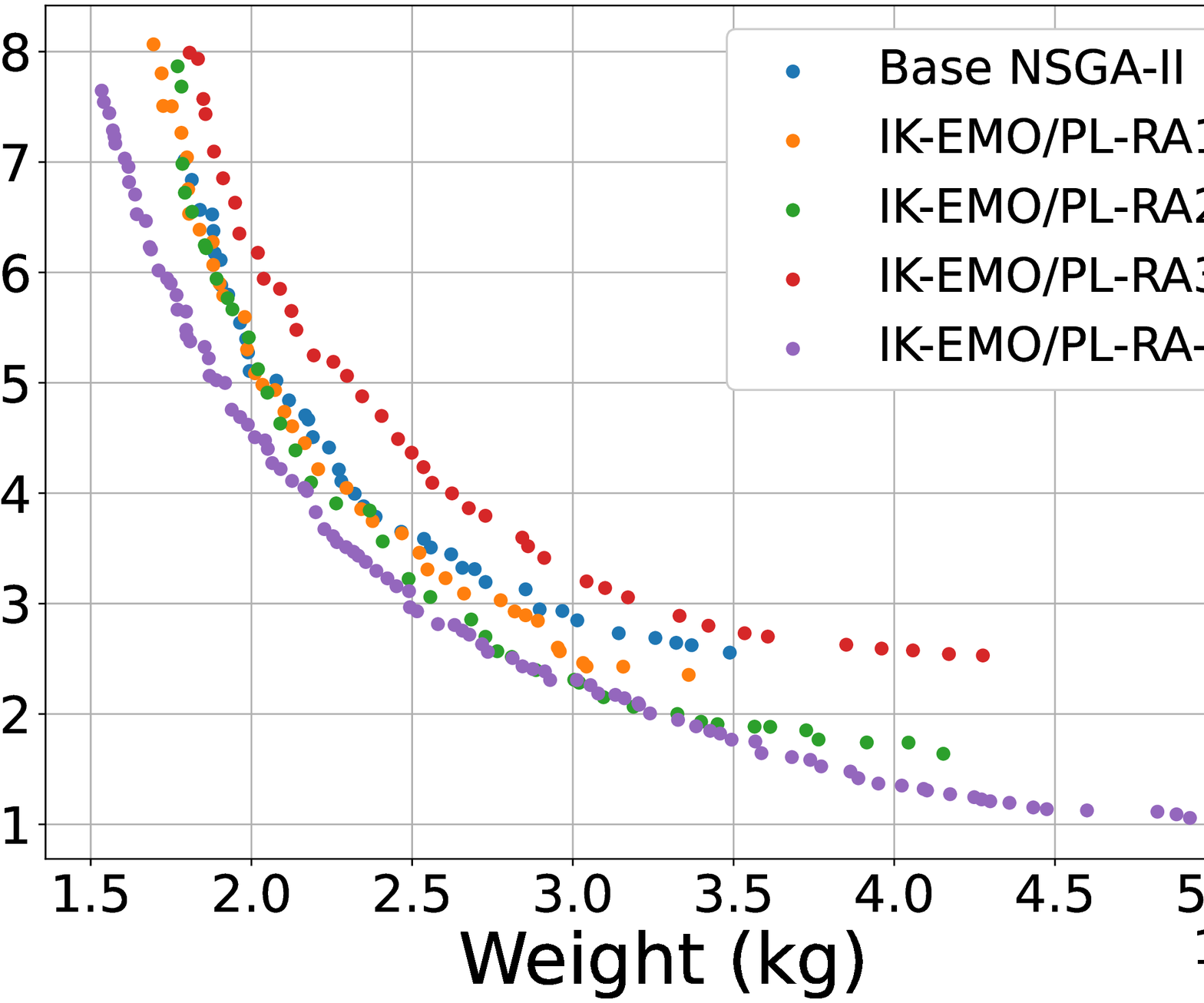}
	\caption{ND Front for one run.}
	\label{fig:pfTruss1100}
\end{subfigure}\hfill
\begin{subfigure}{0.48\textwidth}
	\centering\includegraphics[width=0.8\columnwidth]{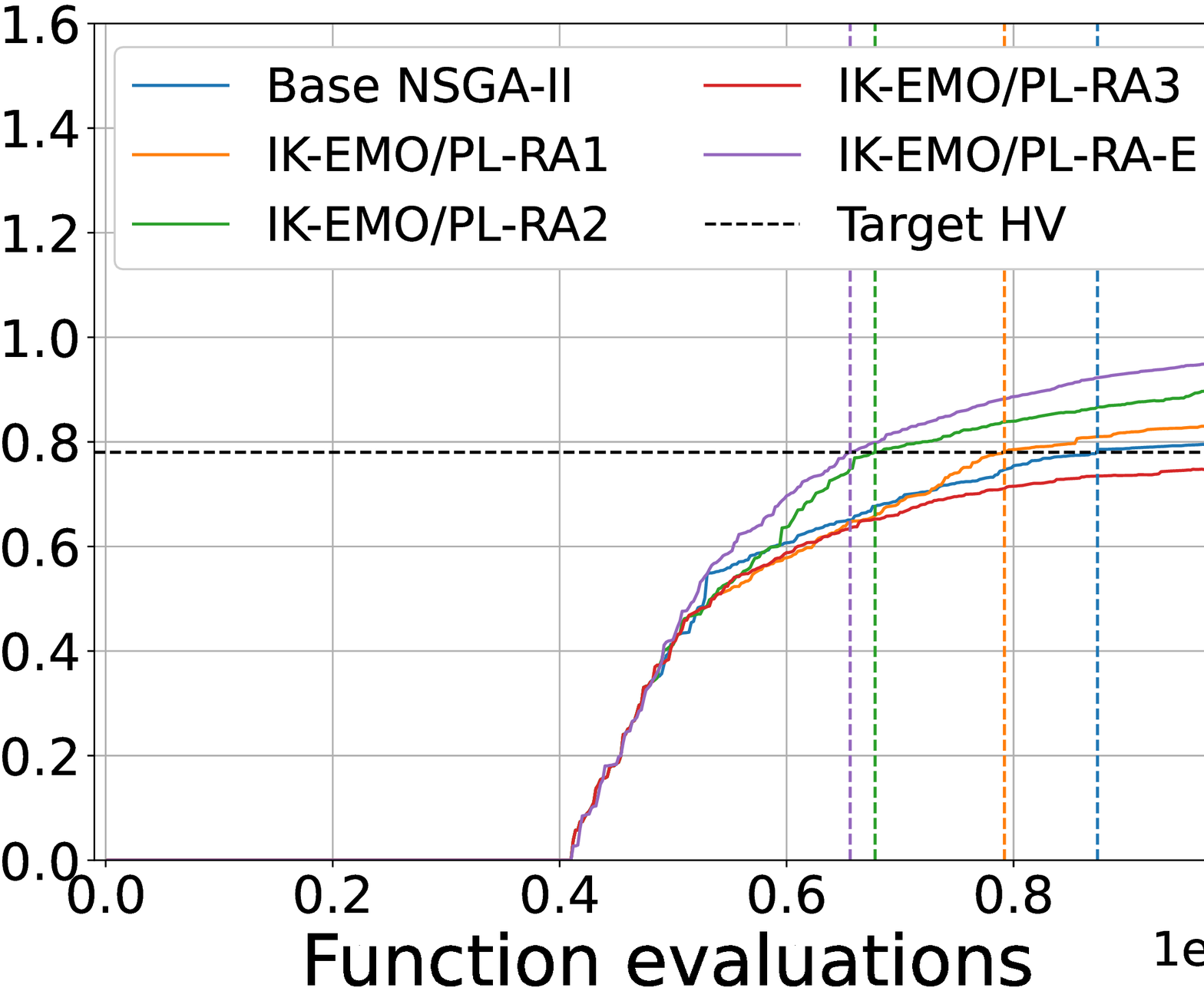}
	\caption{HV plot over 20 runs.}
	\label{fig:hvTruss1100}
\end{subfigure}
\caption{ND fronts and hypervolume plots obtained by \IKEMO\ with RU2 and power law repair agents for 1,100-member truss design problem.}
\label{fig:Truss1100}
\end{figure*}
Base NSGA-II is outperformed by most repair operators. As in the previous two problems, power-law-based approaches perform better than inequality-based approaches, but the ensemble-based approach performs overall the best with an intermediate use of repair (RU2). More information are put in supplementary document.
 
\section{Summary of results}
For every problem we have a total of 11 different algorithms including base NSGA-II and 10 repair schemes. For each problem, ranking based on FEs to achieve the target HV for four users is summarized in Table~\ref{tab:overallRank}. An algorithm with statistically similar performance to the best performing algorithm is assigned a rank of 1. More details are presented in the supplementary document.
It is seen that the top ranked algorithm is PL-RA-E followed by mixed PL-RA2+IQ-RA2, and mixed PL-RA-E+IQ-RA-E, highlighting the superiority of the ensemble approach.
\begin{table*}[hbt]
	\renewcommand{\arraystretch}{1.2}
	\centering
	\caption{Ranking of different repair agents on multiple problems. A detailed breakdown is provided in the supplementary material.}
	\label{tab:overallRank}
	\begin{tabular}{|l@{\hspace{2pt}}|p{1cm}@{\hspace{2pt}}|p{1cm}@{\hspace{2pt}}|p{1cm}@{\hspace{2pt}}|p{1cm}@{\hspace{2pt}}|p{1.1cm}@{\hspace{2pt}}|p{1cm}@{\hspace{2pt}}|p{1cm}@{\hspace{2pt}}|p{1cm}@{\hspace{2pt}}|p{1.05cm}@{\hspace{2pt}}|p{1.15cm}@{\hspace{2pt}}|p{1.25cm}|} \hline
		Problem & Base & PL-RA1 & PL-RA2 & PL-RA3 & PL-RA-E & IQ-RA1 & IQ-RA2 & IQ-RA3 & IQ-RA-E & PL-RA2 +IQ-RA2 & PL-RA-E +IQ-RA-E \\ \hline
		Beam 78-var 	 & 9 & 6 & \textbf{1} & 7 & \textbf{1} & 11 & 5 & 9 & 8 & \textbf{1} & \textbf{1} \\\hline
		Beam 118-var  & 11 & 5 & 4 & 6 & \textbf{1} & 8 & 7 & 10 & 9 & \textbf{1} & \textbf{1} \\\hline
		Power 115-var & 7 & 5 & \textbf{1} & 7 & \textbf{1} & 10 & 10 & 6 & 9 & \textbf{1} & 4 \\\hline
		Power 243-var & 10 & 5 & 3 & 11 & \textbf{1} & 7 & 8 & 8 & 6 & 4 & \textbf{1} \\\hline
		Truss 1179-var & 10 & 5 & \textbf{1} & 11 & \textbf{1} & 9 & 6 & 8 & 6 & \textbf{1} & \textbf{1} \\\hline\hline
		Final Rank & 11 & 5 & 4 & 9 & \textbf{1} & 10 & 6 & 8 & 7 & 2 & 2 \\\hline
	\end{tabular}
\end{table*}

\section{Synchronous vs asynchronous user interaction}
\label{sec:syncAsync}
In the previous sections, it is assumed that a user's feedback will be available soon after the learned rules are presented to the user.
However, in real world applications it is more likely that users will take some finite time to come up with a preferred ranking of the rules. Pausing the optimization (synchronous user interaction) until the user provides feedback may be inefficient for problems with expensive function evaluations. Continuing the optimization tasks while the user finalizes their feedback (asynchronous user interaction) is a practical and promising approach. We implement a simplistic asynchronous scenario here to investigate the effect of delayed feedback from users. 
We consider both 118- and 300-bus OPF problems for this purpose. The user feedback lag ($T_{U}$) is expressed in terms of the number of FEs that could have been executed between the time the user is presented with a set of rules and the time when the user is ready with some feedback (ranking of rules). For simplicity, $T_{U}$ is assumed to be constant for every round of user interaction. In the synchronous user interaction case, $T_U$ is undefined, as the optimization is put on hold until the user specifies a ranking of the rules. The learning interval ($T_L$) -- number of FEs executed between two consecutive rule learning tasks -- and repair interval ($T_R$) -- number of FEs executed between two consecutive repair operations -- are set to 500 FEs, (with 50 population members, this means after every 10 generations). 
When the user provides feedback--- a ranking of the previous rule set provided to them---new rules having identical structure to previous rules are given priority, and the rest are discarded. But, instead of using previous rules' statistics (means and standard deviations of $c$, for example), statistics of the new rules are used to repair. The optimization with repair operations then proceeds with the updated preferred ranking of rules. 

In the asynchronous user interaction case, once a VRG is constructed for the first time, \IKEMO\ is ready to provide information to the user about the learned rules every $T_L$ FEs if the user is available. However, we let the optimization proceed normally without waiting for user feedback and perform a repair operation using the learned rules. 
After $T_U$ FEs, the user is ready to provide feedback. Then, the user feedback on the immediate past rules is combined with the latest learned rules, as in the synchronous case, and repair is performed using the common rules but with latest rules' statistics. If $T_U \leq T_L$, the learned rules available to \IKEMO\ will be the same as the ones provided to the user before. Repair is performed immediately after the user provides the feedback. If $T_U > T_L$, the latest learned rules may be different than the ones the user was provided. It is to be noted that in the asynchronous case, $T_R$ is not an adjustable parameter, since repair is performed as soon as one learning phase is complete, or the user has provided some feedback. A figure illustrating both processes in detail is provided in the supplementary document. In this study, we use different $T_{U}$ values from 125 to 500 FEs ($T_U \leq T_L$) and from 1,000 to 4,000 FEs $(T_U > T_L)$. 
\begin{table*}[hbt]
	\centering
	\caption{FEs required to achieve HV$\protect^T$ for IEEE 118 and 300-bus OPF problems over 20 runs with multiple instances of asynchronous user interaction. Best performing algorithm for each row is marked in bold. Statistically similar results to the best are marked in italics.}
	\begin{tabular}{|c|c|c|c|c|c|c|c|c|}
		\hline
		\multirow{2}{*}{Problem} & \multirow{2}{*}{HV$^T$} &  \multirow{2}{*}{Base NSGA-II} & \multicolumn{6}{c|}{Decision-making lag ($T_{U}$)}\\
		\cline{4-9} & & & 125 & 250 & 500 & 1000 & 2000 & 4000\\
		\hline
									  IEEE 118-bus & 0.74 
									  &	6.5k $\pm$ 0.6k 
									  & \textbf{4.2k $\pm$ 0.1k} 
									  & \textit{4.4k $\pm$ 0.3k} 
									  & \textit{4.5k $\pm$ 0.4k} 
									  & \textit{4.8k $\pm$ 0.4k} 
									  & 5.2k $\pm$ 0.2k 
									  & 6.0k $\pm$ 0.5k\\
		
										IEEE 300-bus & 0.70 
										& 17.5k $\pm$ 1.2k 
										& \textbf{14.4k $\pm$ 0.3k} 
										& \textit{14.5k $\pm$ 0.1k} 
										&  \textit{14.4k $\pm$ 0.5k} 
										&  \textit{14.7k $\pm$ 0.4k} 
										& 15.1k $\pm$ 0.2k 
										& 16.0k $\pm$ 0.3k \\
		\hline
	\end{tabular}
	\label{tab:asyncResults}
\end{table*}
Table~\ref{tab:asyncResults} shows that for both IEEE 118-bus and 300-bus systems, a lag of up to 1,000 FEs gives statistically similar performance to the synchronous case. For lags of 2,000 FEs or higher, the performance deteriorates. But even with large lags, \IKEMO\ is robust enough to perform better than base NSGA-II. A quicker user feedback with $T_U \leq T_L$ produces the optimal performance, as expected. With a large lag time for user decision, feedback based on old rules has a detrimental effect. Thus, asynchronous interaction is suitable for cases where the objective evaluation is overly expensive, providing users relatively more time to make a decision on preferential ranking of rules. 

A second study evaluates the effect of user feedback lag time for a fixed overall computational time of $T_c$ units. For the synchronous case, the effective number of FEs allocated for the optimization operations becomes small, since a part of $T_c$ is now consumed by the user to make a decision. For multiple values of $T_{U}$, we compare the final HV obtained for asynchronous interaction with non-zero lag cases with that of the synchronous user interaction case. Table~\ref{tab:SyncResults} shows such a comparison with lag values varying from 125 to 4,000 FEs for a fixed overall execution time of $T_c=20,000$ FEs.
\begin{table*}[hbt]
	\centering
	\caption{Final median HV obtained for IEEE 118 and 300-bus OPF problems over 20 runs with multiple instances of synchronous user interaction. Best performing algorithm for row is marked in bold. Algorithms with statistically similar performance to the best performing algorithm are marked in italics. Results are presented graphically in the supplementary document.}
	\begin{tabular}{|c|c|c|c|c|c|c|c|c|c|c|c|c|c|}
		\hline
		\multirow{2}{*}{Problem} & $T_c$ &   \multicolumn{2}{c|}{$T_{U}$ = 125}  & \multicolumn{2}{c|}{$T_{U}$ = 250}  & \multicolumn{2}{c|}{$T_{U}$ = 500} & \multicolumn{2}{c|}{$T_{U}$ = 1,000}  & \multicolumn{2}{c|}{$T_{U}$ = 2,000} & \multicolumn{2}{c|}{$T_{U}$ = 4,000} \\
		\cline{3-14} &  & Sync & Async & Sync & Async & Sync & Async & Sync & Async & Sync & Async & Sync & Async\\
		\hline
		IEEE 118-bus & 20k  
		& \textbf{0.86} & \textit{0.84} 
		& \textbf{0.86} & \textit{0.84} 
		& 0.80 & 0.79 
		& 0.74 & 0.79 
		& 0.69 & 0.77 
		& 0.51& 0.76\\
		
		IEEE 300-bus & 20k 
		& \textbf{0.84} & \textbf{0.84} 
		& \textit{0.83} & \textit{0.81} 
		& 0.74 & \textit{0.82} 
		& 0.71 & \textit{0.80} 
		& 0.66 & \textit{0.80} 
		& 0.43& 0.77\\
		\hline
	\end{tabular}
	\label{tab:SyncResults}
\end{table*}
From the results, we observe that the performance of the synchronous case drops drastically when $T_{U}$ is increased for both 118- and 300-bus OPF problems. This is because for a large lag in making decisions, more time is wasted in the decision-making and less execution time is provided for running the optimization operations. For the asynchronous case, the performance drops slowly with $T_U$,
since \IKEMO\ is able to use the full computational budget of 20k FEs for both 118- and 300-bus cases. This outweighs any performance loss caused by using outdated rules. 

From the results presented in this section, it is evident that user feedback lag is an important practical factor that will have an effect in an interactive optimization procedure. These preliminary results suggest that for a small lag, both synchronous and asynchronous implementations are viable options. However, for a large anticipated lag, asynchronous implementation may provide an advantage.

\section{Conclusions and future work}
\label{sec:conclusion}
In this paper, we have proposed the \IKEMO\ framework which interleaves interactive optimization with knowledge augmentation to obtain better quality solutions faster. Power law, inequality, and mixed rules are extracted, together with their degrees of statistical adherence, from the ND solutions at a regular interval of generations. A computationally efficient graph data structure-based (VRG) knowledge processing method has been proposed to store and process multiple pairwise variable interactions. A user is then expected to provide a ranking of the learned rules based on his/her perception of the validity of the rules. A repair agent has been proposed to utilize the VRG with user-supplied ranking to repair offspring solutions. The study has created six repair schemes with three different degrees --- tight, medium, and loose -- of rule adherence. A mixed power law and inequality based repair has also been used. Finally, three ensemble-based repair schemes which adaptively use power law, inequality or both have been proposed. These 10 repair schemes have been implemented with four different rule usage schemes RU1-RU4, using 10\% (conservative), to 100\% (liberal) of the learned rules.

The proposed framework has been applied to three large-scale two-objective practical problems: 78- and 118-variable stepped beam design problems, 115- and 243-variable optimal power flow problems, and a 1,179-variable truss design problem.  
Experimental results on all problems have consistently shown that (i) usage of a moderate number of rules (20\%) combined with a moderate degree of rule adherence produces the best performance, and (ii) power law rules, individually, produce better performance than inequality rules. Moreover, ensemble-based repair operators provide comparable performance to the best performing individual repair operators. Use of ensembles eliminates the need to experiment to find the right rule adherence for a new problem. \IKEMO\ is also able to work with very low population sizes, even for a large-scale problem. 

A preliminary study on a practical aspect---the inevitable lag time between presenting learned rules to the user and obtaining feedback from the user---has been made. Results on the OPF problem have shown that the proposed framework is able to maintain similar performance up to a certain lag period, beyond which the user response has been found to be too slow for the algorithm to maintain the same level of performance. 

This study opens up a number of avenues for future work. In this work, the learning and repair intervals are kept fixed. The effect of these parameters need to be studied more closely. In many problems, a rule may not stay valid across the entire Pareto-optimal front. Locally present rules may exist in certain parts of the Pareto-optimal front. Ways to extract local rules and repair a MOEA's offspring population members accordingly will introduce additional challenges but may result in faster convergence. 
Traditional user preference information including, but not limited to, relative importance of objective functions and preferred regions of the Pareto-optimal front, can also potentially be integrated into this type of framework. Lastly, the asynchronous user interaction study is practical and must be investigated more thoroughly. Nevertheless, this study has clearly demonstrated a viable way to extract variable interaction knowledge from intermediate optimization iterations and to use relevant and vetted knowledge back in the optimization algorithm for updating offspring solutions to constitute a computationally fast search process. More such practice-oriented studies must now accompany evolutionary optimization applications to make them more worthy for practical problem solving tasks.  



\bibliographystyle{IEEEtran}
\bibliography{./references/library}

\end{document}


\title{Supplementary Document of \\ An Interactive Knowledge-based Multi-objective Evolutionary Algorithm Framework for Practical Optimization Problems}


\author{Abhiroop~Ghosh,
	        Kalyanmoy~Deb,
	        Erik~Goodman, and
	        Ronald~Averill
	\thanks{Authors are with Michigan State University, East Lansing, MI 48824, USA, e-mail: \{ghoshab1, kdeb, goodman, averillr\}@msu.edu (see https://www.coin-lab.org).}}


\maketitle
\begin{abstract}
Additional supporting materials to the main paper are provided in this supplementary document. These materials presents additional results illustrating the main concepts presented in the paper. 
\end{abstract}

\begin{IEEEkeywords}
Multi-objective optimization, `innovization', knowledge-based optimization.
\end{IEEEkeywords}

\section{Simply-supporterd stepped beam design}
\begin{figure*}[htb]
	\centering
	\begin{subfigure}{0.48\textwidth}
		\includegraphics[width=\columnwidth]{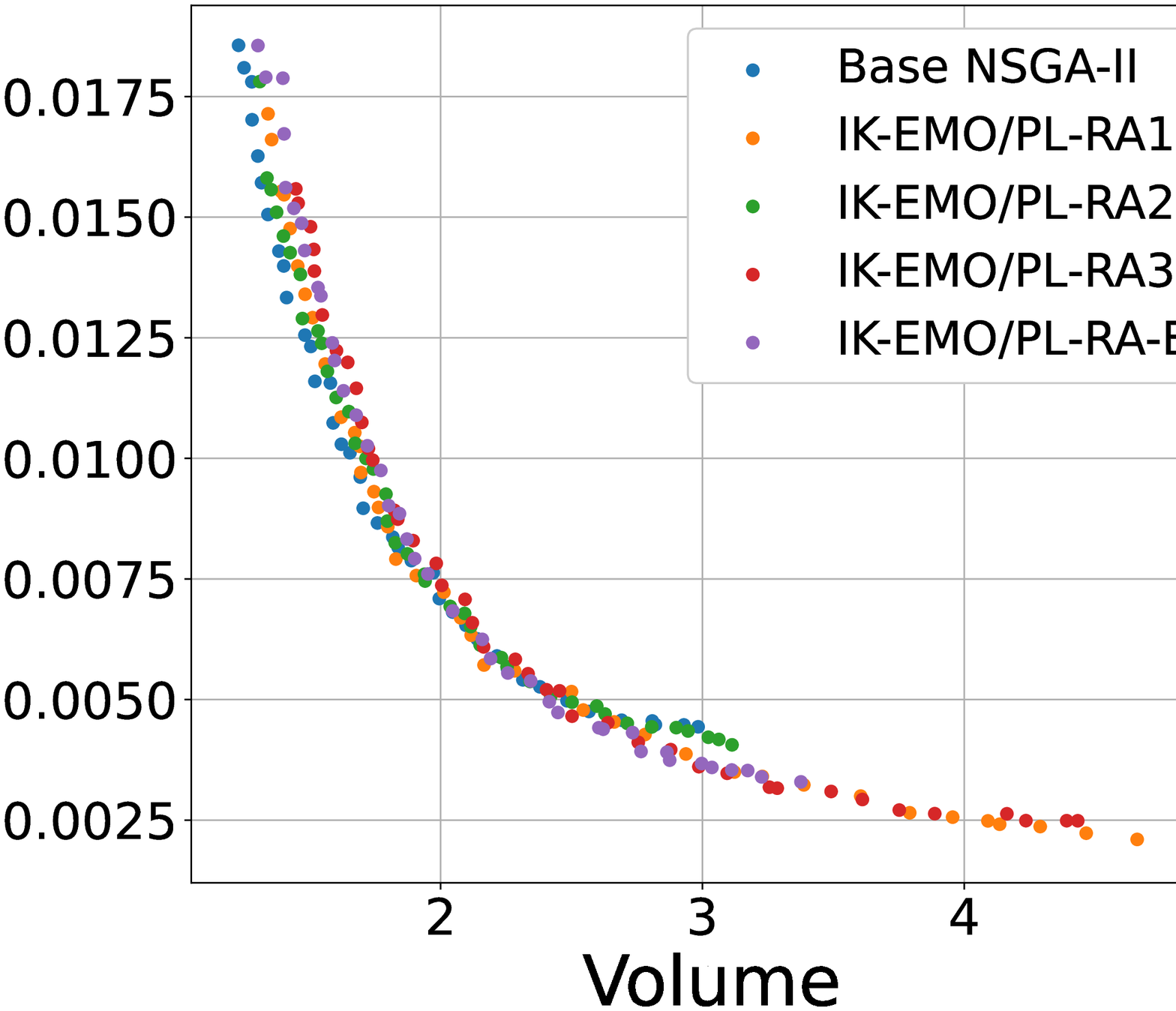}
		\caption{Pareto Front for one run.}
		\label{fig:pfSB39}
	\end{subfigure}
	\begin{subfigure}{0.48\textwidth}
		\includegraphics[width=\columnwidth]{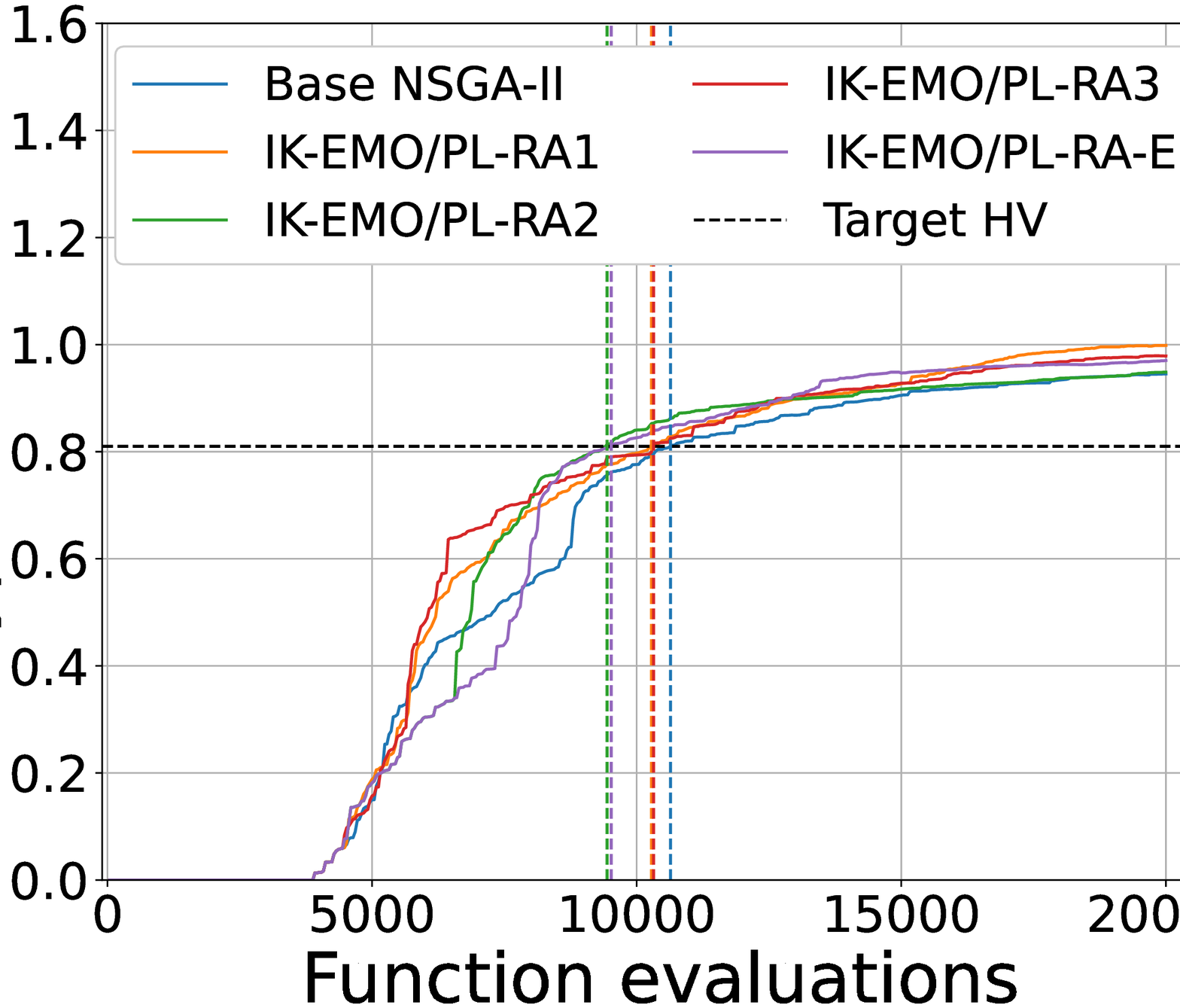}
		\caption{HV plot over 20 runs.}
		\label{fig:hvSB39}
	\end{subfigure}
	\caption{ND fronts and hypervolume plots obtained by \IKEMO\ with RU2 and power law repair agents for 
	39-segment stepped beam problem.}
	\label{fig:sb39results}
\end{figure*}

This section presents some additional details on the results presented in the main document for the simply-supporterd stepped beam design problem. Figure \ref{fig:sb39results} shows the Pareto Front for one run and the median HV plot over 20 runs for the 39-segment stepped beam problem.
	
\section{Optimal Power Flow problem}
\begin{figure*}[htb]
	\centering
	\begin{subfigure}{0.48\textwidth}
		\includegraphics[width=\columnwidth]{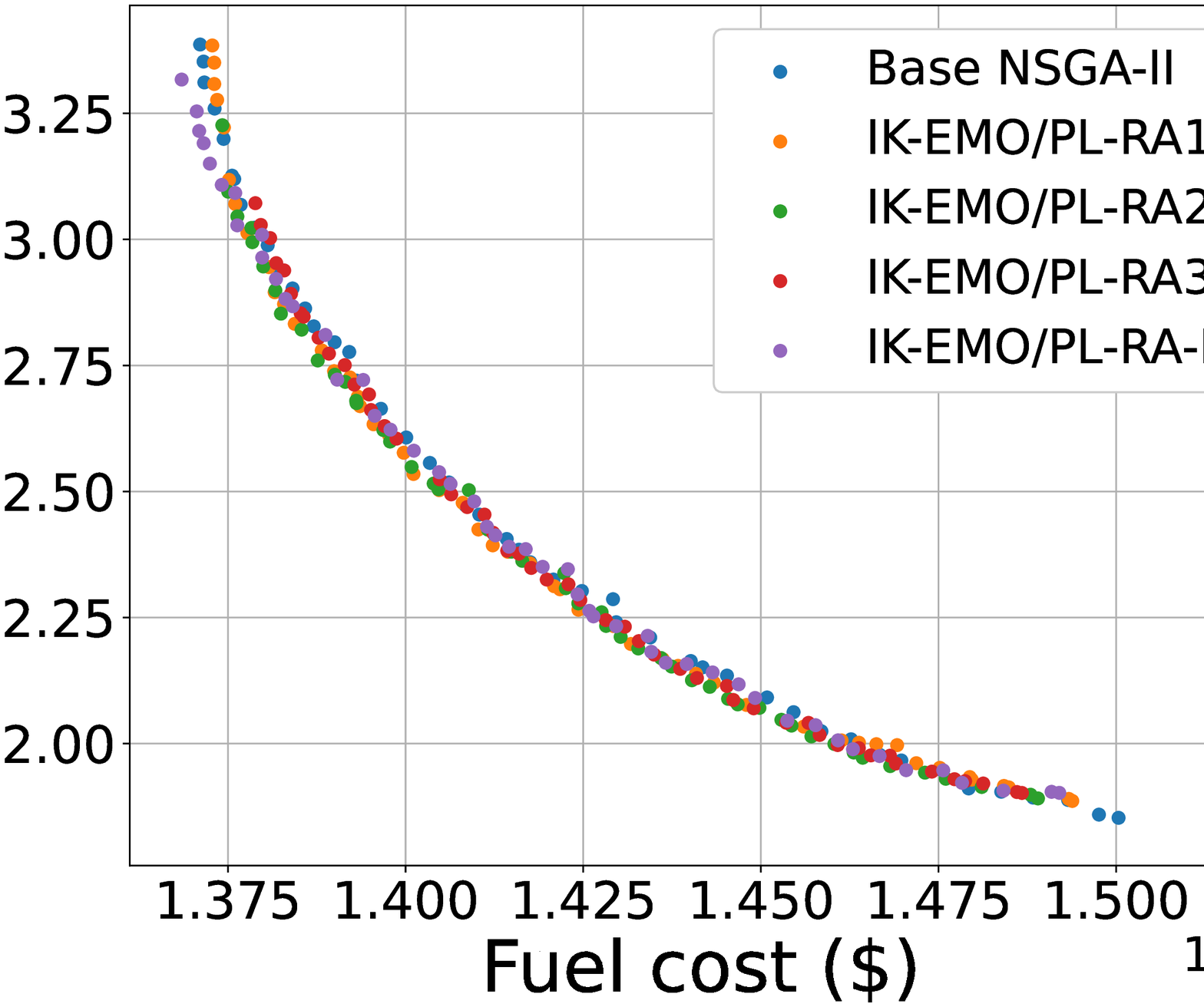}
		\caption{ND Front for one run.}
		\label{fig:pfOpf118}
	\end{subfigure}
	\begin{subfigure}{0.48\textwidth}
		\includegraphics[width=\columnwidth]{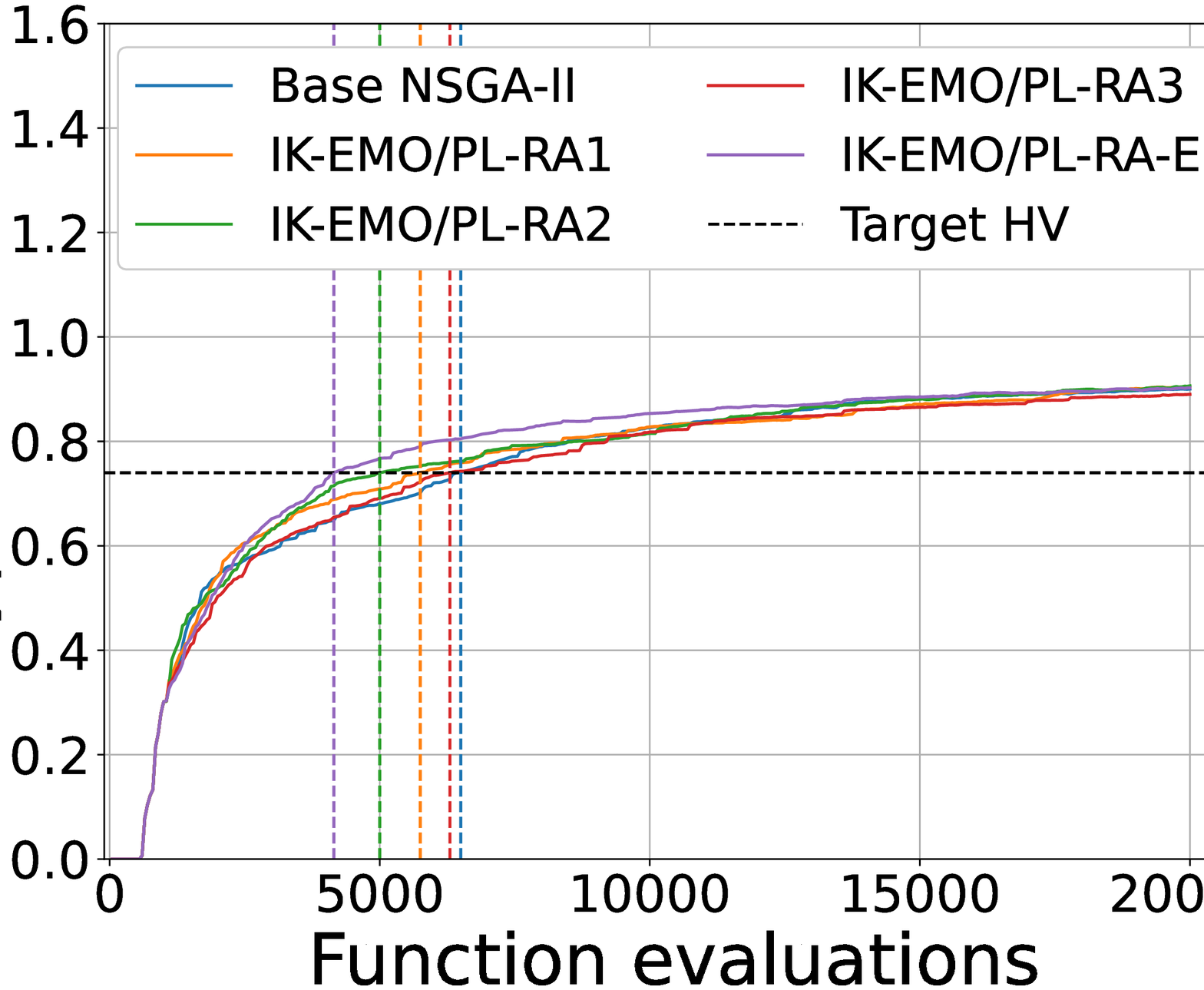}
		\caption{HV plot over 20 runs.}
		\label{fig:hvOpf118}
	\end{subfigure}
	\caption{ND front and hypervolume plots obtained by \IKEMO\ with RU2 and power law repair agents for IEEE 118-bus OPF problem.}
	\label{fig:opf118results}
\end{figure*}

This section presents some additional details on the results presented in the main document for the optimal power flow (OPF) problem. Figure \ref{fig:opf118results} shows the Pareto Front for one run and the median HV plot over 20 runs for the 118-bus OPF problem.

\section{Truss Design Problem}
\label{sec:truss_problem}
\subsection{Problem background and formulation}
\begin{figure}[hbt]
	\centering
	\includegraphics[width=\columnwidth,trim=100mm 25mm 100mm 25mm,clip]{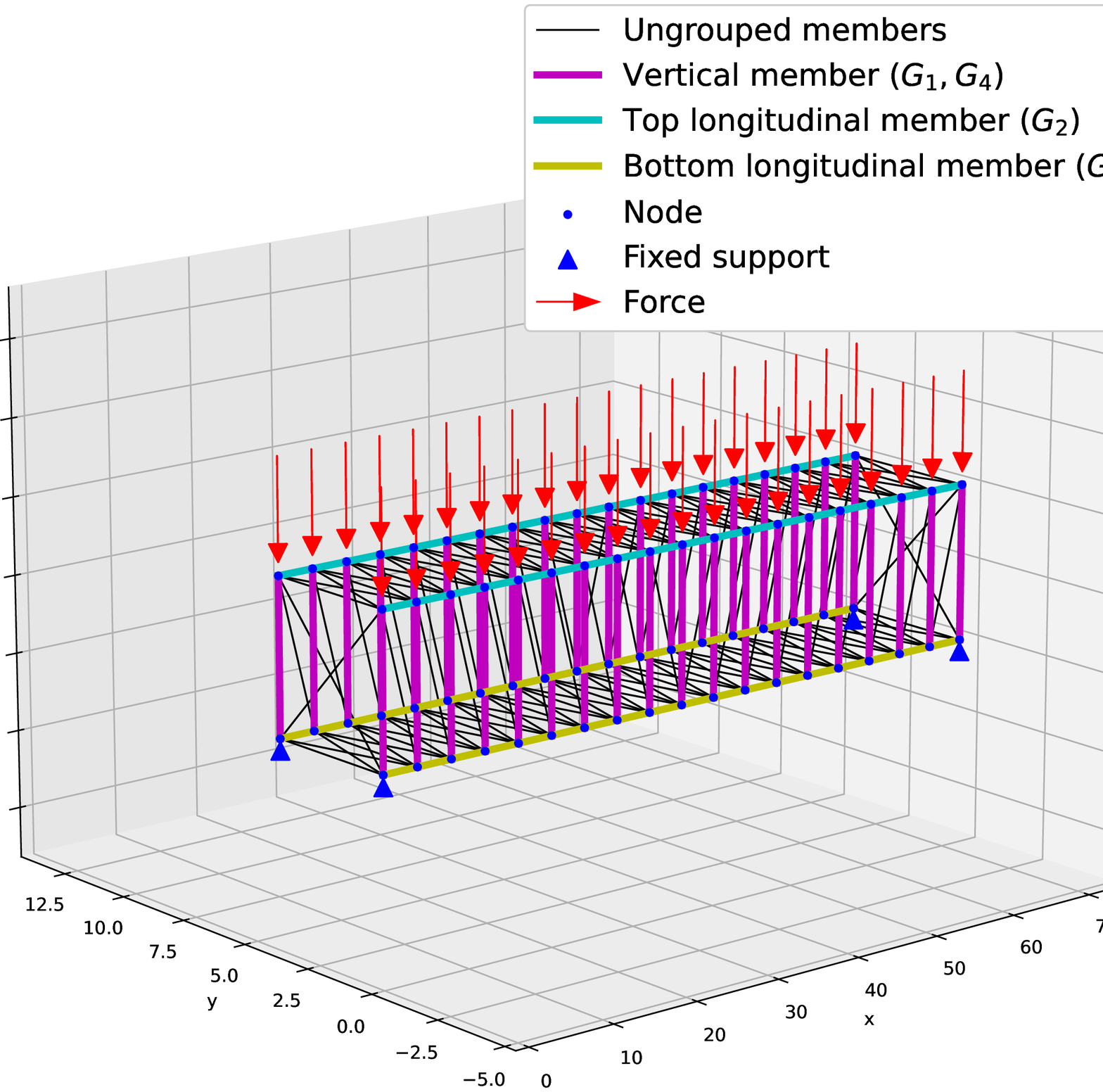}
	\caption{260-member truss with grouped variables.}
	\label{fig:groupedTruss}
\end{figure}

A number of optimization algorithms to solve a truss design problem can be found in the literature \cite{Ahrari2015,Ahrari2020,Bandyopadhyay2015}. In this paper, a 3D truss design problem has been created based on the formulation provided in \cite{Ghosh2021}. A bi-objective optimization problem can be defined with the objectives being: (a) minimize weight, and (b) minimize compliance. There are two types of design variables: size variables, defined by member radii ($\mathbf{r}$), and shape variables, defined by length of the vertical members ($\mathbf{L}_v$) parallel to the $z$-axis. This problem can be scaled by adjusting the number of shape variables ($n_s$). Two constraints are to be satisfied: (a) stress in each member should be less than 200 MPa, (b) displacement of each node should be less than 20 mm. The problem formulation is shown below:

\begin{align}
\label{eq:optim_obj}
\text{Minimize}\quad & f_{1}(\mathbf{r}, \mathbf{L}_v)   = \rho \sum_{i=1}^{n_m}
V_i(\mathbf{r}, \mathbf{L}_v), \\
\text{Minimize}\quad &  f_{2}(\mathbf{r}, \mathbf{L}_v) = \  \textbf{F}(\mathbf{r}, \mathbf{L}_v) \cdot \textbf{D}(\mathbf{r}, \mathbf{L}_v),\\
\label{eq:stress_constr}
\text{Subject to}\quad & \sigma_i(\mathbf{r}, \mathbf{L}_v) \leq S, \ \mbox{ for } i=1,2\ldots,n_m, \\
\label{eq:displacement_constr}
& |D_j(\mathbf{r}, \mathbf{L}_v)| \leq \delta, \ \text{ for }  j = 1, 2, \ldots, n_n,\\
& r^L \leq r_i \leq r^U, \mbox{ for } i=1,2\ldots,n_m, \\
& l^L \leq l_j \leq l^U, \mbox{ for } j=1,2\ldots,n_n,
\end{align}

where $V_i$ is the volume of the $i^{th}$ member, $\rho$ is the material density (7,000 kg/m$^3$), $n_m$ is the total number of members, $n_n$ is the total number of nodes, $\mathbf{F}$ and $\mathbf{D}$ are vectors consisting of the forces in members and displacements at nodes, respectively. $\sigma_i$ is the stress developed in the $i$-th member of the truss and is restricted to material strength $S=200$ MPa. Radius and length variables are restricted to [5, 300] mm and [4, 30] m, respectively. The absolute deflection of nodes is restricted to $\delta=20$ mm. While $f_1$ is relatively easy to compute, $f_2$ and two constraint evaluations require a finite element simulation, making the process computationally expensive. For this paper, we consider an 1100-member truss with 1179 decision variables. A large number of decision variables makes this problem a large scale multi-objective optimization problem (LSMOP). Figure \ref{fig:groupedTruss} shows a scaled-down version of the truss with 260 members.

%

\subsection{Experimental results and discussion}
\begin{figure*}[htb]
\centering
\begin{subfigure}{0.48\textwidth}
	\includegraphics[width=\columnwidth]{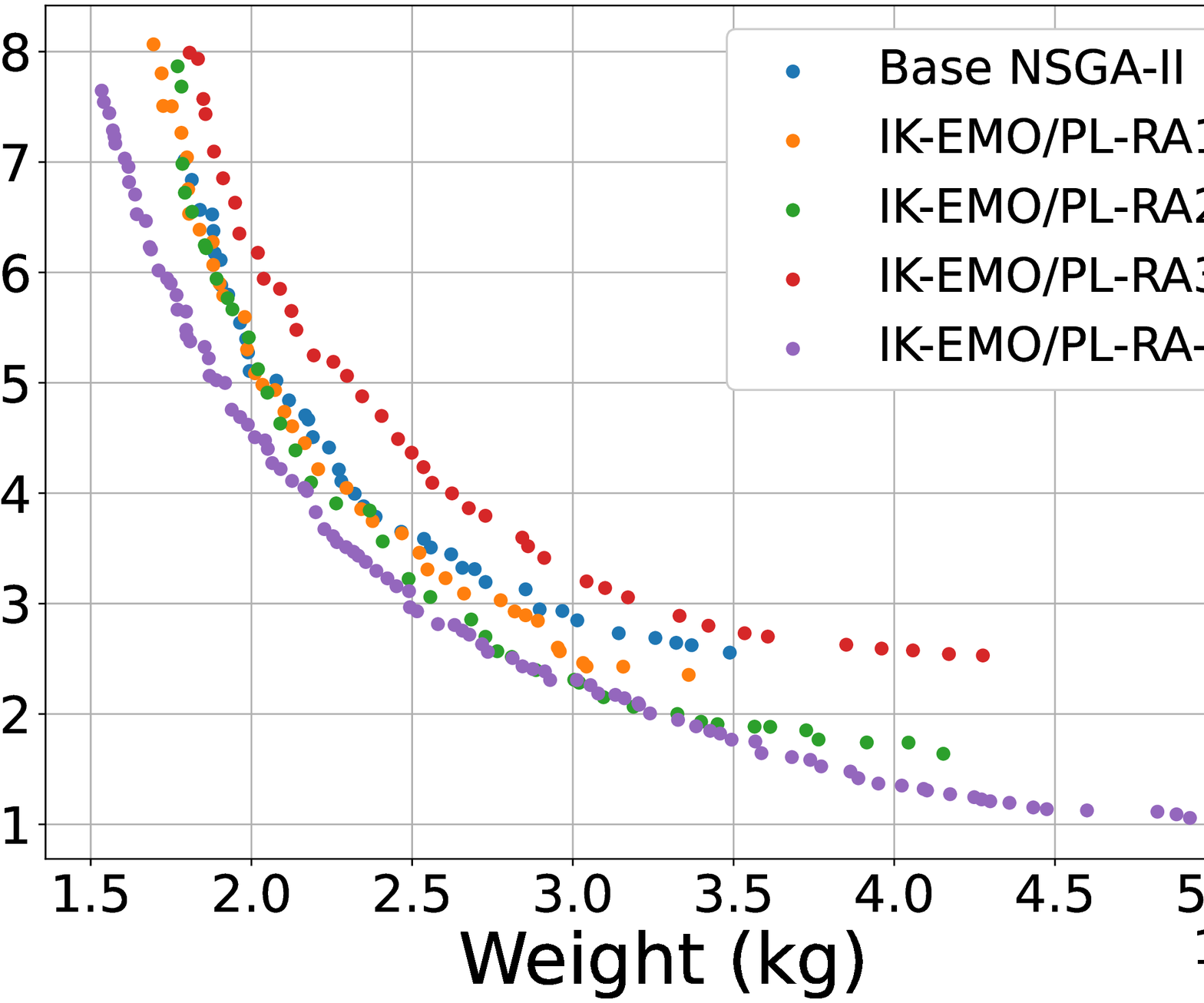}
	\caption{Pareto Front for one run.}
	\label{fig:pfTruss1100}
\end{subfigure}
\begin{subfigure}{0.48\textwidth}
	\includegraphics[width=\columnwidth]{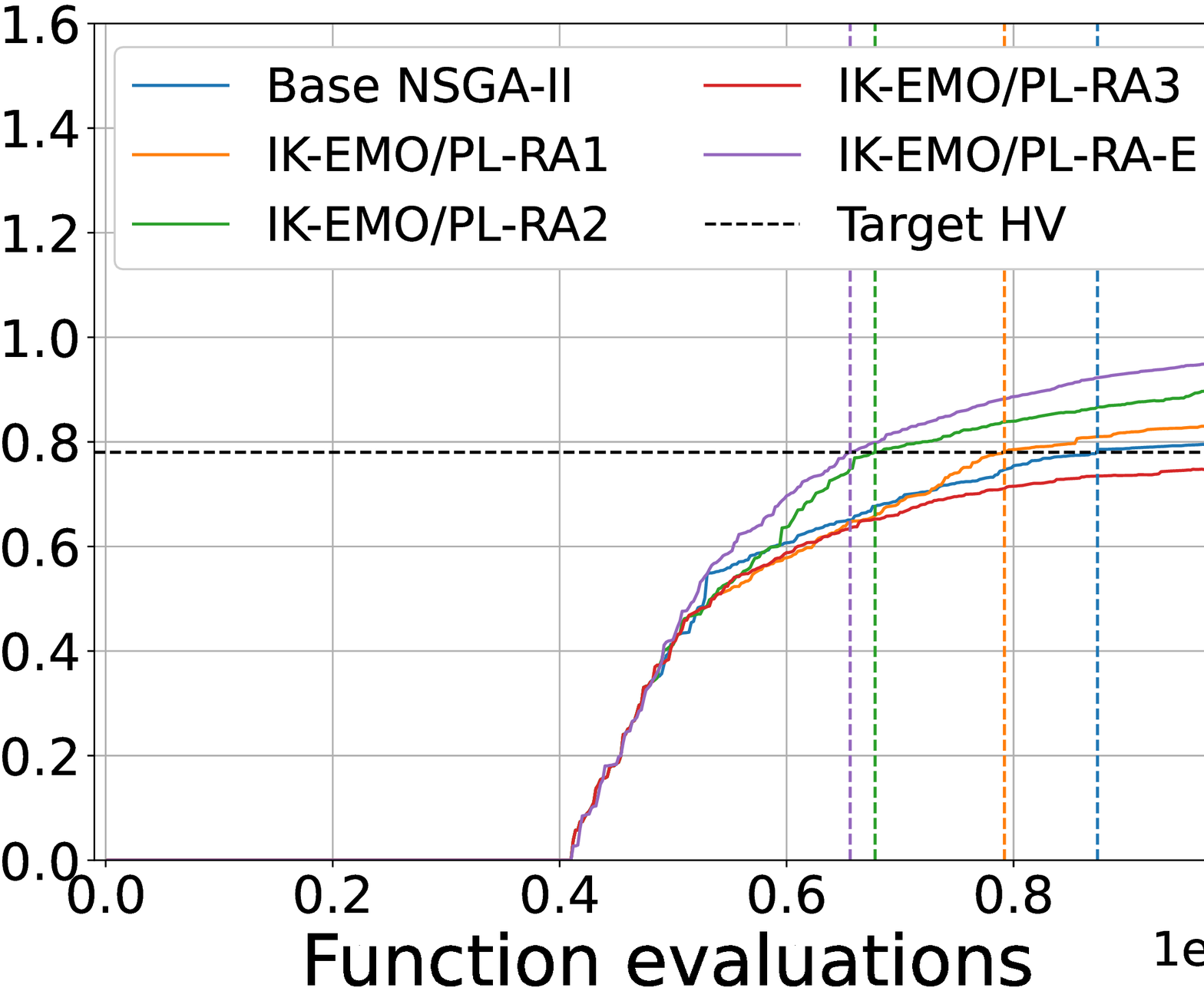}
	\caption{HV plot over 20 runs.}
	\label{fig:hvTruss1100}
\end{subfigure}
\caption{Pareto fronts and hypervolume plots obtained by \IKEMO\ with RU2 and power law repair agents for 1100-member truss design problem.}
\label{fig:Truss1100}
\end{figure*}

Results on the 1100-member truss problem is given in Table \ref{tab:TrussResults1380}. The Pareto Fronts obtained in 1 run using the power law repair operators for RU2 are shown in Figure \ref{fig:pfTruss1100}. The corresponding median HV plots over the course of the optimization are shown in Figure \ref{fig:hvTruss1100}.
\begin{table*}[hbt]
	\centering
	\caption{FEs required to achieve HV$\protect^T$ = 0.78 for 1100-member truss. Best performing algorithm for row is marked in bold. Best performing algorithm in each column is marked by a shaded gray box. Algorithms with statistically similar performance to the best algorithm column-wise are marked in italics. The corresponding Wilcoxon p-values are given in braces.}
	\begin{tabular}{|c|c|c|c|c|c|}
		\hline
		Rule Type & & RU1 & RU2 & RU3 & RU4 \\
		\hline
		None 						& Base &  874k $\pm$ 10k (p = 0.0043)  & 874k $\pm$ 10k (p = 0.0017) &  874k $\pm$ 10k (p = 0.0062)  &  874k $\pm$ 10k (p = 0.0053) \\
		\hline
		\hline
		\multirow{4}{*}{Power law}  & PL-RA1 & 802k $\pm$ 8k (p=0.0057) & 792k $\pm$ 8k (p=0.0129) & \textbf{786k $\pm$ 13k} (p=0.0140) & 812k $\pm$ 6k (p=0.0023)\\
									& PL-RA2 & \textit{680k $\pm$ 11k} (p=0.0633) & \textbf{\textit{678k $\pm$ 15k}} (p=0.0793) & \cellcolor{gray!25}688k $\pm$ 10k  & \cellcolor{gray!25}744k $\pm$ 17k \\
									& PL-RA3 & \textbf{963k $\pm$ 21k} (p=0.0005) & 1M (HV=0.74) & 1M (HV=0.71)  & 1M (HV=0.66) \\
									& PL-RA-E & \cellcolor{gray!25}672k $\pm$ 9k  & \cellcolor{gray!25}\textbf{656k $\pm$ 18k} & \textit{693k $\pm$ 16k} (p=0.1016) & \textit{754k $\pm$ 24k} (p=0.1163)\\
		\hline
		\hline
		\multirow{4}{*}{Inequality rule} & IQ-RA1 & 843k $\pm$ 15k (p=0.0015) & 828k $\pm$ 20k (p=0.0115) & \textbf{822k $\pm$ 23k} (p=0.0169) & 851k $\pm$ 8k (p=0.0325)\\
										& IQ-RA2 & 836k $\pm$ 12k (p=0.0039) & 838k $\pm$ 19k (p=0.0248) &  \textbf{803k $\pm$ 18k} (p=0.0465) & 834k $\pm$ 6k (p=0.0318)\\
										& IQ-RA3 & 839k $\pm$ 21k (p=0.0036) & \textbf{819k $\pm$ 29k} (p=0.0219) & 826k $\pm$ 10k (p=0.0454) & 837k $\pm$ 5k (p=0.0414)\\
										& IQ-RA-E &  840k $\pm$ 27k (p=0.0024) & 816k $\pm$ 22k (p=0.0351) & \textbf{798k $\pm$ 13k} (p=0.0311) & 843k $\pm$ 11k (p=0.0223)\\
		\hline
		\hline
		\multirow{2}{*}{Mixed} & PL-RA2 + IQ-RA2 & \textit{682k $\pm$ 10k} (p=0.0669) & \textbf{\textit{677k $\pm$ 14k}} (p=0.0816) & \textit{691k $\pm$ 8k} (p=0.0772) & \textit{751k $\pm$ 10k} (p=0.0522)\\
							   & PL-RA-E + IQ-RA-E & \textit{676k $\pm$ 9k} (p=0.0714) & \textit{\textbf{659k $\pm$ 12k}} (p=0.0814) & \textit{696k $\pm$ 13k} (p=0.0611) &\textit{752k $\pm$ 23k} (p=0.0699)\\
		\hline
	\end{tabular}
	\label{tab:TrussResults1380}
\end{table*}

The observations from the results are outlined below.

\subsubsection{General observations}
Base NSGA-II is outperformed by all repair operators except for PL-RA3, which fails to achieve the target HV in most cases. This shows PL-RA3 being a subpar repair operator introducing too much uncertainty into the optimization process compared to the others. However, despite the low population size (almost 1/10th the number of variables), the repair operators are able to generate better solutions on an LSMOP. FE savings upto 25\% compared to base NSGA-II is obtained.

\subsubsection{Power law vs inequality rules}
Similar to the previous two problems, power law repair operators are the best performer for each RU. Inequality rule-based repair operators in general result in a worse performance compared to power law-based repair operators except for the PL-RA3 case. These results show that power laws are in general a better choice compared to inequality rules.

\subsubsection{Best performing algorithm for each RU}
PL-RA-E is the best performer for RU1 and RU2, with PL-RA2 showing a comparable performance. For RU3 and RU4, PL-RA2 is the best performer, with PL-RA-E showing a comparable performance. PL-RA3 fails to achieve the target. For RU4, PL-RA1 is the best, with PL-RA2 and PL-RA-E giving statistically similar performance. As in the case of the previous problem, PL-RA-E is either the best or gives a statistically similar performance to the best. PL-RA2 performance shows that an intermediate amount of trust in the learned power law is the optimal choice.

Figure \ref{fig:Truss1100} illustrates the results with RU2 combined with the power law repair operators for both problem cases. The difference in the quality of solutions obtained after 1 million FEs is more prominent than the previous two problems due to this being an LSMOP. PL-RA-E gives the best solutions are PL-RA3 is significantly worse. In the median HV plots it is seen that the FEs required to reach HV$^T$ for PL-RA-E is close to PL-RA2. Base NSGA-II, PL-RA2 and PL-RA3 show a worse performance than the others.

\subsubsection{Relative performance of each RU}
It can be seen from Table \ref{tab:TrussResults1380}  that RU2 gives the best performance in 5 out of 10 cases and RU3 gives the best performance in 4 out of 10 cases. This reinforces the previous results by showing that a moderate amount of rule usage gives better performance.

\subsubsection{Mixed relation repair agents}
All of the mixed relation repair agents (PL-RA2+IQ-RA2) and (PL-RA-E+IQ-RA-E), have a comparable performance to the best algorithm for each RU. Similar to previous problems, the worse performance of PL-RA3 and inequality repair operators do not affect the performance.

\section{Summary of results}
For every problem we have a total of 11 different algorithms including base NSGA-II and 4 RUs. Table \ref{tab:overallRank} presents a ranking of all the algorithms for each RU based on the FEs required to reach the target HV. An algorithm with statistically similar performances to the best is given a rank of 1. The results are collected and an overall ranking is assigned to each algorithm. It is seen that the top ranked algorithm is PL-RA-E followed by PL-RA2+IQ-RA2, PL-RA-E+IQ-RA-E and PL-RA2. If the ideal repair agent or rule types are not known beforehand, the ensemble operator with mixed relationships (PL-RA-E+IQ-RA-E) can be used without significant performance drop.
\begin{table*}[!hbt]
	\renewcommand{\arraystretch}{1.2}
	\centering
	\caption{Ranking of different repair agents on multiple problems.}
	\label{tab:overallRank}
	\begin{tabular}{|l@{\hspace{2pt}}|p{1cm}@{\hspace{2pt}}|p{1cm}@{\hspace{2pt}}|p{1cm}@{\hspace{2pt}}|p{1cm}@{\hspace{2pt}}|p{1.1cm}@{\hspace{2pt}}|p{1cm}@{\hspace{2pt}}|p{1cm}@{\hspace{2pt}}|p{1cm}@{\hspace{2pt}}|p{1.05cm}@{\hspace{2pt}}|p{1.15cm}@{\hspace{2pt}}|p{1.25cm}|} \hline
		RU & Base & PL-RA1 & PL-RA2 & PL-RA3 & PL-RA-E & IQ-RA1 & IQ-RA2 & IQ-RA3 & IQ-RA-E & PL-RA2 +IQ-RA2 & PL-RA-E +IQ-RA-E \\ \hline
		\multicolumn{12}{|c|}{78-variable stepped beam design} \\\hline
			RU1 & 4 & 3 & 1 & 3 & 1 & 5 & 2 & 5 & 4 & 1 & 1 \\
			RU2 & 4 & 2 & 1 & 2 & 1 & 3 & 2 & 4 & 3 & 1 & 1 \\
			RU3 & 5 & 4 & 1 & 2 & 1 & 5 & 3 & 4 & 5 & 1 & 1 \\
			RU4 & 2 & 2 & 1 & 5 & 1 & 4 & 3 & 2 & 2 & 1 & 1 \\\hline
		Rank Sum & 15 & 11 & 4 & 12 & 4 & 17 & 10 & 15 & 14 & 4 & 4 \\\hline
		Rank 	 & 9 & 6 & \textbf{1} & 7 & \textbf{1} & 11 & 5 & 9 & 8 & \textbf{1} & \textbf{1} \\\hline
		\multicolumn{12}{|c|}{118-variable stepped beam design} \\\hline
		RU1 & 7 & 2 & 1 & 3 & 1 & 5 & 6 & 4 & 5 & 1 & 1\\
		RU2 & 8 & 1 & 2 & 3 & 1 & 5 & 4 & 7 & 6 & 1 & 1 \\
		RU3 & 7 & 2 & 1 & 3 & 1 & 3 & 4 & 6 & 5 & 1 & 1 \\
		RU4 & 9 & 3 & 2 & 4 & 1 & 8 & 5 & 7 & 6 & 1 & 1 \\\hline
		Rank Sum & 31 & 8 & 6 & 13 & 4 & 21 & 19 & 24 & 22 & 4 & 4 \\\hline
		Rank & 11 & 5 & 4 & 6 & \textbf{1} & 8 & 7 & 10 & 9 & \textbf{1} & \textbf{1} \\\hline
		\multicolumn{12}{|c|}{115-variable optimal power flow} \\\hline
		RU1 & 3 & 2 & 1 & 5 & 1 & 4 & 4 & 3 & 3 & 1 & 1 \\
		RU2 & 5 & 2 & 1 & 3 & 1 & 7 & 6 & 4 & 3 & 1 & 1 \\
		RU3 & 7 & 3 & 1 & 5 & 1 & 4 & 6 & 4 & 8 & 1 & 2 \\
		RU4 & 2 & 1 & 1 & 4 & 1 & 4 & 3 & 5 & 4 & 1 & 1 \\\hline
		Rank Sum & 17 & 8 & 4 & 17 & 4 & 19 & 19 & 16 & 18 & 4 & 5 \\\hline
		Rank & 7 & 5 & \textbf{1} & 7 & \textbf{1} & 10 & 10 & 6 & 9 & \textbf{1} & 4 \\\hline
		\multicolumn{12}{|c|}{243-variable optimal power flow} \\\hline
		RU1 & 7 & 1 & 2 & 8 & 1 & 3 & 6 & 5 & 4 & 1 & 1 \\
		RU2 & 7 & 2 & 1 & 7 & 1 & 3 & 5 & 4 & 6 & 1 & 1 \\
		RU3 & 7 & 4 & 3 & 10 & 1 & 9 & 8 & 6 & 2 & 5 & 1 \\
		RU4 & 3 & 2 & 1 & 7 & 1 & 5 & 4 & 8 & 6 & 1 & 1 \\\hline
		Rank Sum & 24 & 9 & 7 & 32 & 4 & 20 & 23 & 23 & 18 & 8 & 4 \\\hline
		Rank & 10 & 5 & 3 & 11 & \textbf{1} & 7 & 8 & 8 & 6 & 4 & \textbf{1} \\\hline
		\multicolumn{12}{|c|}{1179-variable truss design} \\\hline
		RU1 & 7 & 2 & 1 & 8 & 1 & 6 & 3 & 4 & 5 & 1 & 1 \\
		RU2 & 7 & 2 & 1 & 8 & 1 & 5 & 6 & 4 & 3 & 1 & 1 \\
		RU3 & 7 & 2 & 1 & 8 & 1 & 5 & 4 & 6 & 3 & 1 & 1 \\
		RU4 & 7 & 2 & 1 & 8 & 1 & 6 & 3 & 4 & 5 & 1 & 1 \\\hline
		Rank Sum & 28 & 8 & 4 & 32 & 4 & 22 & 16 & 18 & 16 & 4 & 4 \\\hline
		Rank & 10 & 5 & \textbf{1} & 11 & \textbf{1} & 9 & 6 & 8 & 6 & \textbf{1} & \textbf{1} \\\hline\hline
		Final Rank-Sum & 47 & 26 & 10 & 42 & 5 & 45 & 36 & 41 & 38 & 8 & 8 \\\hline
		Final Rank & 11 & 5 & 4 & 9 & \textbf{1} & 10 & 6 & 8 & 7 & 2 & 2 \\\hline
	\end{tabular}
\end{table*}

\section{Synchronous vs asynchronous user interaction}
\begin{figure*}[htb]
	\centering
	\begin{subfigure}{0.8\textwidth}
		\includegraphics[width=\columnwidth]{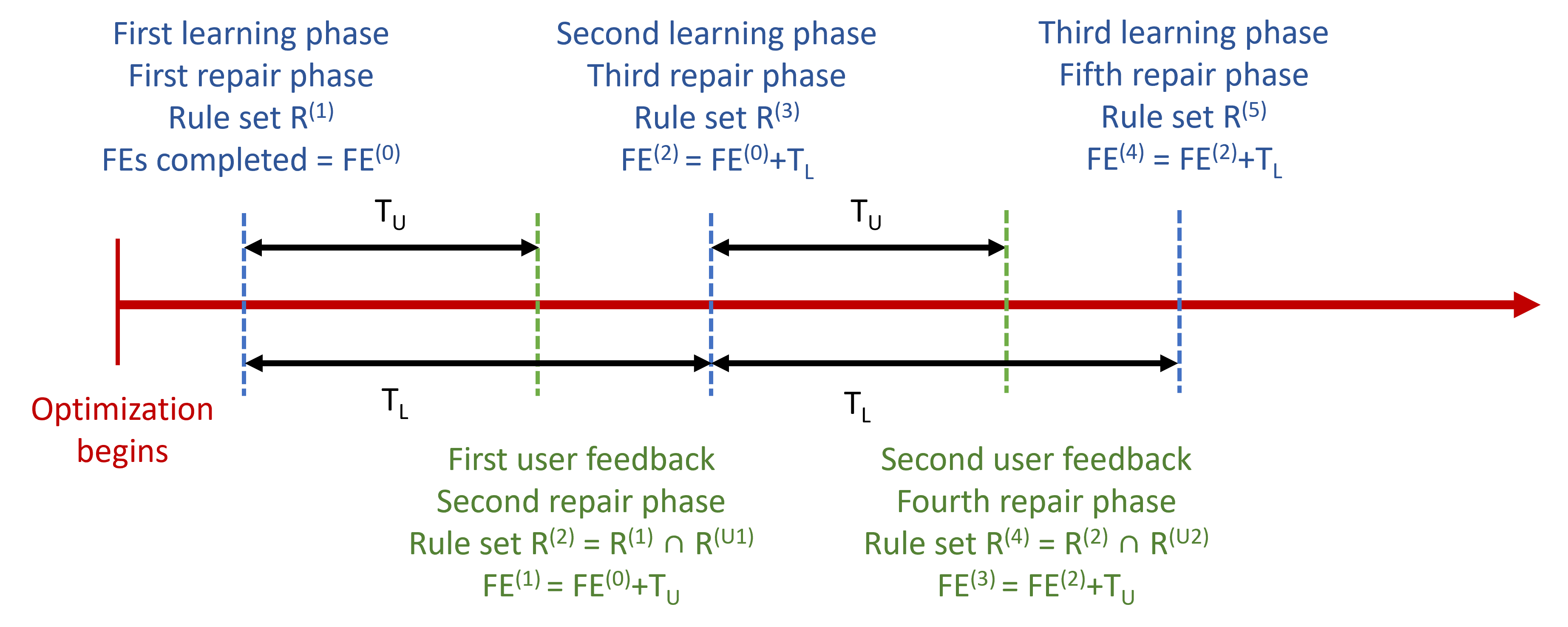}
		\caption{$T_U \leq T_L$.}
		\label{fig:asyncTULessThanTL}
	\end{subfigure}
	\begin{subfigure}{0.8\textwidth}
		\includegraphics[width=\columnwidth]{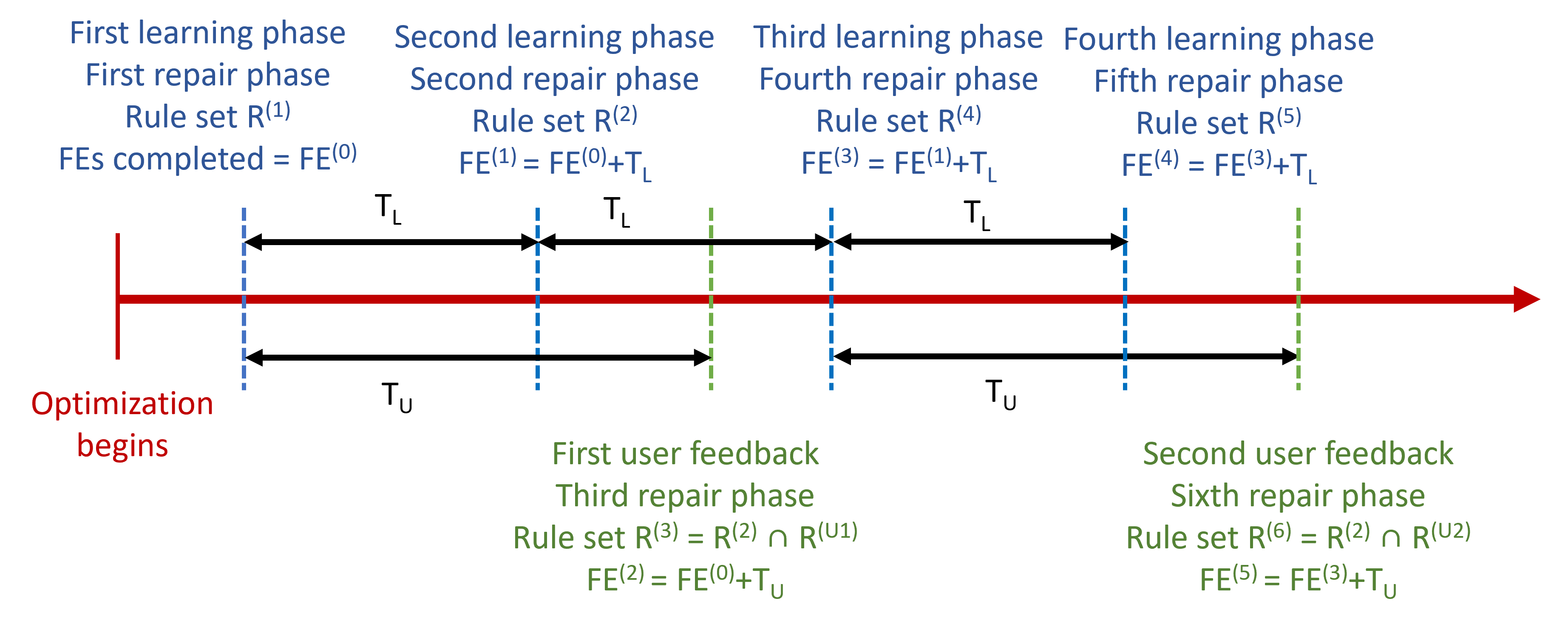}
		\caption{$T_U > T_L$.}
		\label{fig:asyncTUGreaterThanTL}
	\end{subfigure}
	\caption{Learning, repair and user feedback operations over the optimization run. Blue dotted lines represent the learning phase and the associated repair phase. At every learning phase taking place at intervals of $T_L$ FEs, a rule set (R) is sent to the user if he/she is available. Green dotted lines represent when user feedback is complete.}
	\label{fig:asyncTUTL}
\end{figure*}

This section presents some additional information about the synchronous vs asynchronous user interaction study. Figures \ref{fig:asyncTULessThanTL} and \ref{fig:asyncTUGreaterThanTL} illustrate the learning, repair and user feedback operations over the optimization run for the asynchronous case where $T_U \leq T_L$ and $T_U > T_L$, respectively. Here we always perform a repair whenever a learning phase is complete, or when the user provides a feedback. In the latter case, \IKEMO\ combines the learned rules and the user feedback as described in the main document. This new rule set is used to perform repairs. For the case where $T_U \leq T_L$, no new rules are learned between the time the user is given the rules and when the user is ready with feedback. In the case of $T_U > T_L$, the algorithm may learn new rules multiple times and use it to perform repairs while the user is processing the rules.

\begin{figure*}[htb]
	\centering
	\begin{subfigure}{0.48\textwidth}
		\includegraphics[width=\columnwidth]{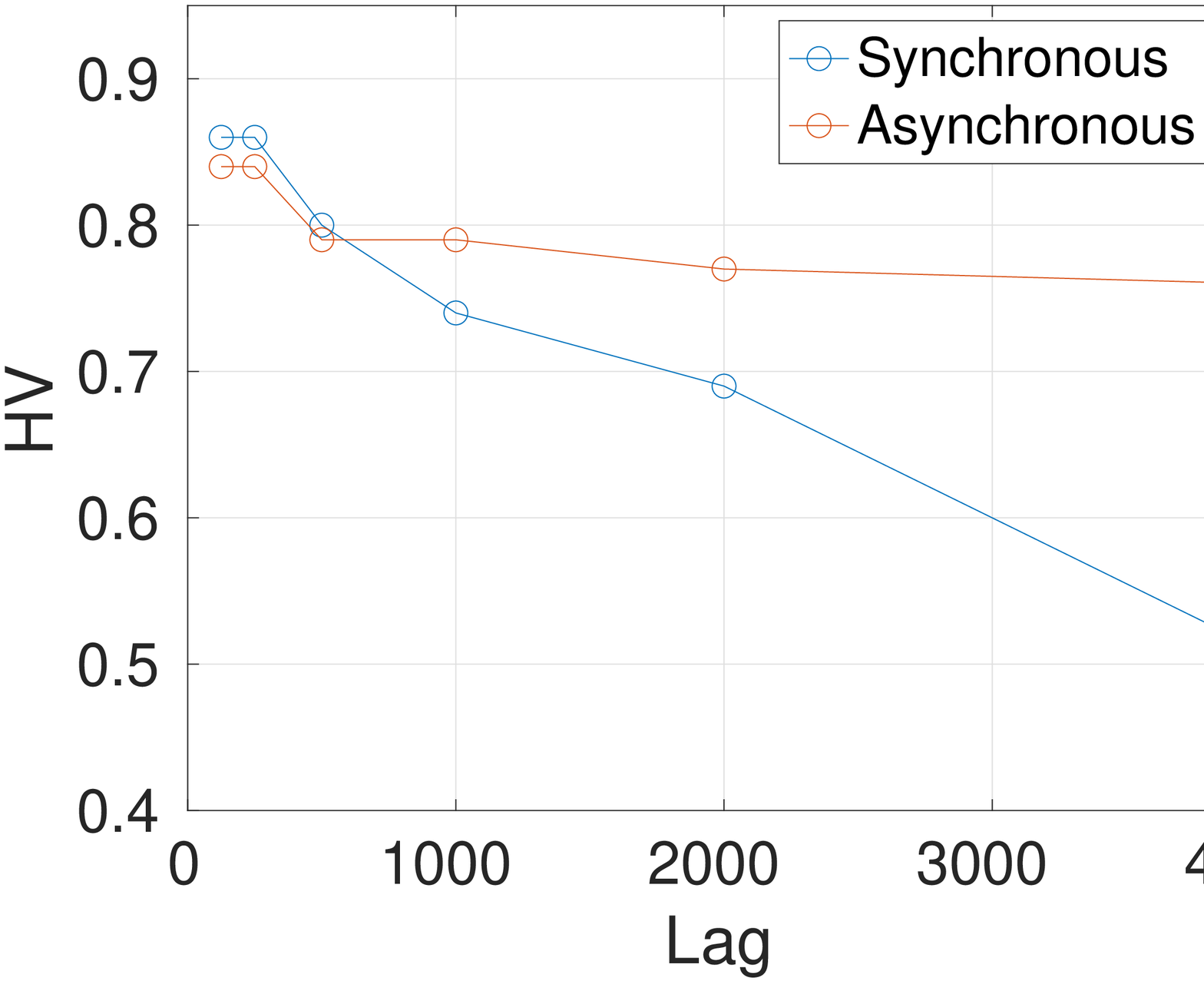}
		\caption{IEEE 118-bus OPF problem}
		\label{fig:syncvsasynchv118bus}
	\end{subfigure}
	\begin{subfigure}{0.48\textwidth}
		\includegraphics[width=\columnwidth]{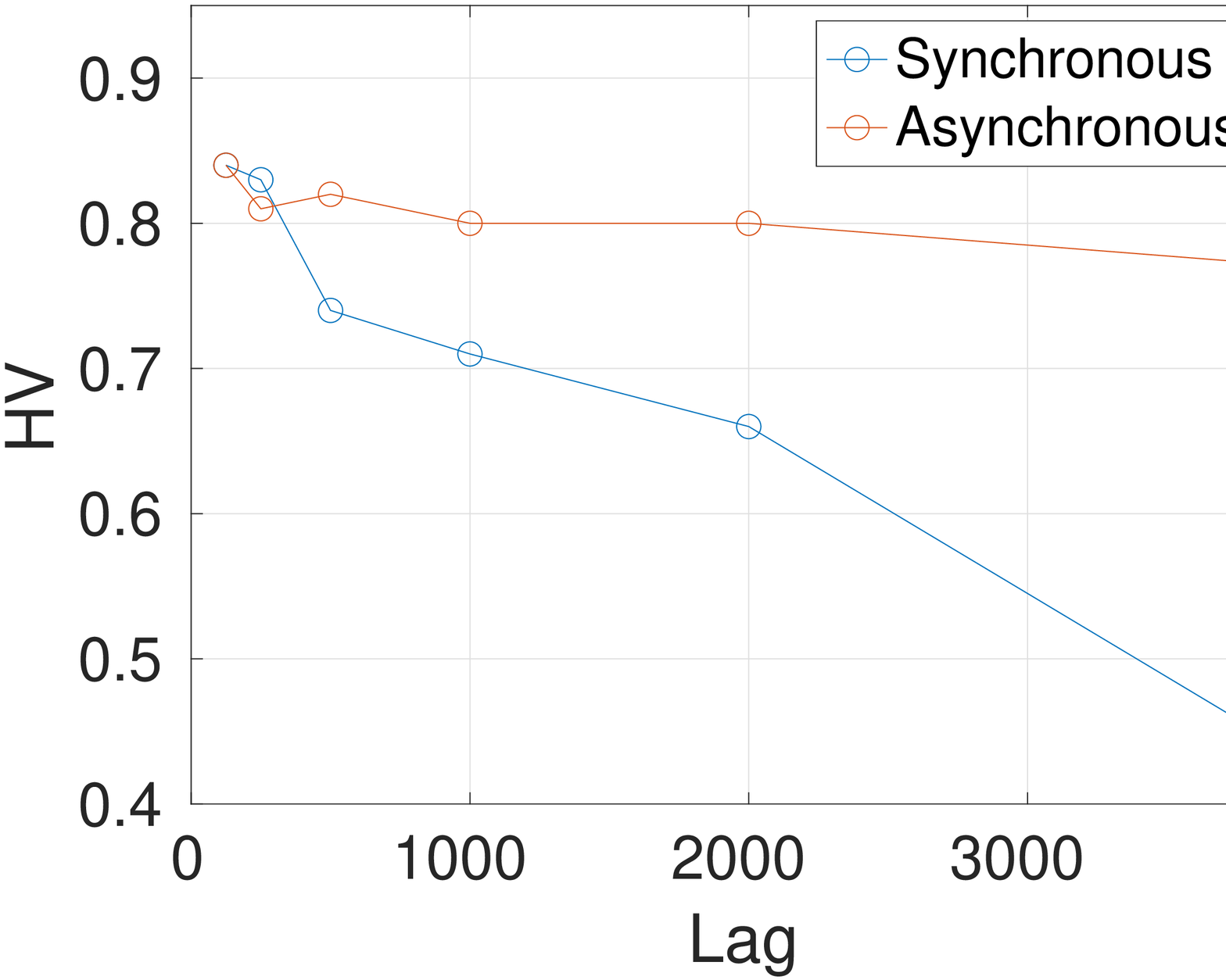}
		\caption{IEEE 300-bus OPF problem}
		\label{fig:syncvsasynchv300bus}
	\end{subfigure}
	\caption{Variation of final median HV over 20 runs with different lag values (in terms of time for executing different number of FEs) for the synchronous and asynchronous cases on the 118-bus and 300-bus OPF problems. For large $T_U$, asynchronous implementation is better.}
	\label{fig:syncVsAsyncHVOPF}
\end{figure*}

Figure \ref{fig:syncVsAsyncHVOPF} shows the variation of the final median HV over 20 runs with changing lag ($T_U$) for the 118 and 300-bus OPF problem. It can be seen that for the synchronous case, the performance drops very fast since the number of effective FEs are lowered due to the lag. For the asynchronous case, the performance drop is much less. Thus, for problems with expensive function evaluations, asynchronous user interaction can be a good option.

\section{Conclusions}
This supplementary document with additional results and discussion, clearly supports our conclusions provided in the main study. Experimental results on all problems have consistently shown that moderate amount of rule usage combined with a moderate degree of rule adherence produce the best performance. Power law rules, individually, perform better than inequality rules. Ensemble-based repair operators provide a comparable performance to the best performing individual repair operators.

\clearpage
\bibliographystyle{IEEEtran}
\bibliography{./references/library}